\documentclass[10pt,twocolumn,letterpaper]{article}

\usepackage[pagenumbers]{cvpr}              

\usepackage[dvipsnames]{xcolor}

\definecolor{cvprblue}{rgb}{0.21,0.49,0.74}
\usepackage[pagebackref,breaklinks,colorlinks,citecolor=cvprblue]{hyperref}

\usepackage{caption}
\usepackage{multirow}
\usepackage{chemarrow}
\usepackage{array}
\usepackage{arydshln}

\usepackage[accsupp]{axessibility} %

\def\paperID{12069} %
\def\confName{CVPR}
\def\confYear{2024}

\title{Intrinsic Image Diffusion for Indoor Single-view Material Estimation}

\author{Peter Kocsis\\
Technical University of Munich\\
\and
Vincent Sitzmann\\
MIT EECS\\
\and
Matthias Nie{\ss}ner\\
Technical University of Munich\\
}

\definecolor{yellow2}{RGB}{255, 195, 55}

\makeatletter
\apptocmd\@maketitle{{\myfigure{}\par}}{}{}
\makeatother

\begin{document}
\newcommand{\myfigure}{
    \centering
    \vspace{-36pt}
    \vbox{%
    	\hsize\textwidth
    	\linewidth\hsize
    	\centering
    	\normalsize
    	\tt\href{https://peter-kocsis.github.io/IntrinsicImageDiffusion/}{peter-kocsis.github.io/IntrinsicImageDiffusion/}
    }
    \includegraphics[width=\textwidth]{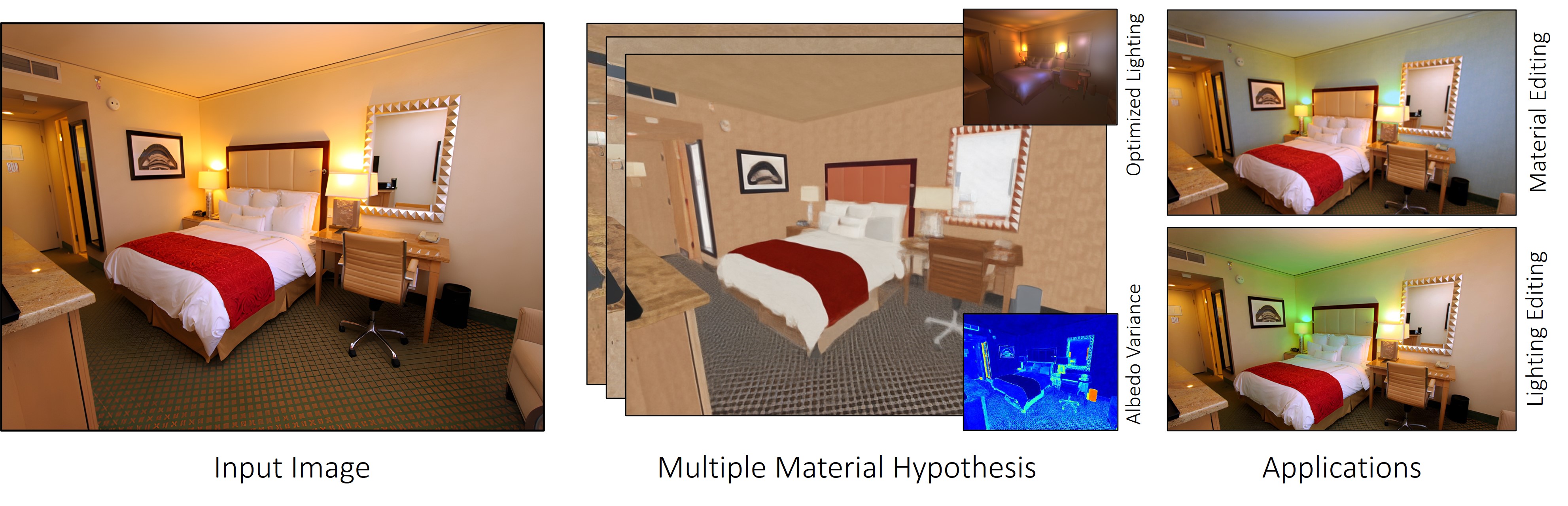}
        
  \vspace{-15pt}
\captionsetup{type=figure}\caption{\textbf{Intrinsic Image Diffusion. } 
        We present Intrinsic Image Diffusion for single-view material estimation of indoor scenes. 
        Since appearance decomposition is a highly ambiguous task, we propose to use a conditional generative model to predict multiple solutions and utilize the strong prior of recent diffusion models \cite{LDM}.
        Our approach gives detailed and consistent material estimations on complex indoor scenes, outperforming recent state-of-the-art methods and allows for high-quality controllable lighting optimization. 
        }
    \label{fig:teaser}
    \vspace{9pt}
}

\maketitle

\begin{abstract}
\vspace{-12pt}
We present Intrinsic Image Diffusion, a generative model for appearance decomposition of indoor scenes.
Given a single input view, we sample multiple possible material explanations represented as albedo, roughness, and metallic maps. 
Appearance decomposition poses a considerable challenge in computer vision due to the inherent ambiguity between lighting and material properties and the lack of real datasets. 
To address this issue, we advocate for a probabilistic formulation, where instead of attempting to directly predict the true material properties, we employ a conditional generative model to sample from the solution space. 
Furthermore, we show that utilizing the strong learned prior of recent diffusion models trained on large-scale real-world images can be adapted to material estimation and highly improves the generalization to real images. 
Our method produces significantly sharper, more consistent, and more detailed materials, outperforming state-of-the-art methods by $1.5dB$ on PSNR and by $45\%$ better FID score on albedo prediction. 
We demonstrate the effectiveness of our approach through experiments on both synthetic and real-world datasets.
\end{abstract}

\vspace{0pt}
\section{Introduction}
\vspace{-3pt}
Intrinsic image decomposition is a long-standing challenge in computer vision. 
The goal is to predict geometric, material, and lighting properties from a single image, enabling numerous downstream applications, such as content editing by changing material or lighting conditions. 
However, the fundamental difficulty comes from the fact that we can only observe the visual appearance of objects as a result of a complex interplay of lighting and material, making the decomposition inherently ambiguous.

Recent data-driven algorithms \cite{ComplexInvIndoor, IRISFormer, ComplexInvIndoorMC} show significant improvement by utilizing large-scale synthetic datasets \cite{ComplexInvIndoor} rendered photo-realistically \cite{ComplexInvIndoorMC}. 
They achieve impressive results in predicting the overall material properties of objects in the scene; however, accurately capturing high-frequency details continues to pose a challenge.

We believe the limitations observed in existing methods can be attributed to their deterministic treatment of the appearance decomposition problem. 
Decomposing the appearance into lighting and material properties is a highly ambiguous task since lighting properties can be baked into materials and vice versa.
By imposing the constraint of providing a single solution, deterministic models tend to predict the local or global average of the solution space, resulting in the aforementioned issues. 
We propose embracing the probabilistic nature of the problem, enabling a more comprehensive exploration of the solution space.

\begin{figure*}
  \centering
  \includegraphics[width=\textwidth]{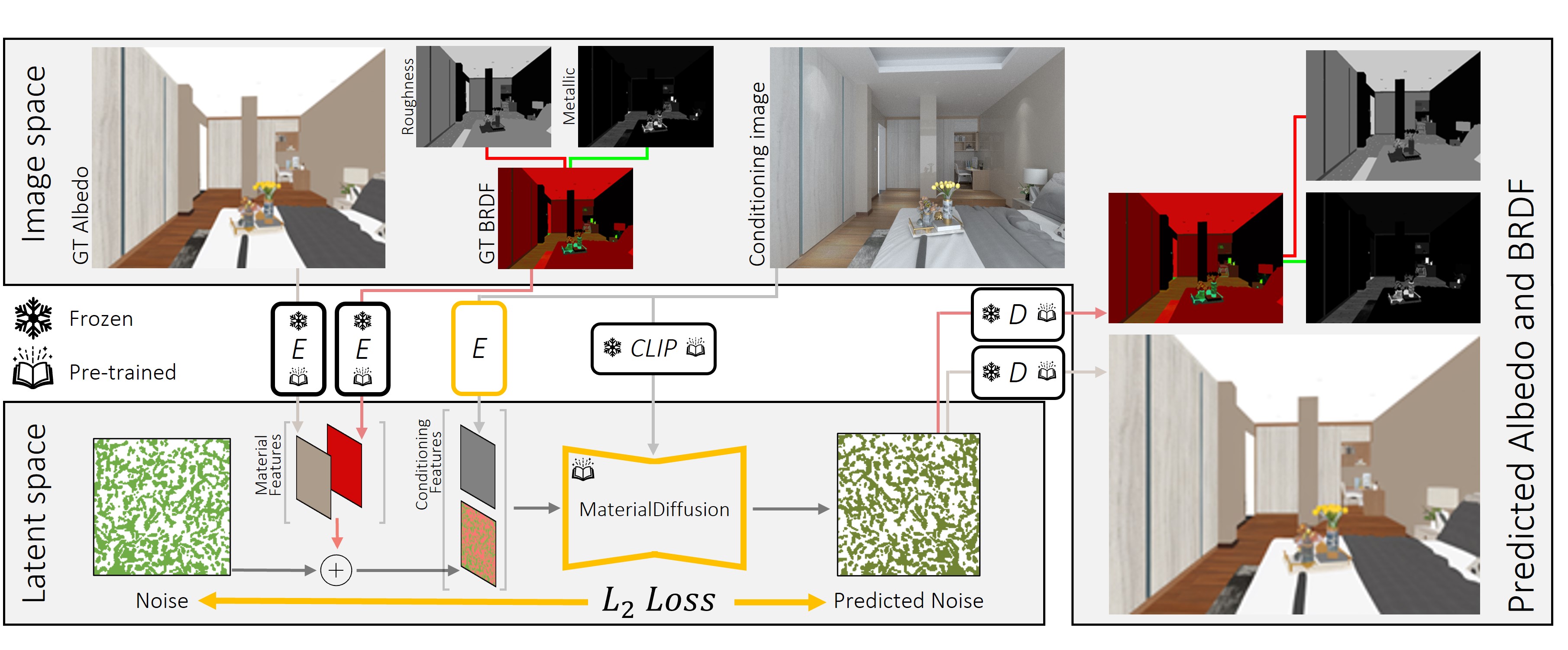}
  \vspace{-21pt}
  \caption{\textbf{Training pipeline.} 
  We train a conditional diffusion model to predict albedo and BRDF properties (roughness and metallic) given a single input image. We adapt the learned prior of Stable Diffusion \cite{LDM} by fine-tuning it on the synthetic InteriorVerse \cite{ComplexInvIndoorMC} dataset. 
  \textcolor{yellow2}{Models being trained are marked with yellow.} 
  (i) First, we separately encode the ground-truth (GT) albedo and BRDF properties with a fixed encoder to obtain the material feature maps. We also encode the conditioning image with a trainable encoder. 
  (ii) We add noise to the material features and use our conditional diffusion model to predicted the noise. 
  (iii) The training is supervised with L2 loss between the original and predicted noise. 
  (iv) Using the predicted noise, the predicted material properties can be decoded separately. }
  \label{fig:method:overview}
  \vspace{-12pt}
\end{figure*}

In this work, we propose to formulate the task of single-view appearance decomposition as a probabilistic problem. 
We develop a generative model capable of sampling from the solution space conditioned on the input image. 
Specifically, we build upon the recent advancements of generative models and train a conditional diffusion model \cite{DDPM, LDM} to predict the underlying material properties of a single image. 
With our approach, we aim to overcome the limitations of deterministic models and provide a more comprehensive and accurate representation of the underlying appearance decomposition problem.

Since no large-scale real-world intrinsic image decomposition dataset is available yet, previous methods usually train on synthetic data and may fine-tune on the real IIW dataset \cite{IIW} to reduce the domain gap. 
Instead, we utilize the strong learned prior of diffusion models and fine-tune a pre-trained Stable Diffusion V2 \cite{LDM} model to adapt its image prior to the task of material estimation using the photo-realistic synthetic InteriorVerse dataset of \cite{ComplexInvIndoorMC}.

Our approach is not restricted to predicting a single solution; thus, it provides clear and sharp possible material explanations for a given observation without local or global averaging. 
Furthermore, by adapting the strong prior of pre-trained diffusion models, we aim to bring two key benefits.
First, our objective is to reduce the domain gap caused by fine-tuning on synthetic imagery.
Second, the prior of real images can already contain important cues that can help in material estimation, such as semantic and perceptual information \cite{ComplexInvIndoorMC}. 
In summary, our main contributions are:
\begin{enumerate}
    \item We formulate the ambiguous appearance decomposition problem probabilistically and utilize diffusion models to sample solutions. 
    \item We leverage the real-world image prior of diffusion models for material estimation, achieving $77.6\%$ FID and $4.04dB$ PSNR improvement in albedo prediction. 
    \item Using our material predictions, we optimize for lighting as multiple point lights and a global environment map. 
\end{enumerate}

\vspace{-3pt}
\section{Related work} 
\vspace{-3pt}

\noindent\textbf{Material estimation. }
The field of intrinsic image decomposition in computer graphics and vision has a long and extensive history, with research dating back to the 1970s.
Earlier works aimed to decompose an image of a single object \cite{MITII} into reflectance (albedo) and shading (irradiance) maps using heuristic approaches. 
Some of these methods include considering the image gradients \cite{Retinex, jin2023estimating}, employing chromaticity clustering \cite{ChromacityClustering}, utilizing depth cues \cite{DBLP:conf/iccv/ChenK13}, or incorporating complex priors on the geometry, material, and lighting \cite{ShapeFromShading, DBLP:conf/cvpr/BarronM13}.  
\cite{IIW, SAW} took a significant step towards and introduced real-world datasets of indoor scenes for intrinsic image decomposition. 
These datasets include manually annotated relative comparisons of material properties, providing a valuable benchmark for evaluation. 
Additionally, \cite{IIW} proposed a conditional-random-field-based algorithm demonstrating impressive improvement over prior methods. 

Recent material estimation methods are usually part of a full inverse rendering framework \cite{I2SDF, MAIR, 3DSVLIndoor, INR, Nimier, IPT} and often use a learning-based algorithm for complex material predictions trained on increasingly photo-realistic datasets \cite{ComplexInvIndoor, ComplexInvIndoorMC} possibly combined with real data \cite{DBLP:conf/cvpr/LiS18a, CGIntrinsics}.
There has been great improvement in the model architectures, including UNet-based \cite{3DSVLIndoor, PBLightEditing, ComplexInvIndoorMC}, or cascaded networks \cite{DBLP:journals/tog/LiXRSC18, ComplexInvIndoor}, and Transformers \cite{IRISFormer}. 
However, all of these methods use a feed-forward model to decompose the input image.
We propose to formulate the task of appearance decomposition probabilistically and use a conditional generative model to sample multiple possible solutions. 

Another line of work focuses on material acquisition, i.e. obtaining high-quality material properties of a single object. 
\cite{UMat} introduced a similar idea for modeling the acquisition from a single scan probabilistically to address the disambiguity between specular and diffuse terms. 
Our method focuses on the single image multi-object material estimation and utilizes pre-trained image prior to better generalize.

\noindent\textbf{Generative models. }
The field of generative architectures has recently gained significant attention with diffusion-based models \cite{DBLP:conf/icml/Sohl-DicksteinW15, DBLP:journals/corr/abs-2107-00630, DDPM, DDIM, DBLP:journals/corr/abs-2011-13456, LDM} contributing notable advancements to the field as highlighted in the surveys of \cite{DiffusionSurvey,DiffusionSTAR}. 
These models have demonstrated remarkable fidelity utilizing large-scale text-image datasets \cite{LAION5B}.
Moreover, these models possess a strong and robust prior that has proven valuable for various downstream applications across different domains \cite{DiffusionSurvey}, such as relighting \cite{bhattad2023StylitGAN} and even for extracting intrinsic properties \cite{IntrinsicLora, DMP, DBLP:conf/nips/BhattadMHF23} determinisically.

In our research, we leverage diffusion models to generate multiple samples from a solution space, harnessing their inherent probabilistic behavior. 
Building upon this, we propose to exploit the learned prior of recent diffusion models and transfer it to our specific task of material estimation. 
Similar to approaches such as \cite{Zero123D, ControlNet}, we fine-tune StableDiffusion \cite{LDM}. 
However, our unique contribution lies in targeting the material space instead of the image space.
By adopting this approach, we aim to leverage the similarities between the material and image domains and utilize important cues for material estimation learned from real-world images, while simultaneously mitigating the domain gap.

\vspace{-3pt}
\section{Method}
\vspace{-3pt}

We aim to generate possible material properties $\hat{m} \in \mathbb{R}^{H \times W \times 5}$ consisting of albedo, roughness, and metallic from a single input image $x \in \mathbb{R}^{H \times W \times 3}$ ($\hat{m} \sim q(m|x)$). 
To this end, we train a latent diffusion model conditioned on the input image.
We provide more details about the diffusion process in \cref{sec:method:diffusion} and our material estimation in \cref{sec:method:material}. 
We illustrate our training pipeline in \cref{fig:method:overview}.

\noindent\textbf{Dataset}
We train our model on the photo-realistic InteriorVerse \cite{ComplexInvIndoorMC} dataset, which consists of more than $50000$ rendered images with albedo, roughness, and metallic maps from over $4000$ high-quality scenes. 
All the images and material maps are HDR in linear space. 

\subsection{Material representation}
\label{sec:method:material}
Estimating the material from a single image requires to decouple the material properties from the lighting. 
This task brings two main challenges.
First, the appearance decomposition is highly ambiguous. 
Multiple joint material and lighting explanations can explain the same input view.
The most obvious ambiguity is the scaling; the material can be scaled with an arbitrary global value, whereas the lighting can be downscaled accordingly. 
Furthermore, shading effects, such as shadows and specular highlights, can be baked into the material, resulting in valid explanations for different lighting conditions.
Second, the material models used in computer graphics are hand-crafted approximations of real-world reflectance properties. 
Although they might be physically based, they don't exactly capture the real reflectance, leading to a representational gap. 

In our work, we follow \citet{ComplexInvIndoorMC} and represent the material with the physically-based GGX microfacet BRDF \cite{GGX} consisting of albedo, roughness, and metallic properties. 
We create an image of the BRDF parameters, where channel R is the roughness, G is metallic, and B is always zero. 
To get the latent material features for the diffusion process, we separately encode the BRDF properties and the albedo and concatenate them, as shown in \cref{fig:method:overview}.

\begin{figure*}[t!]
  \centering
  \setlength\tabcolsep{1.25pt}
  \begin{tabular}{c|cc|c|c}
        \includegraphics[width=0.19\textwidth]{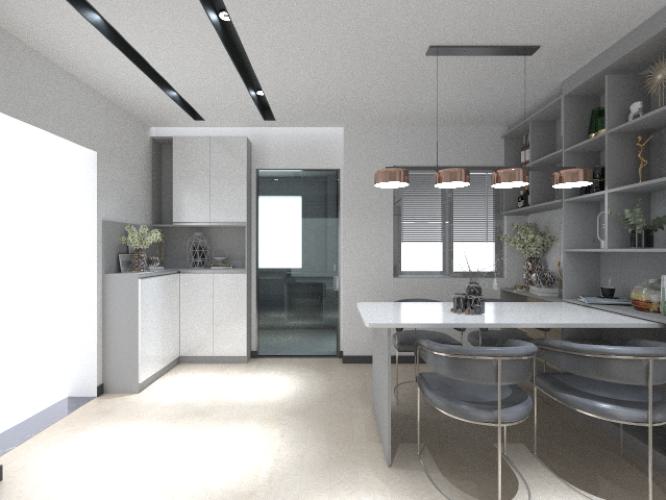}&
        \includegraphics[width=0.19\textwidth]{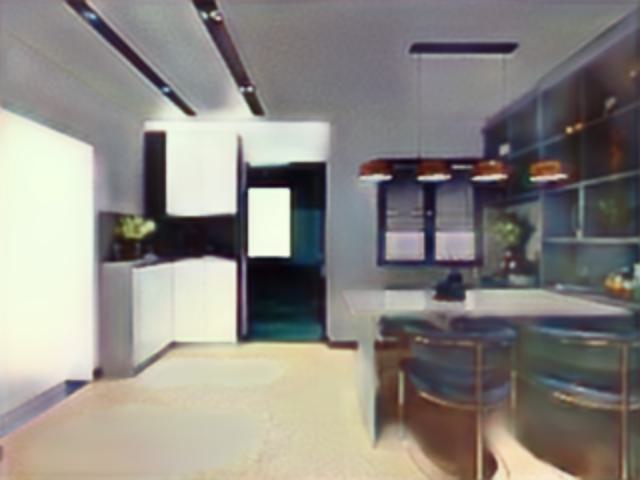}&
        \includegraphics[width=0.19\textwidth]{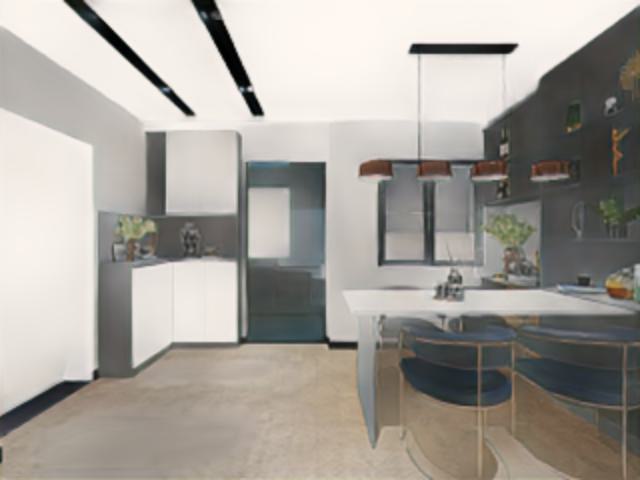}&
        \includegraphics[width=0.19\textwidth]{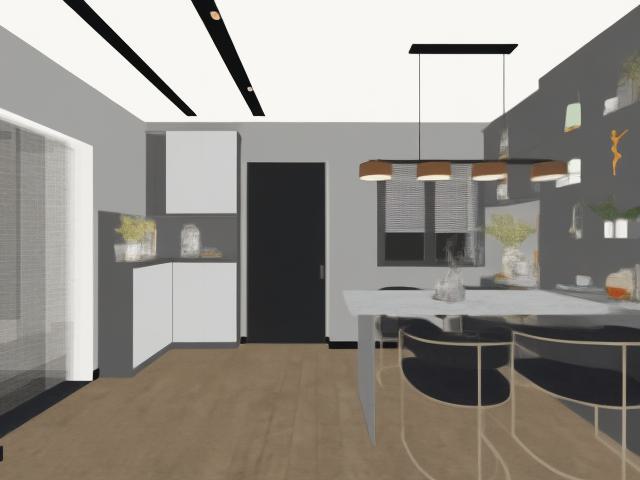}&
        \includegraphics[width=0.19\textwidth]{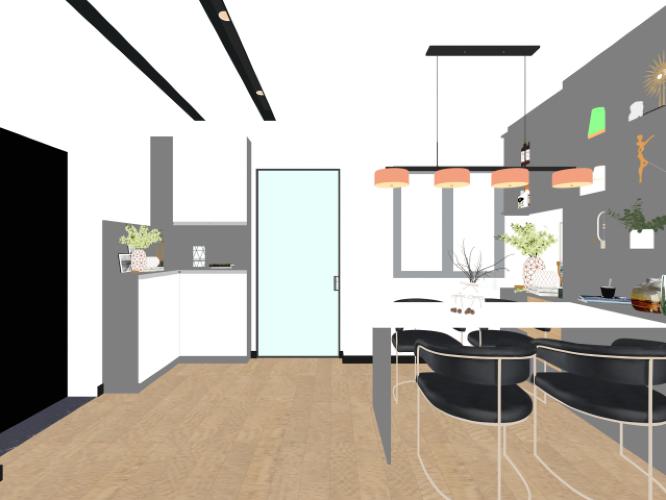} \\

        \includegraphics[width=0.19\textwidth]{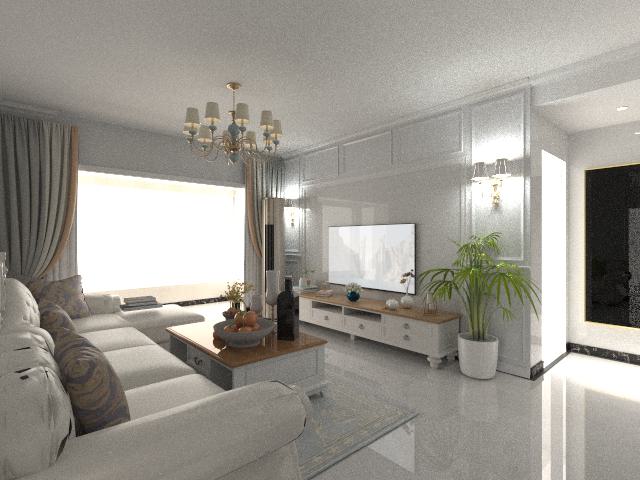}&
        \includegraphics[width=0.19\textwidth]{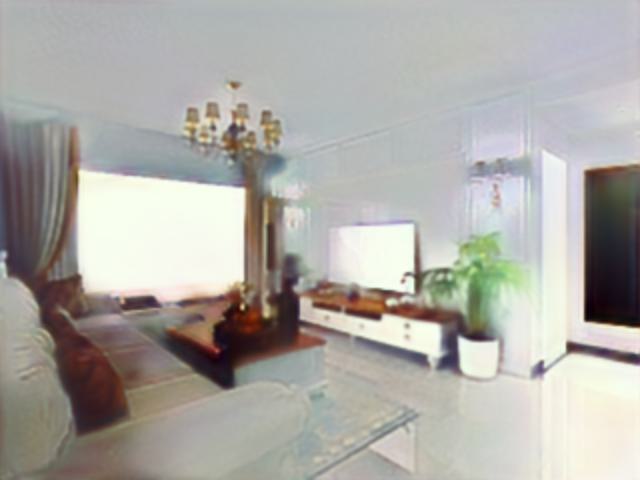}&
        \includegraphics[width=0.19\textwidth]{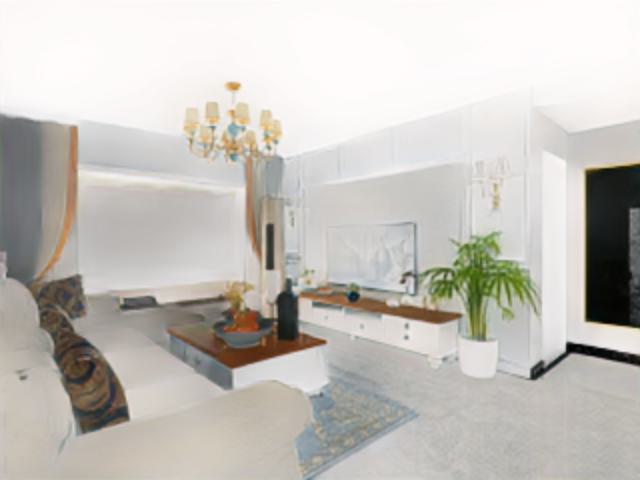}&
        \includegraphics[width=0.19\textwidth]{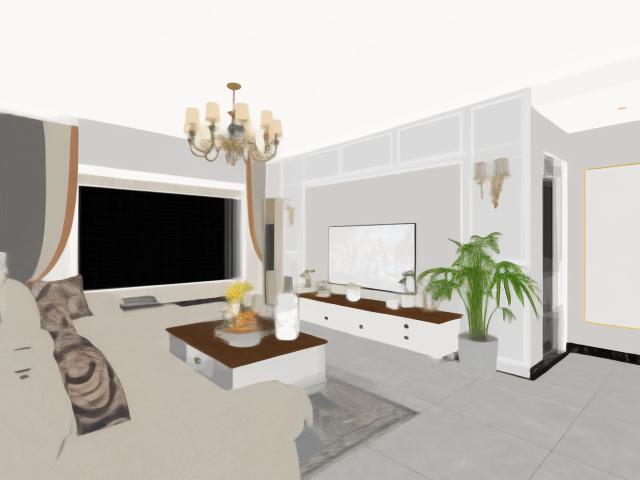}&
        \includegraphics[width=0.19\textwidth]{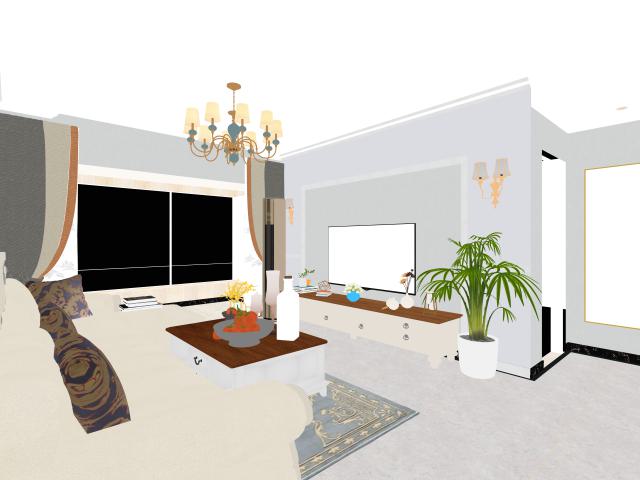} \\

        \includegraphics[width=0.19\textwidth]{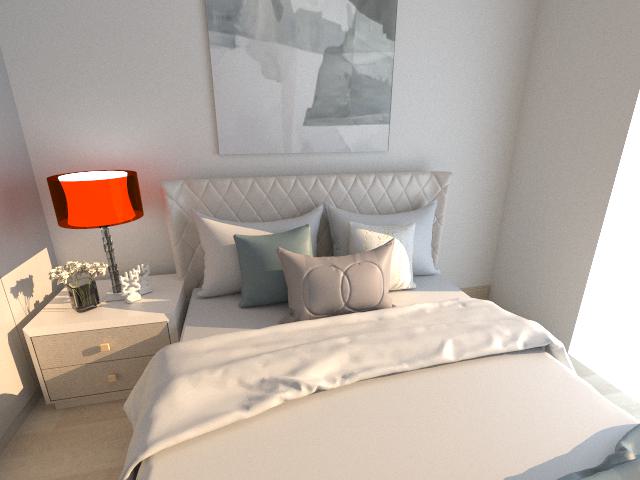}&
        \includegraphics[width=0.19\textwidth]{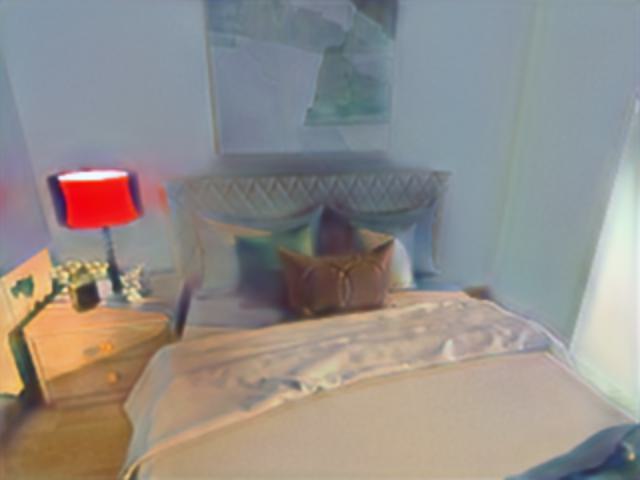}&
        \includegraphics[width=0.19\textwidth]{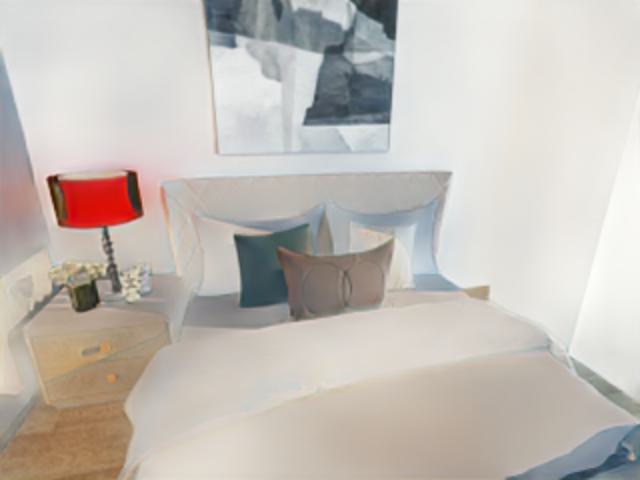}&
        \includegraphics[width=0.19\textwidth]{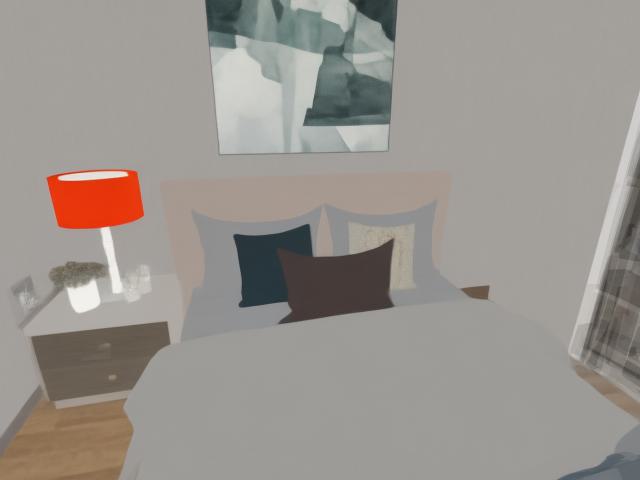}&
        \includegraphics[width=0.19\textwidth]{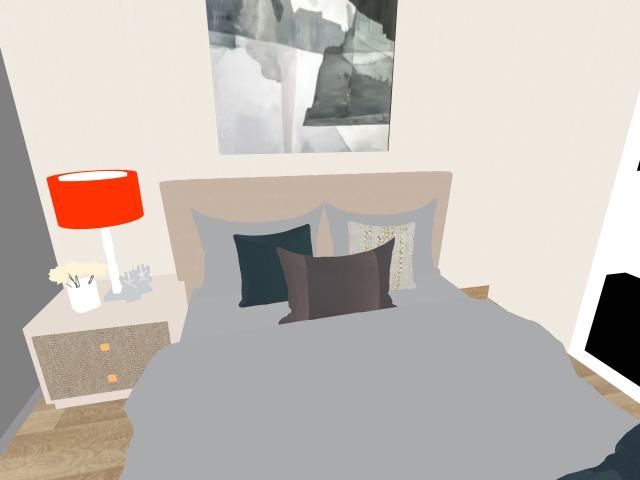} \\
        Input & \citet{ComplexInvIndoor} & \citet{ComplexInvIndoorMC} & Ours - Mean
        & GT
  \end{tabular}
  \vspace{-6pt}
  \caption{\textbf{Synthetic evaluation.} 
  We qualitatively compare the predicted albedos of our to the baselines \cite{ComplexInvIndoor, ComplexInvIndoorMC} on the InteriorVerse dataset \cite{ComplexInvIndoorMC}. 
  Both of the baselines produce smoothed results, often with baked-in lighting, specularities, or shadows. 
  In contrast, our method gives sharp and detailed predictions with consistent textures. 
  See supplementary for more results with roughness and metallic predictions.}
  \label{fig:experiments:synthetic:main}
  \vspace{-6pt}
\end{figure*}

\begin{figure}
  \setlength\tabcolsep{1.25pt}
  \centering
  \begin{tabular}{cc|c}
        \includegraphics[width=0.22\columnwidth]{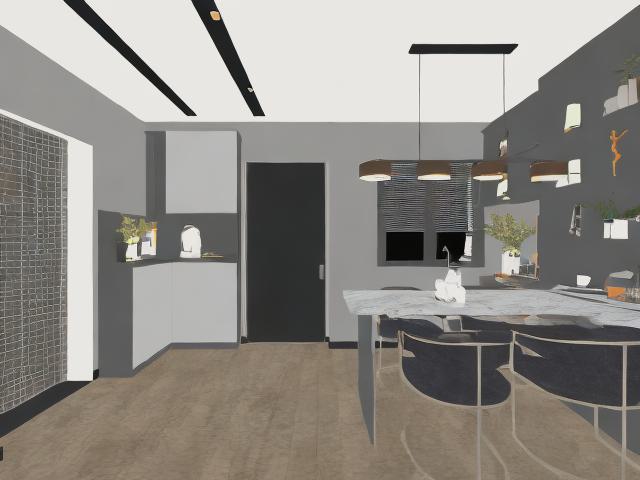} &
        \includegraphics[width=0.22\columnwidth]{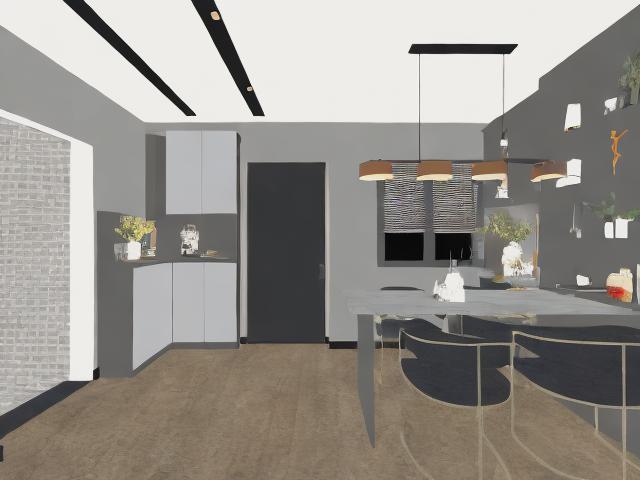} & 
        \multirow{8}[2]{*}[38.5pt]{\includegraphics[width=0.498\columnwidth]{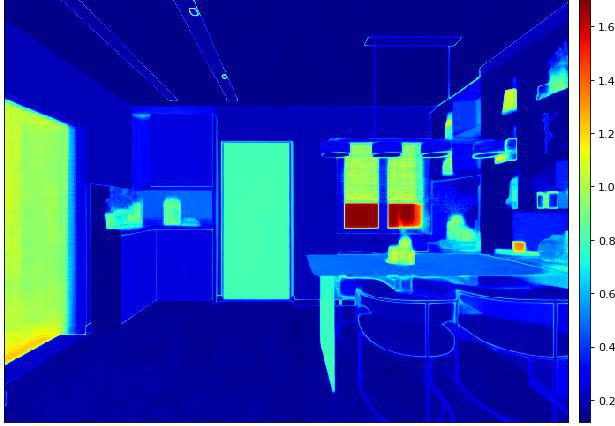}}\\
        \includegraphics[width=0.22\columnwidth]{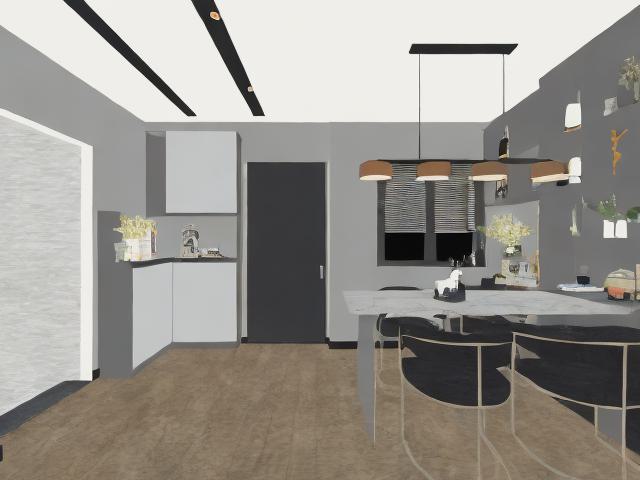} &
        \includegraphics[width=0.22\columnwidth]{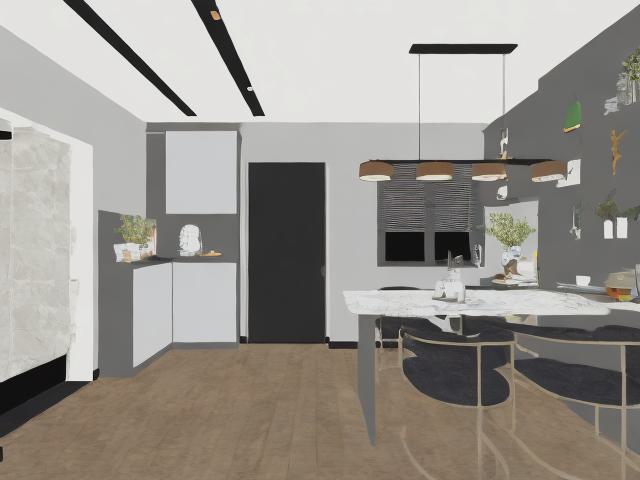} & \\
        \multicolumn{2}{c|}{Samples} & Variance 
  \end{tabular}
  \vspace{-6pt}
  \caption{\textbf{Sample diversity.} We show multiple samples for a single scene and visualize the variance of the images across $100$ samples. Specular and emissive objects have higher variance since their material properties are highly ambiguous. }
  \label{fig:experiments:synthetic:diversity}
  \vspace{-6pt}
\end{figure}

\subsection{Training}
\label{sec:method:diffusion}
Diffusion models have recently shown excellent capabilities in learning a prior over high-dimensional input data \cite{DDPM, LDM}.
These methods use an iterative denoising process to transform pure Gaussian noise into a data sample from the training distribution. 
Training them on large-scale datasets \cite{LAION5B} gives them an extremely strong prior over real-world images. 

Our goal is to utilize their learned prior and adjust it to the material estimation task. 
We fine-tune the pre-trained text-conditional Stable Diffusion V2 (SD) \cite{LDM} on our task on synthetic data, similar to \cite{Zero123D}.
As \cite{LDM}, we follow the latent diffusion architecture with pre-trained frozen encoder $\mathcal{E}$ and decoder $\mathcal{D}$. 
To condition our diffusion model, we input the image information in two ways to the model.
First, we encode it with a trained encoder $\mathcal{E}^*$ and concatenate it to the input noisy features. 
Our encoder $\mathcal{E}^*$ has the same architecture as the pre-trained encoder $\mathcal{E}$, but it is initialized randomly and gives $3$-channel features; thus, the overall input channel size is $11$. 
We found that naive downscaling to the latent dimension, as in the original depth-conditioned SD, causes loss in the high-frequency details. 
Using the frozen pre-trained encoder $\mathcal{E}$ performs worse, suggesting that the task of material estimation requires a different set of features for optimal predictions.  
Second, we use the CLIP \cite{CLIP, OpenCLIP} image embedding as a cross-attention conditioner. 

In each training step, we take a batch of images with the corresponding albedo and BRDF maps. 
We encode both material maps separately, then concatenate them in the latent space to get the material features. 
For each image, we sample a timestep $t \sim [1, 1000]$ and noise the material features accordingly \cite{LDM}. 
We predict the added noise $\epsilon$ conditioned on the input image $x$. 
Our training objective is to minimize the distance between the original and the predicted noise, as shown on \cref{fig:method:overview}. 
\begin{equation}
    L = \mathbb{E}_{m,\epsilon \sim \mathcal{N}(0,I),t}[||\epsilon - \epsilon_\theta(\mathcal{E}(m) + \epsilon, t, x||_2^2]
\end{equation}

In contrast to prior methods, our approach introduces a novel strategy for mitigating the domain gap.
Prior work trains on synthetic data and subsequently fine-tunes on a smaller real-world dataset (IIW \cite{IIW}) to adapt the synthetic prior to real-world scenarios. 
We take a different approach by fine-tuning a prior trained on large-scale real image data and adapting it to our specific task on synthetic data \cite{Zero123D}.

\begin{figure}
  \centering
  \setlength\tabcolsep{1.25pt}
  \begin{tabular}{ccccc|c}
        \includegraphics[width=0.15\columnwidth]{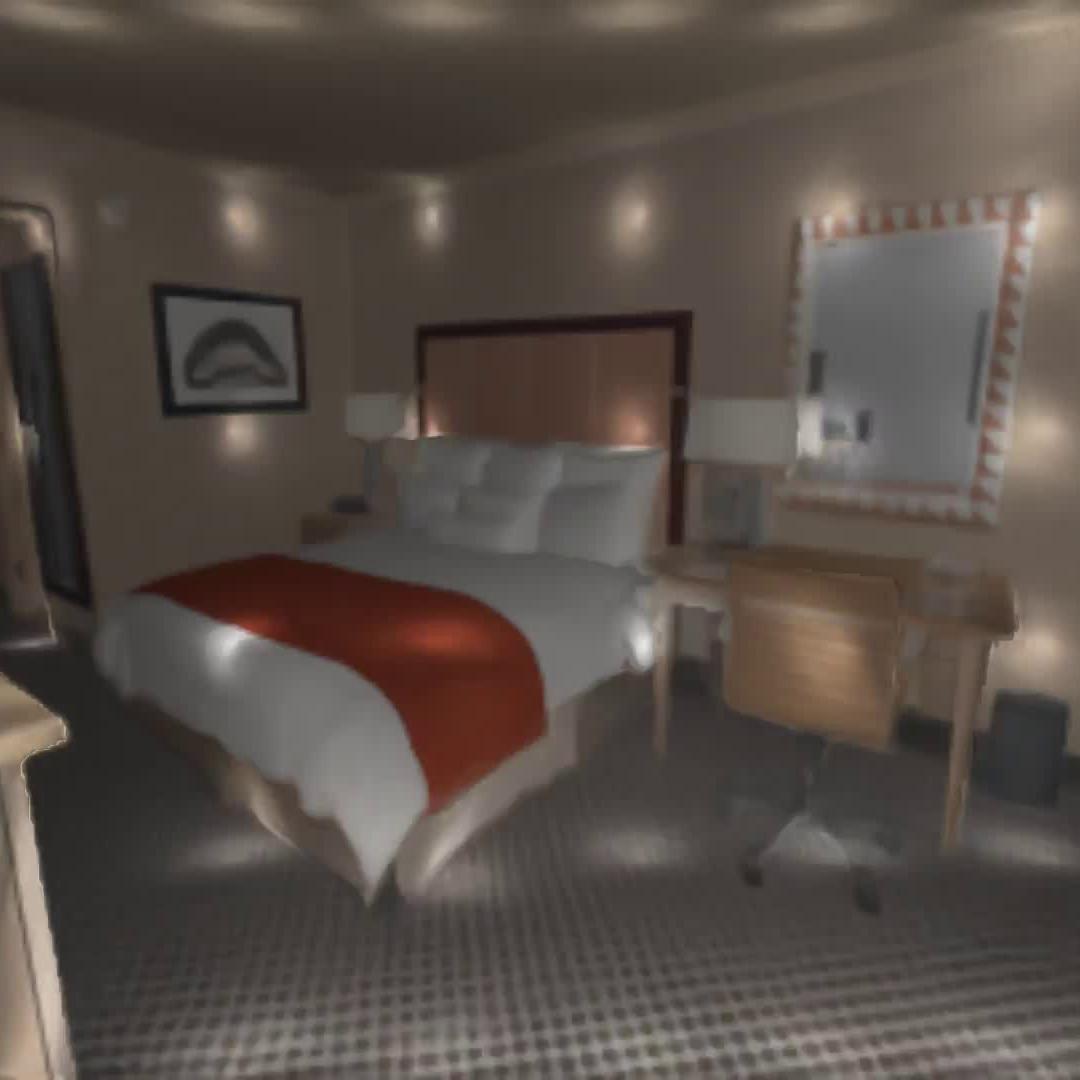}&
        \includegraphics[width=0.15\columnwidth]{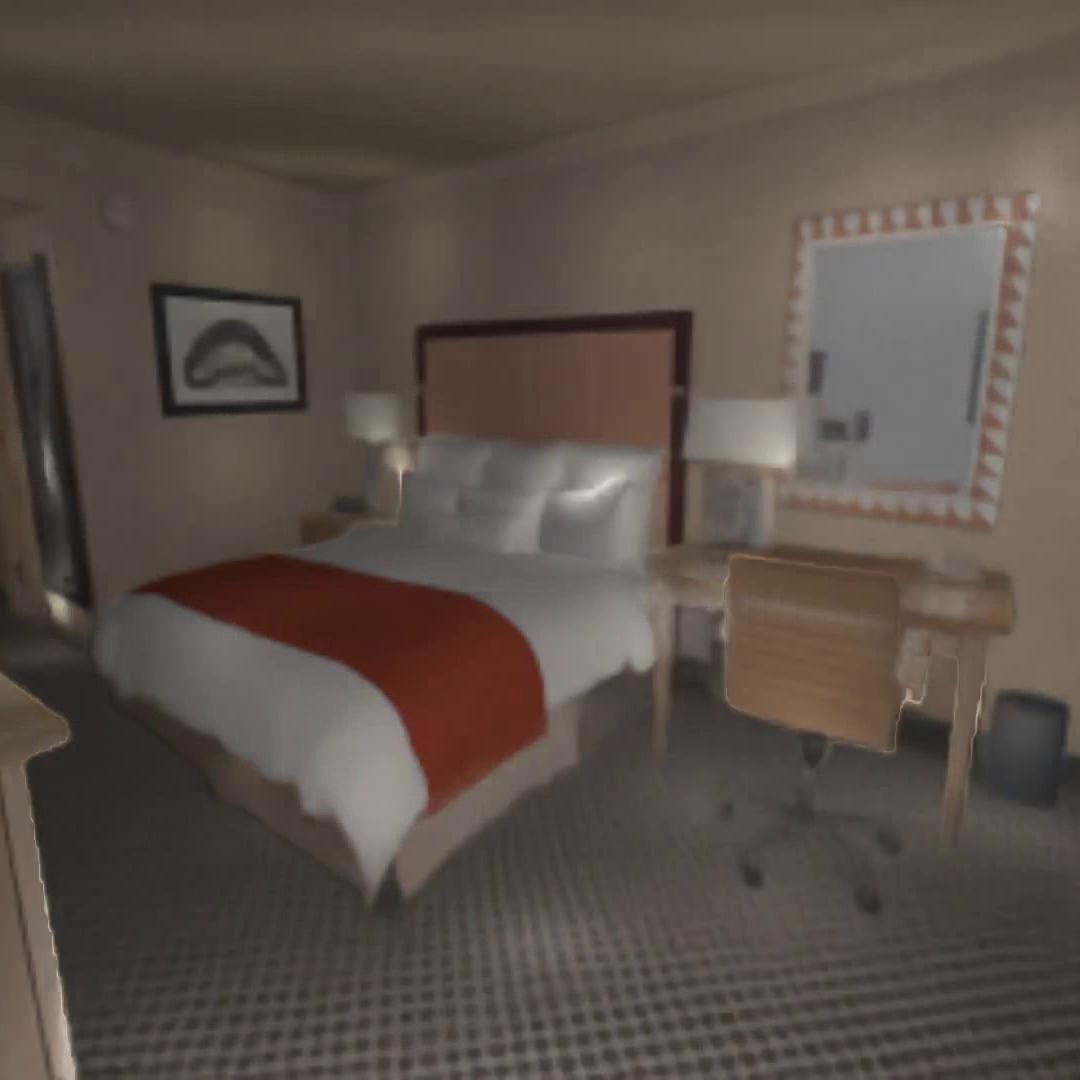}&
        \includegraphics[width=0.15\columnwidth]{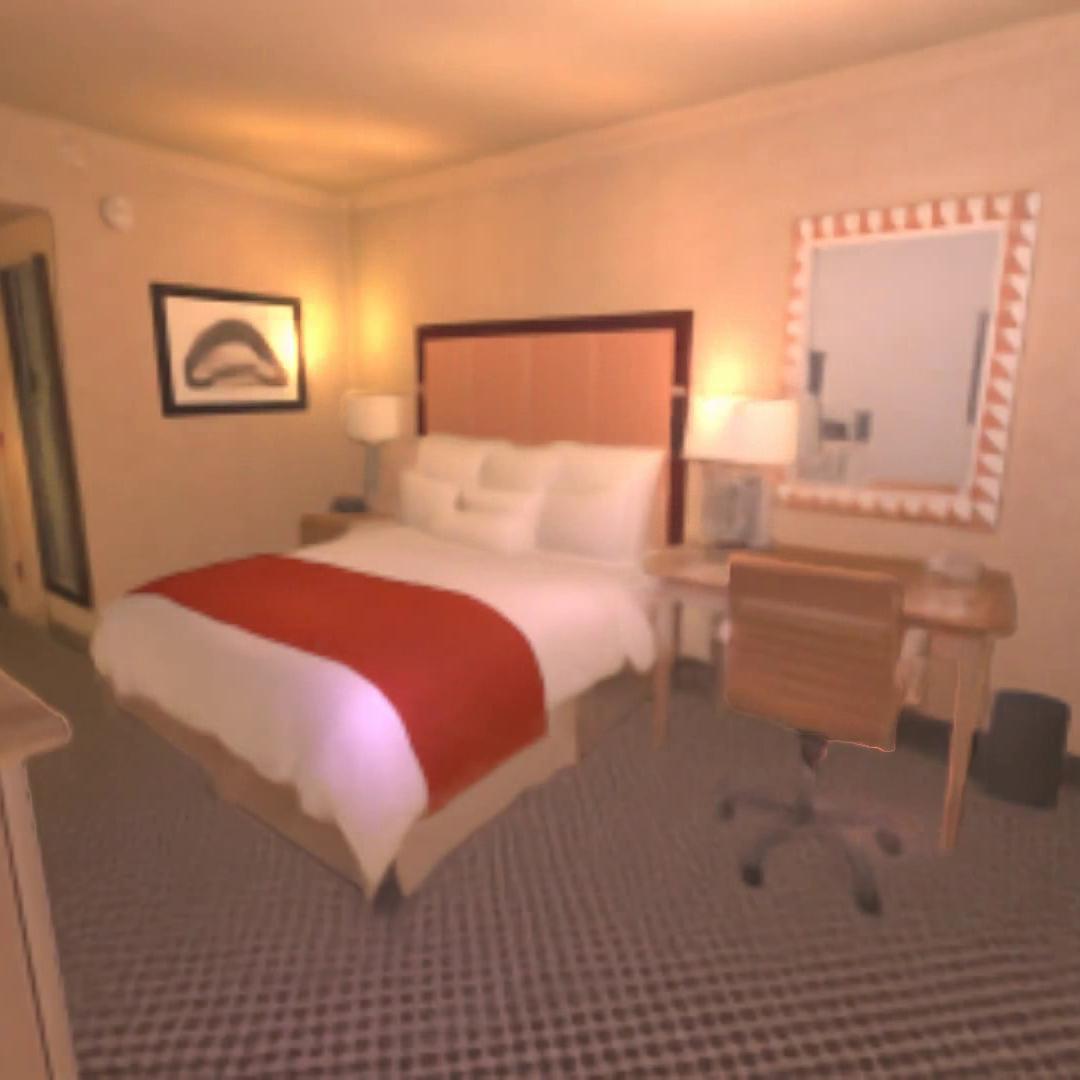}&
        \includegraphics[width=0.15\columnwidth]{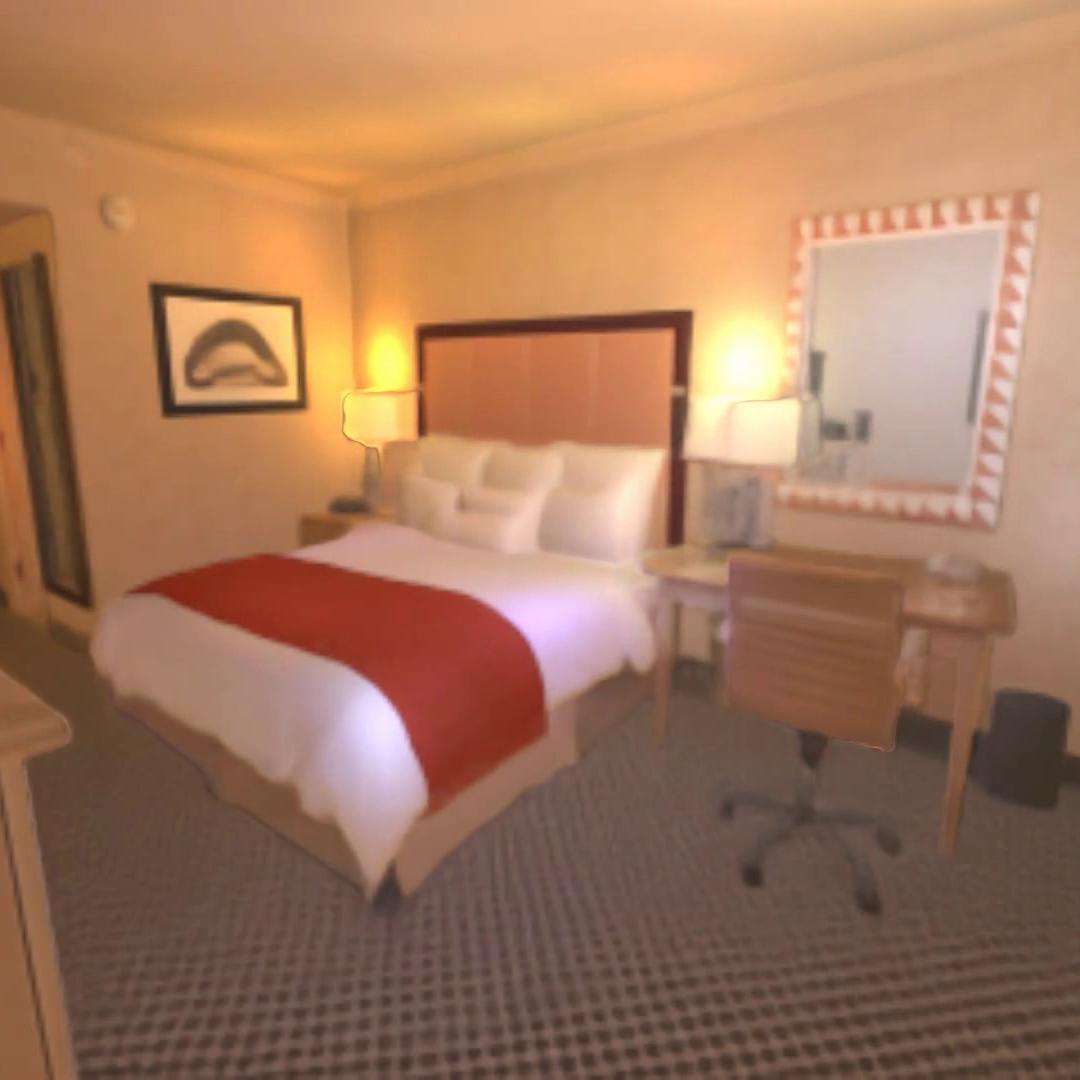}&
        \includegraphics[width=0.15\columnwidth]{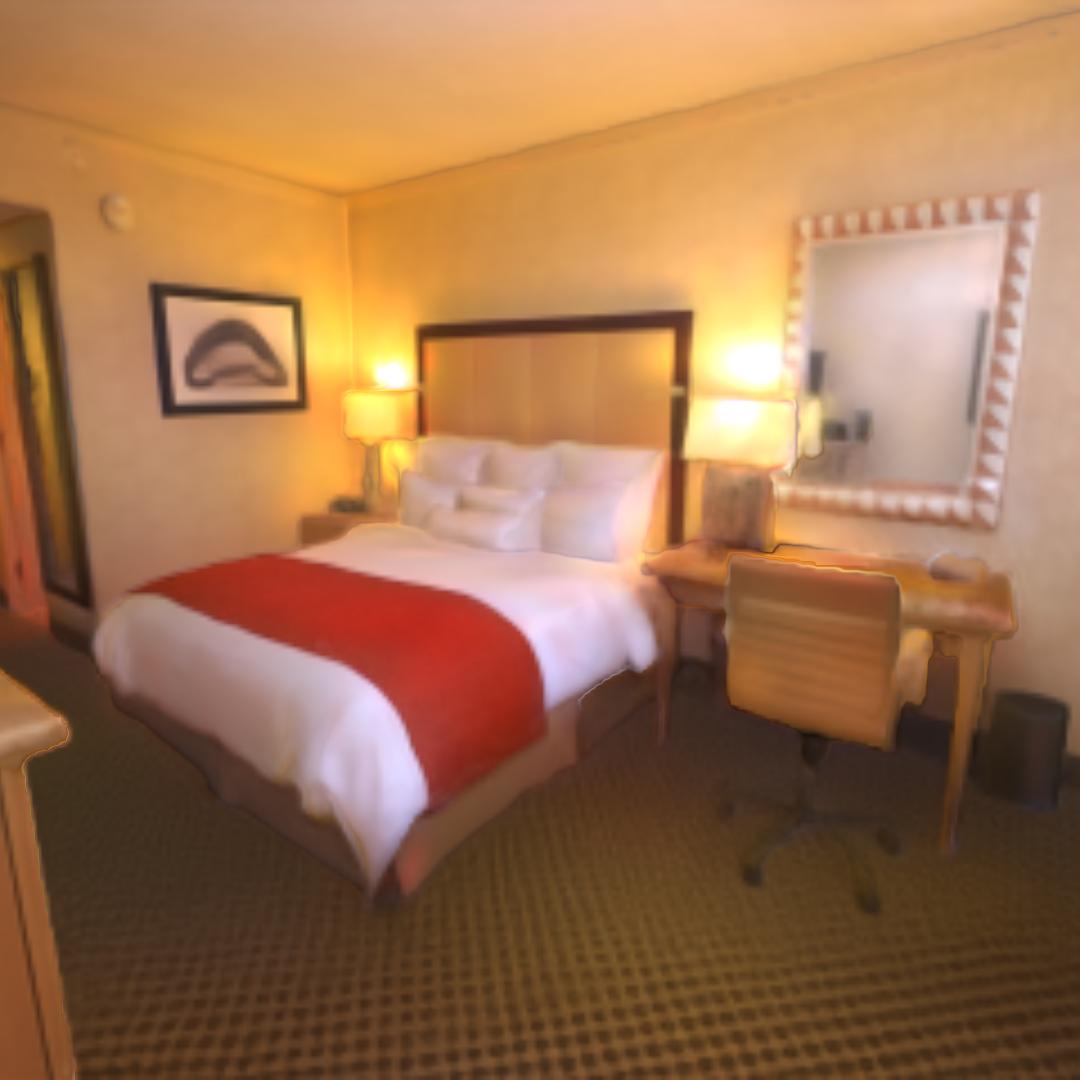}&
        \includegraphics[width=0.15\columnwidth]{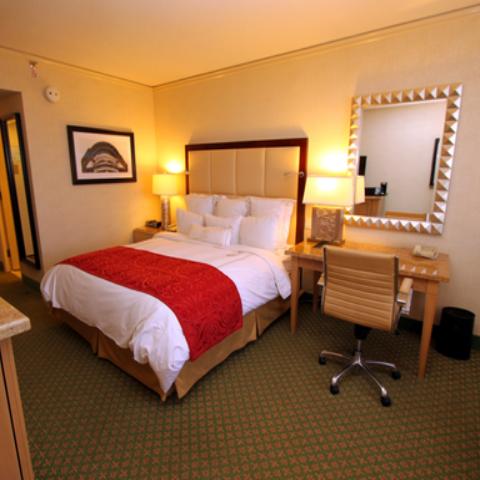} \\
        
        \multicolumn{5}{c|}{$\autorightarrow{}{\hspace*{0.75\columnwidth}}$}\vspace{-11pt} & \\
        \multicolumn{5}{c|}{Optimization} & GT
  \end{tabular}
  \vspace{-6pt}
  \caption{\textbf{Lighting Optimization.} 
  We optimize for $N_{light}$ point light sources with SG emission profile together with global incident lighting. 
  Our representation is expressive enough to capture detailed emissions yet controllable for relighting purposes. 
  }
  \label{fig:method:lighting}
  \vspace{-6pt}
\end{figure}

\subsection{Inference}
At test time, we use the regular diffusion process \cite{LDM} to sample material features conditioned on a single image. 
We split the predicted material features into albedo and BRDF features and decode them separately to obtain a possible material explanation. 
We sample multiple possible solutions for one image and decode the albedo and BRDF predictions separately. 

\subsection{Lighting optimization}
\label{sec:method:lighting}
Given our high-fidelity, consistent material predictions, we optimize for the lighting of the scene. 
We use OmniData \cite{OmniData} to estimate the geometry.
We optimize for global Spherical Gaussians (SG) \cite{ComplexInvIndoor} incident lighting and $N_{light}$ point light sources with SG emission profile, resulting in a controllable yet expressive representation. 
We give more details in the supplemental. 
The optimization progress is shown in \cref{fig:method:lighting}.

\section{Experiments}

In the following, we evaluate our method on both synthetic (\cref{sec:experiment:synthetic}) and real-world (\cref{sec:experiment:real}) datasets.
Furthermore, we ablate the effect of the key design choices of our method in \cref{sec:experiment:ablations}. 
We show qualitative albedo prediction and quantitative material prediction (albedo, roughness, metallic) results. 
More qualitative results with additional ablation can be found in our supplementary material. 

\vspace{3pt}
\noindent\textbf{Experimental setting. }
We fine-tune the pre-trained Stable Diffusion V2 model \cite{LDM} on the InteriorVerse dataset \cite{ComplexInvIndoorMC} with batch size of $40$ for $250$ epochs with a fixed learning rate of $1e\text{-}5$ using the AdamW optimizer \cite{AdamW}.
To make our method robust against exposure changes, we normalize the input image to have a mean of $0.5$ and clip them for the range $[0,1]$.
Furthermore, as expected by the pre-trained encoder $\mathcal{E}$, we transform all input maps to the range $[-1,1]$.
We use random crops of $256 \times 256$ resolution without any padding. 
The training takes $\sim 6$ days on $4$ A6000 GPUs. 

Although we train our model on $256 \times 256$ resolution crops, we evaluate our model on the full image resolution during inference, which is $640 \times 480$ for InteriorVerse \cite{ComplexInvIndoorMC} and $\sim 512 \times 384$ for IIW \cite{IIW}. 
If not stated otherwise, we sample $10$ possible materials for each input image.
We use DDIM sampling \cite{DDIM} with $50$ steps, which takes $17$ seconds for $10$ samples. 

\vspace{3pt}
\noindent\textbf{Baselines. } 
We compare our method with \cite{ComplexInvIndoorMC} and \cite{ComplexInvIndoor}, two recent state-of-the-art inverse rendering frameworks for indoor scenes. 
These methods use a UNet architecture \cite{UNet} to deterministically predict geometric and material properties from a single image. 
Furthermore, we also report results for the optimization-based method of \cite{IIW}; however, this method was tuned on the IIW dataset \cite{IIW}. 

\vspace{3pt}
\noindent\textbf{Metrics. } 
Since we model the appearance decomposition probabilistically, we cannot directly compare our prediction to the GT. 
Therefore, we sample multiple samples for the same view and report the FID score of the predicted albedos alongside the mean of PSNR/SSIM/LPIPS metrics, the metrics of the mean sample, and the nearest sample. 
We calculate the mean sample in the image space. 

Furthermore, albedo is a scale-invariant feature due to the inherent ambiguity between lighting and material intensity. 
Following \cite{ComplexInvIndoor}, we report scale-invariant metrics for the albedo evaluations.

\begin{figure*}
  \setlength\tabcolsep{1.25pt}
  \centering
  \resizebox{0.95\textwidth}{!}{
  \begin{tabular}{c|c|c}
        \begin{tabular}{c}
        IIW \cite{IIW} \\
        \includegraphics[width=0.20\textwidth]{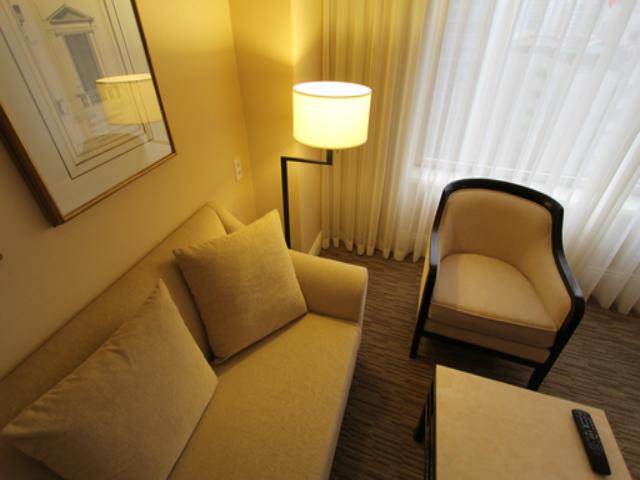}\end{tabular}&
        \begin{tabular}{c}
        \rotatebox{90}{\citet{ComplexInvIndoor}}
        \includegraphics[width=0.205\textwidth]{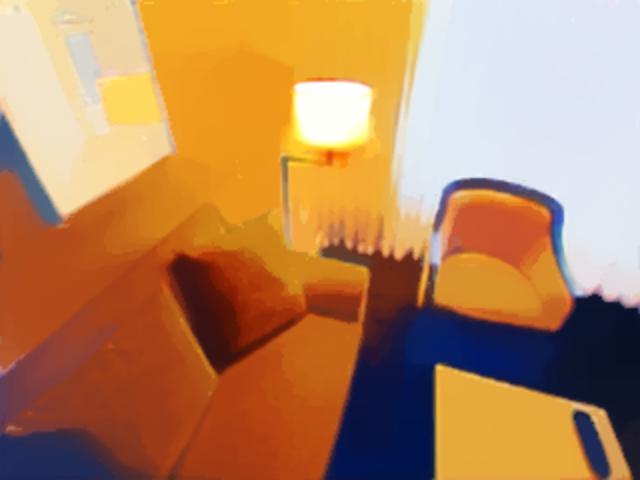}\\
        \rotatebox{90}{\citet{ComplexInvIndoorMC}}
        \includegraphics[width=0.205\textwidth]{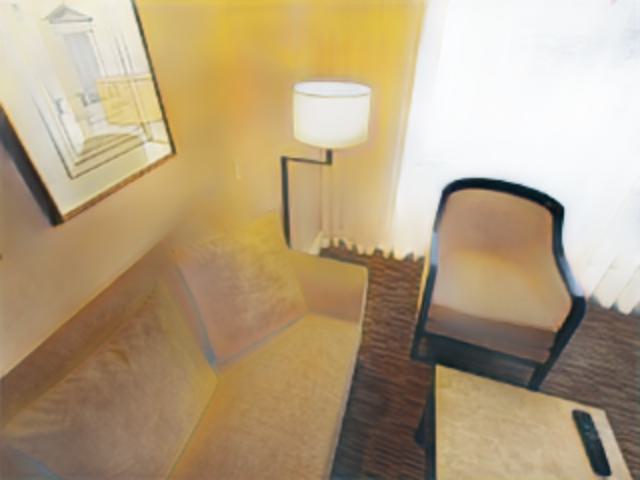}
        \end{tabular}
        &
        \begin{tabular}{c}
        \rotatebox{90}{Mean}
        \includegraphics[width=0.24\textwidth]{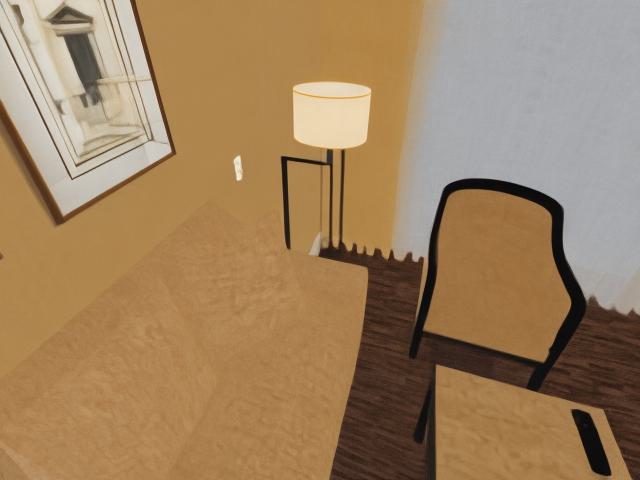}
        \rotatebox{90}{Variance}
        \includegraphics[width=0.26\textwidth]{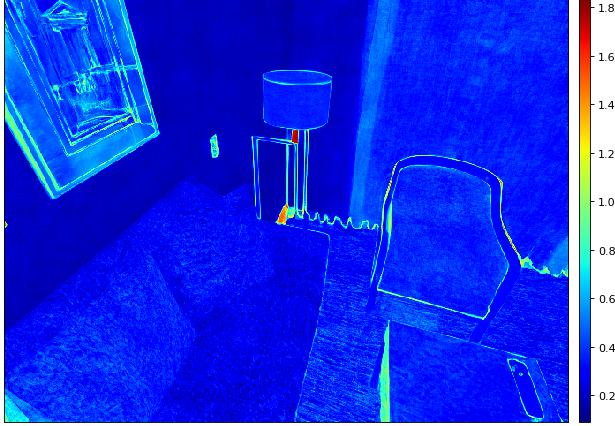}\\
        \rotatebox{90}{Samples}
        \includegraphics[width=0.172\textwidth]{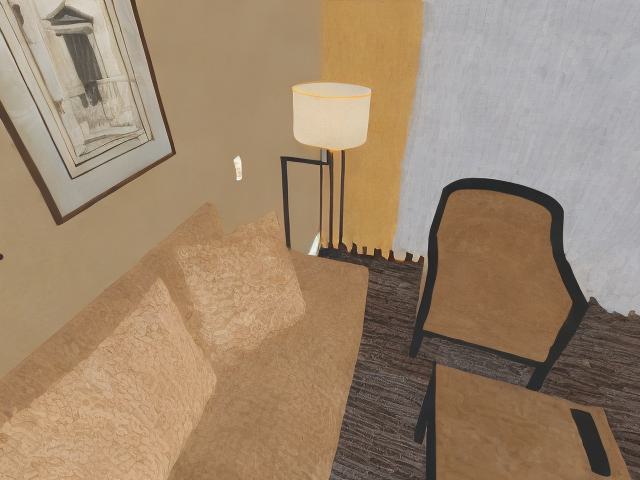}
        \includegraphics[width=0.172\textwidth]{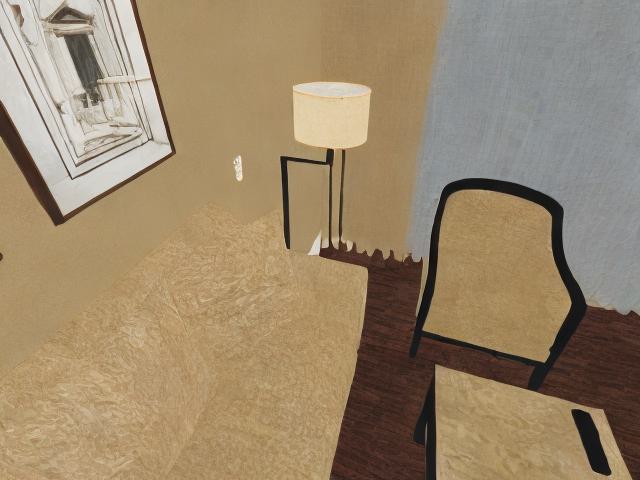}
        \includegraphics[width=0.172\textwidth]{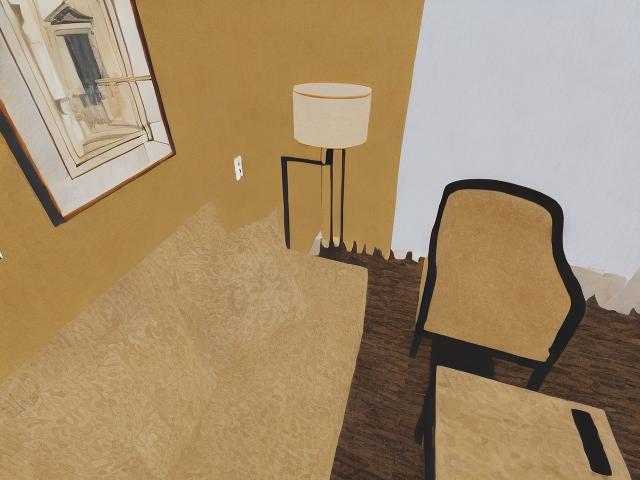}
        \end{tabular} \\

        \begin{tabular}{c}
        IIW \cite{IIW} \\
        \includegraphics[width=0.20\textwidth]{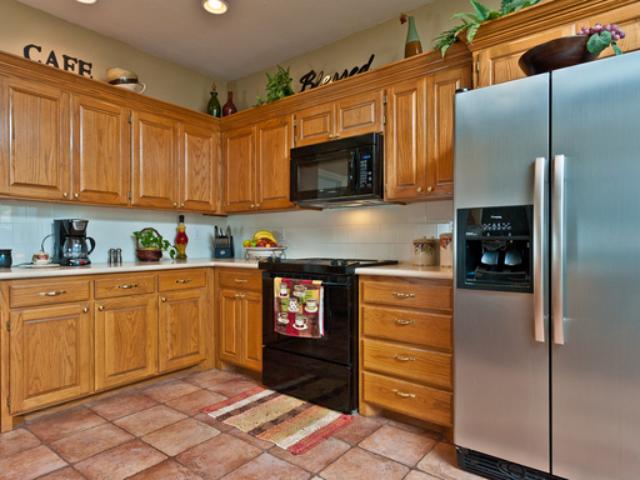}\end{tabular}&
        \begin{tabular}{c}
        \rotatebox{90}{\citet{ComplexInvIndoor}}
        \includegraphics[width=0.205\textwidth]{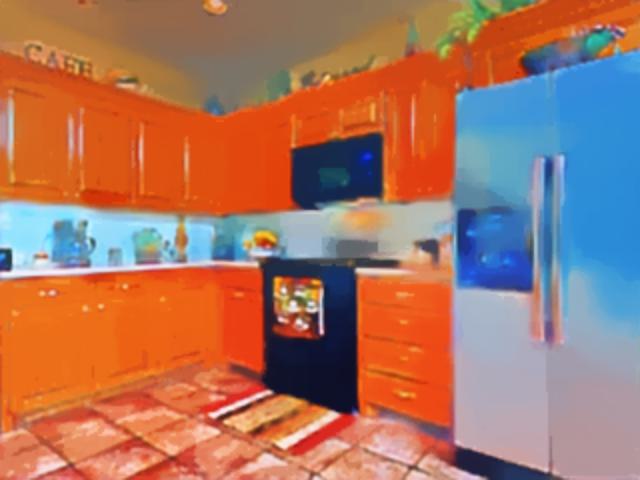}\\
        \rotatebox{90}{\citet{ComplexInvIndoorMC}}
        \includegraphics[width=0.205\textwidth]{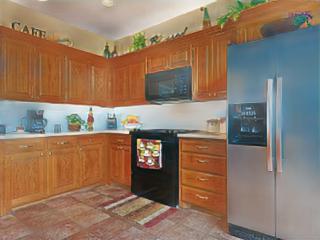}
        \end{tabular}
        &
        \begin{tabular}{c}
        \rotatebox{90}{Mean}
        \includegraphics[width=0.24\textwidth]{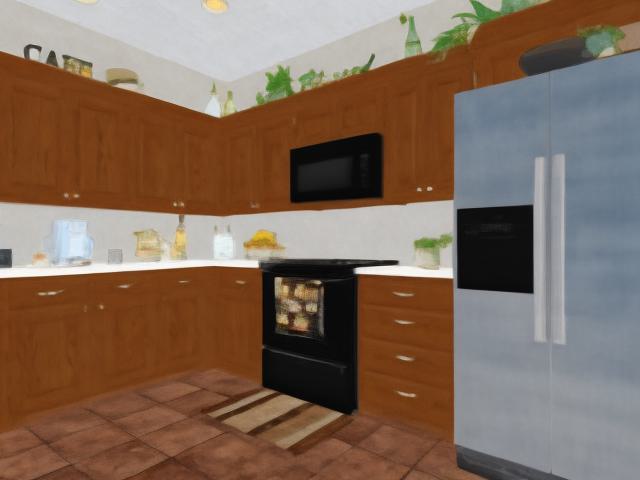}
        \rotatebox{90}{Variance}
        \includegraphics[width=0.26\textwidth]{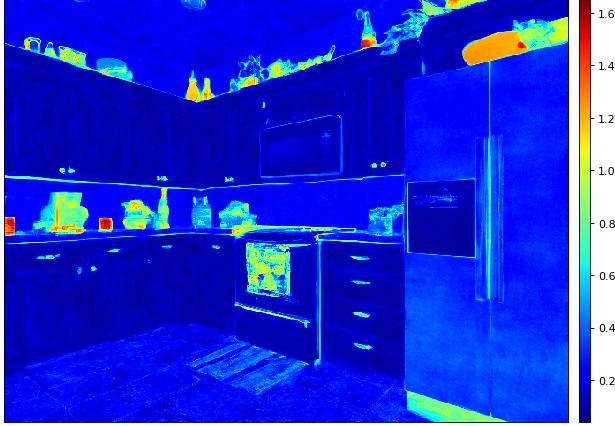}\\
        \rotatebox{90}{Samples}
        \includegraphics[width=0.172\textwidth]{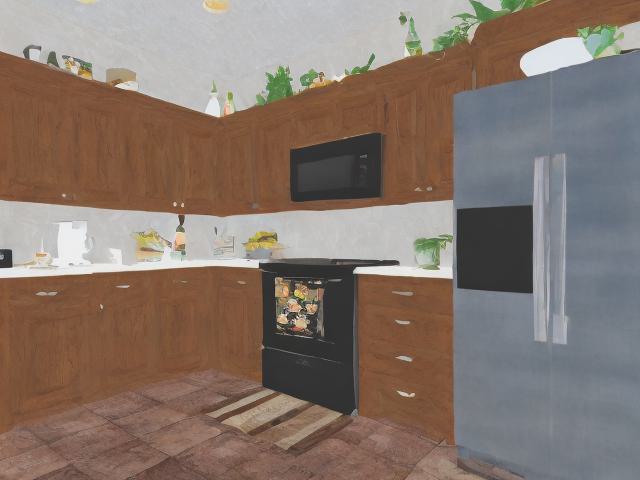}
        \includegraphics[width=0.172\textwidth]{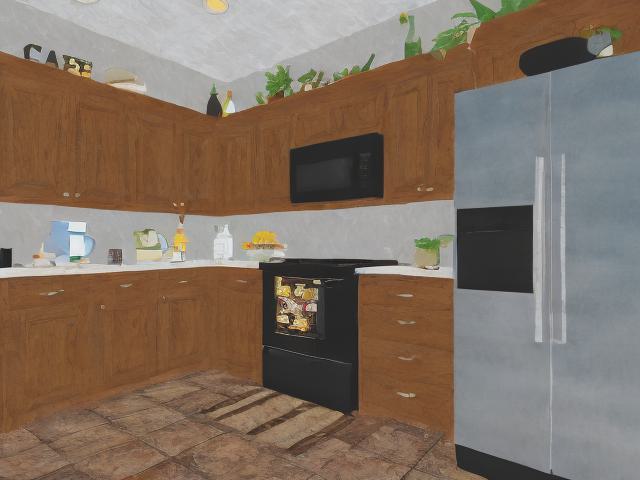}
        \includegraphics[width=0.172\textwidth]{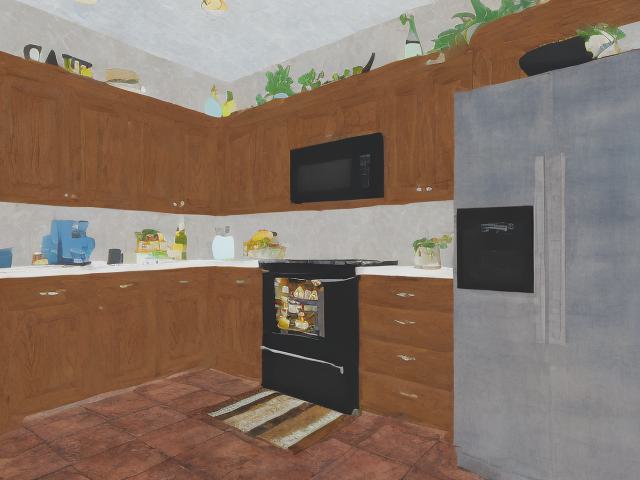}
        \end{tabular} \\
        \midrule

        \begin{tabular}{c}
        ScanNet++ \cite{scannet++} \\
        \includegraphics[width=0.20\textwidth]{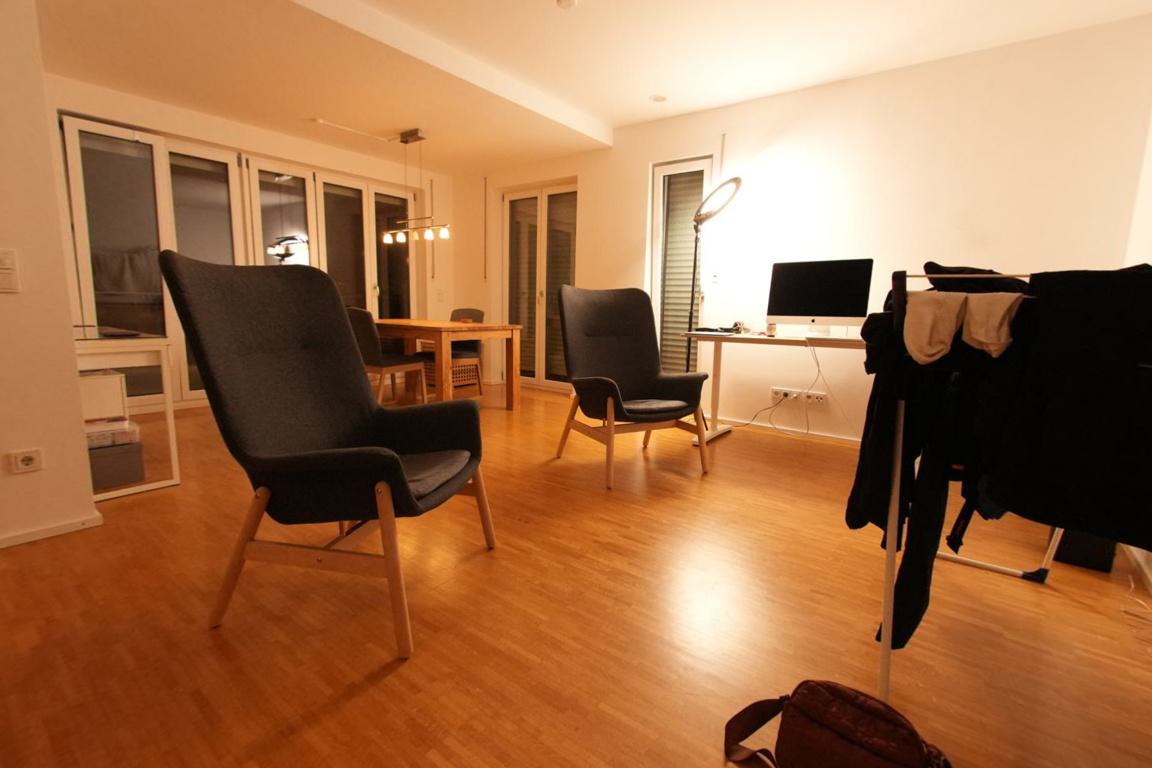}\end{tabular}&
        \begin{tabular}{c}
        \rotatebox{90}{\citet{ComplexInvIndoor}}
        \includegraphics[width=0.205\textwidth]{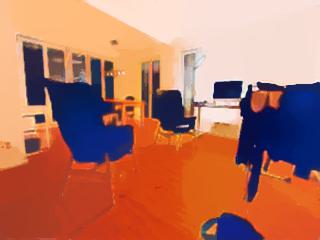}\\
        \rotatebox{90}{\citet{ComplexInvIndoorMC}}
        \includegraphics[width=0.205\textwidth]{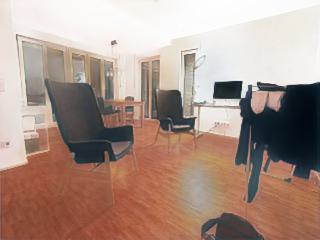}
        \end{tabular}
        &
        \begin{tabular}{c}
        \rotatebox{90}{Mean}
        \includegraphics[width=0.24\textwidth]{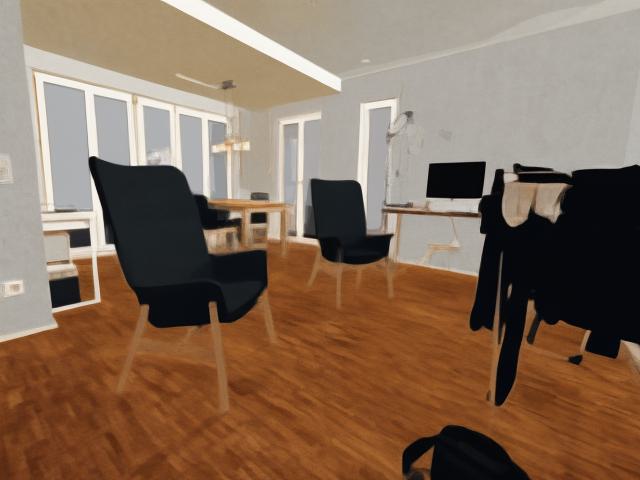}
        \rotatebox{90}{Variance}
        \includegraphics[width=0.26\textwidth]{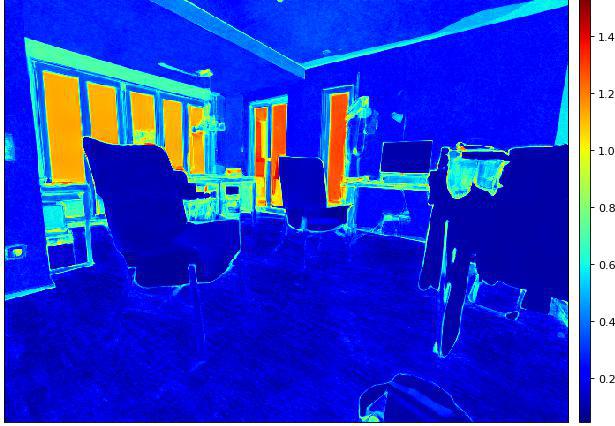}\\
        \rotatebox{90}{Samples}
        \includegraphics[width=0.172\textwidth]{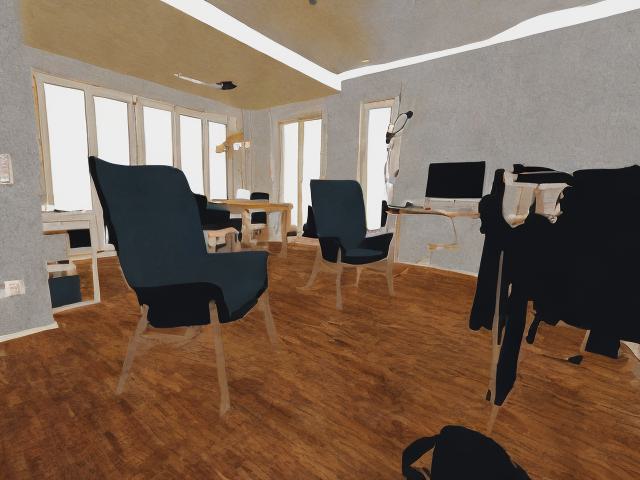}
        \includegraphics[width=0.172\textwidth]{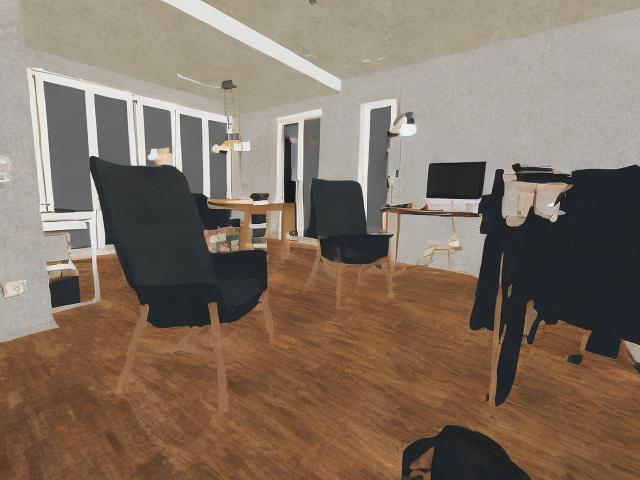}
        \includegraphics[width=0.172\textwidth]{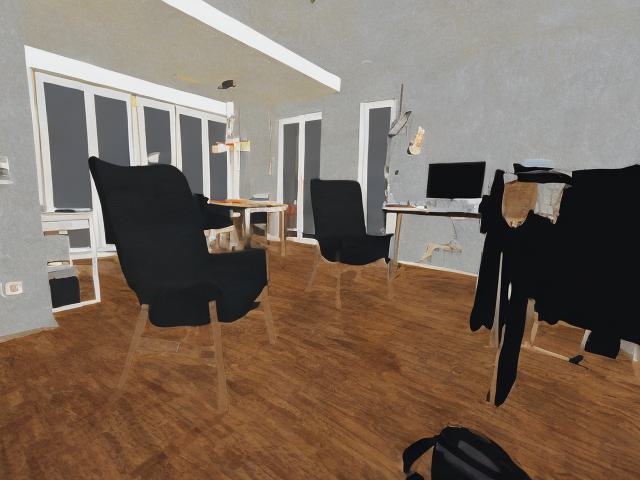}
        \end{tabular} \\

        Input & Baselines & Ours 
  \end{tabular}
  }
  \vspace{-6pt}
  \caption{\textbf{Real-world evaluation.} 
  We qualitatively compare the predicted albedos of our method to the baselines \cite{ComplexInvIndoor, ComplexInvIndoorMC} on the IIW \cite{IIW} and ScanNet++ \cite{scannet++} datasets. 
  Lights and shadows pose a challenge for the previous methods, but our approach gives consistent results on real-world inputs as well. 
  We also visualize the variance of our predictions, showing how the specular, emissive, and small objects have higher uncertainty. 
  See supplementary for more results with roughness and metallic predictions.}
  \label{fig:experiments:real:iiw}
  \vspace{-12pt}
\end{figure*}

\subsection{Synthetic results}
\label{sec:experiment:synthetic}

We evaluate the effectiveness of our material estimation on the test set of the InteriorVerse dataset \cite{ComplexInvIndoorMC}. 
We provide various metrics to evaluate different aspects of our probabilistic model. 

\vspace{3pt}
\noindent\textbf{Fidelity. } 
Our first experiment is about to show that our method predicts more similar albedos to the target albedo distribution.
To assess the distribution similarity, we evaluate the FID score of the predictions in \cref{tab:experiments:synthetic}. 
\cite{ComplexInvIndoorMC} shows great improvement over \cite{ComplexInvIndoor} of $40.60\%$ thanks to the training on the photo-realistic InteriorVerse dataset \cite{ComplexInvIndoorMC} and using an additional perceptual loss helping to utilize semantic information for sharper boundaries.
Our method takes the next step and additionally improves over \cite{ComplexInvIndoorMC} by $44.99\%$.

We show qualitative results in \cref{fig:experiments:synthetic:main}.
\cite{ComplexInvIndoor, ComplexInvIndoorMC} gives reasonable albedo colors overall, but has difficulties with separating strong shadows, lighting, and specular effects from the material and often smoothes along the object boundaries. 
In turn, our approach produces smooth predictions over the same object but also maintains their high-frequency texture details maintaining sharp edges.

\begin{table}
\begin{center}
\resizebox{\columnwidth}{!}{
\begin{tabular}{l|cccc}
  \toprule
   & \multicolumn{4}{c}{InteriorVerse} \\
   & PSNR $\uparrow$ & SSIM $\uparrow$ & LPIPS $\downarrow$ & FID \\
   \midrule 
  IIW \cite{IIW}  & 9.73 {\footnotesize $\pm$ 4.25} & 0.62 {\footnotesize $\pm$ 0.14} & 0.47 {\footnotesize $\pm$ 0.14} & 62.22 \\
  \midrule
  \citet{ComplexInvIndoor}  & 12.31 {\footnotesize $\pm$ 3.32} & 0.68 {\footnotesize $\pm$ 0.11} & 0.52 {\footnotesize $\pm$ 0.11} & 77.79\\
  \citet{ComplexInvIndoorMC}  & 15.92 {\footnotesize $\pm$ 3.93} & 0.78 {\footnotesize $\pm$ 0.09} & 0.34 {\footnotesize $\pm$ 0.09} & 46.21 \\
  Ours - Mean metric & 16.14 {\footnotesize $\pm$ 3.27} & 0.73  {\footnotesize $\pm$ 0.10} & 0.26 {\footnotesize $\pm$ 0.11} & | \\
  Ours - Mean  & 17.42 {\footnotesize $\pm$ 3.08} & \textbf{0.80} {\footnotesize $\pm$ 0.08} & \textbf{0.22} {\footnotesize $\pm$ 0.08} & \textbf{25.42} \\
  Ours - Best  & \textbf{18.43} {\footnotesize $\pm$ 2.97} & 0.77 {\footnotesize $\pm$ 0.09} & 0.26 {\footnotesize $\pm$ 0.11}  & |\\
  \bottomrule
\end{tabular}
}
\end{center}
\caption{\textbf{Quantitative synthetic evaluation.} 
  We quantitatively compare the predicted albedos between our method and the baselines \cite{ComplexInvIndoor, ComplexInvIndoorMC} on synthetic InteriorVerse data \cite{ComplexInvIndoorMC}. 
  Additionally, we report the optimization-based method \cite{IIW}, which was directly tuned on the IIW dataset. 
  We sample $10$ solutions for each view and report the mean of the metrics, the metric of the mean, and the best prediction. 
  (i) Our method has higher fidelity predictions closer to the target albedo distribution, as shown by the FID score. 
  (ii) Our predictions are consistent with the input image, and we can produce predictions close to the GT since the best sample significantly outperforms all the others. 
}
\label{tab:experiments:synthetic}
\end{table}

\begin{table}
\begin{center}
\begin{tabular}{l|ccc}
  \toprule
   FID $\downarrow$ & Albedo & Roughness & Metallic  \\
   \midrule 
  \citet{ComplexInvIndoor}  & 77.79 & 175.73 & - \\
  \citet{ComplexInvIndoorMC}  & 46.21 & 69.92 & 48.71\\
  Ours  & \textbf{25.42} & \textbf{42.02} & \textbf{22.62}\\
  \bottomrule
\end{tabular}
\end{center}
\caption{\textbf{Roughness and metallic evaluation.}
We compare the quality of predicted roughness and metallic maps between our method and the baselines \cite{ComplexInvIndoor, ComplexInvIndoorMC} on the InteriorVerse dataset \cite{ComplexInvIndoorMC}. 
}
\label{tab:experiments:synthetic:brdf}
\vspace{-15pt}
\end{table}

\begin{table}
\begin{center}
\begin{tabular}{l|cc}
  \toprule
   &  \multicolumn{2}{c}{IIW} \\
   &  WHDR $\downarrow$ & PQ $\uparrow$\\
   \midrule 
  IIW \cite{IIW}  & 21.00 {\footnotesize $\pm$ 11.06} & 9.63 \\
  \midrule
  \citet{ComplexInvIndoor}  & 21.99 {\footnotesize $\pm$ 11.06} & 0.46 \\
  \citet{ComplexInvIndoorMC}  & 22.90 {\footnotesize $\pm$ 10.43} & 6.11\\
  Ours - Mean metric & 30.76 {\footnotesize $\pm$ 14.05} & |\\
  Ours - Mean  & 22.02 {\footnotesize $\pm$ 11.99} & \textbf{83.80}\\
  Ours - Best  & \textbf{21.84} {\footnotesize $\pm$ 11.63} & |\\
  \bottomrule
\end{tabular}
\end{center}

\vspace{-6pt}
\caption{\textbf{Quantitative real evaluation.} 
  We quantitatively compare the predicted albedos between our method and the baselines \cite{ComplexInvIndoor, ComplexInvIndoorMC} on real IIW data \cite{IIW}. 
  Additionally, we report the optimization-based method \cite{IIW}, which was directly tuned on the IIW dataset. 
  We sample $10$ solutions for each view and report the mean of the metrics, the metric of the mean, and the best prediction. 
  We reach similar quantitative performance on the real-world IIW dataset; however, our predictions are much sharper and contain more high-frequency details (\cref{fig:experiments:real:iiw,,fig:experiments:real:whdr}) as also assessed by Perceptual Quality in our user-study. 
}
\label{tab:experiments:real}
\vspace{-18pt}
\end{table}

\vspace{3pt}
\noindent\textbf{Consistency. } 
In this section, we show that the predictions are consistent with the input image and it can be a valid solution for the input image. 
To this end, we evaluate in \cref{tab:experiments:synthetic} how close the samples are to the GT albedo. 
We compare the PSNR, SSIM, and LPIPS metrics on the predicted albedos. 
Since appearance decomposition is a highly ambiguous task, a deterministic model \cite{ComplexInvIndoor, ComplexInvIndoorMC} yields just one sample from the solution space. 
Our approach can sample multiple solutions from the solution space.
If we calculate the metric for all the samples independently and average them, we will get slightly better ($+0.22dB$ PSNR) but similar results as \cite{ComplexInvIndoorMC}. 
Taking the mean of the solution space drastically improves the performance ($+1.28dB$ PSNR), leading to the best SSIM. 
We also report the metric of the best samples too, showing that our solution space contains predictions much closer ($+1.01dB$ PSNR) to the ground truth.
However, the ground truth is also just one sample in the solution space; thus, exactly finding it would require exploring this space.  

\vspace{3pt}
\noindent\textbf{Diversity. }
One remaining question about a generative model is whether it produces diverse samples without mode collapse. 
To evaluate this aspect, we sample $100$ material predictions for a single view (top scene of \cref{fig:experiments:synthetic:main}).
We show $4$ predictions in \cref{fig:experiments:synthetic:diversity} and visualize the relative variance of the samples. 
To this end, we normalize the predictions to zero mean and one standard deviation for each channel.
Then, we plot the pixel-wise channel-average standard deviation. 
This variance map can be interpreted as an uncertainty map. 
We can see that large diffuse objects have stable predictions with a small variance. 
However, the prediction of the specular, emissive, or small objects is more uncertain, showing complex ambiguity between lighting and material. 
To quantify this observation for specularity, we investigate the strength of the linear relationship between the albedo variance and metallic properties. 
We have evaluated the Pearson correlation value per image across all pixels for a subset of the test scenes, which ranges between 0.47 and 0.68, underlining our observation. 
Note that perfect correlation should not be expected as specularity is just one source of ambiguity. 
See qualitative comparison in supplemental. 

\vspace{3pt}
\noindent\textbf{Roughness and metallic evaluation.} 
Next to the albedo predictions, we estimate per-pixel BRDF properties as well, such as roughness and metallic. 
We evaluate the fidelity of these predictions in \cref{tab:experiments:synthetic:brdf}. 
\cite{ComplexInvIndoor} does not estimate the metallic properties.
Similar to the advancements observed in albedo predictions, \cite{ComplexInvIndoorMC} demonstrates significant improvement compared to \cite{ComplexInvIndoor}, achieving a remarkable relative enhancement of $60.21\%$ FID in roughness prediction.
Building upon this achievement, our approach surpasses the performance of \cite{ComplexInvIndoorMC} by further improving $39.90\%$ and $53.56\%$ FID in roughness and metallic prediction, respectively.
We provide qualitative results in our supplementary.

\begin{figure}
  \setlength\tabcolsep{1.25pt}
  \centering
  \begin{tabular}{c|ccc}
        \includegraphics[width=0.23\columnwidth]{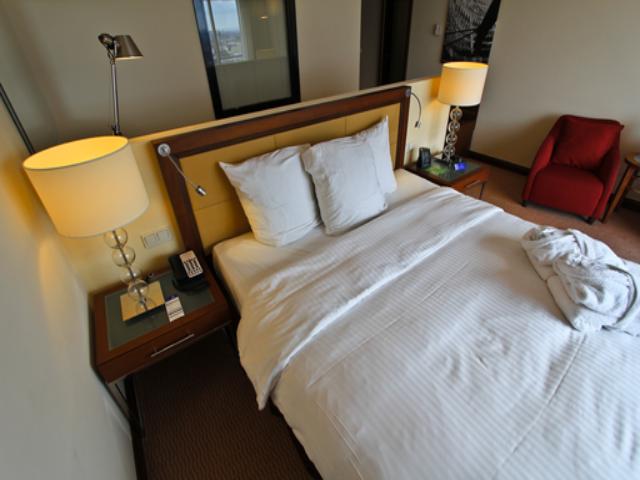}\vspace{-4pt}&
        \includegraphics[width=0.23\columnwidth]{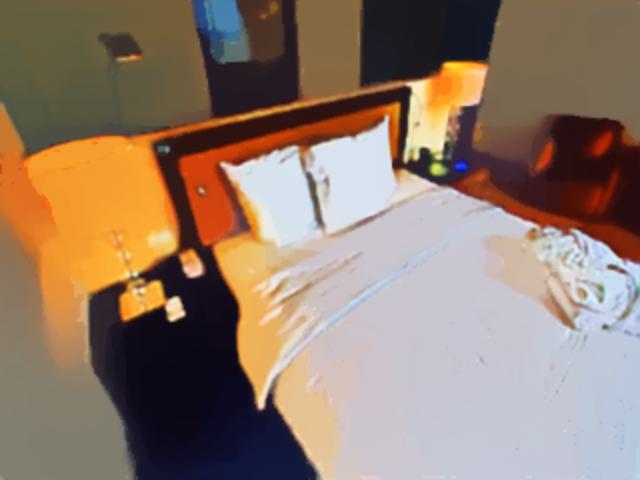}&
        \includegraphics[width=0.23\columnwidth]{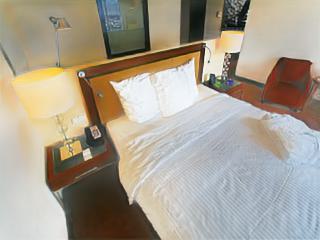}&
        \includegraphics[width=0.23\columnwidth]{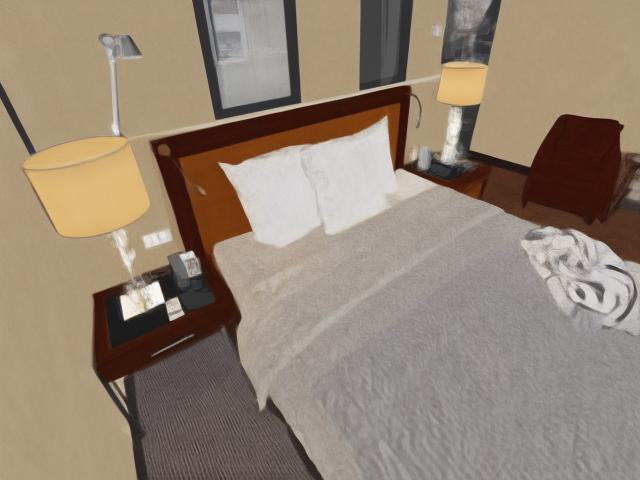}\\
        \small{\hspace{24pt}WHDR} & \small{\hspace{24pt}\textbf{0.14}} & \small{\hspace{24pt}9.51} & \small{\hspace{24pt}3.79} \\
        
        \includegraphics[width=0.23\columnwidth]{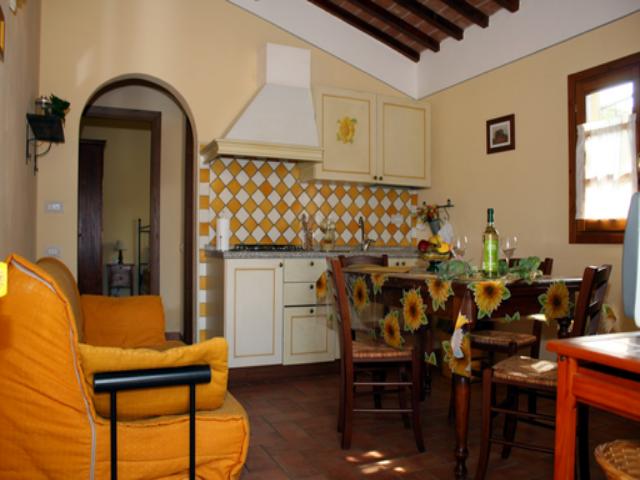}\vspace{-4pt}&
        \includegraphics[width=0.23\columnwidth]{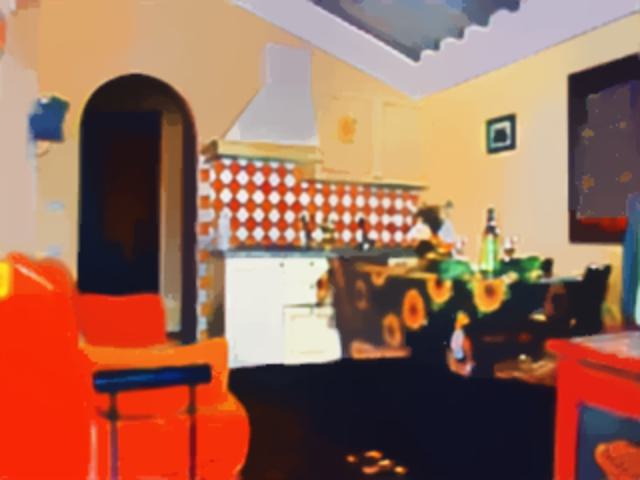}&
        \includegraphics[width=0.23\columnwidth]{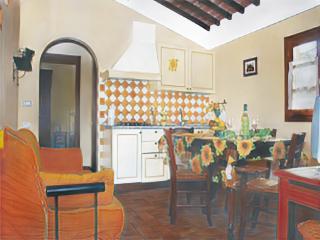}&
        \includegraphics[width=0.23\columnwidth]{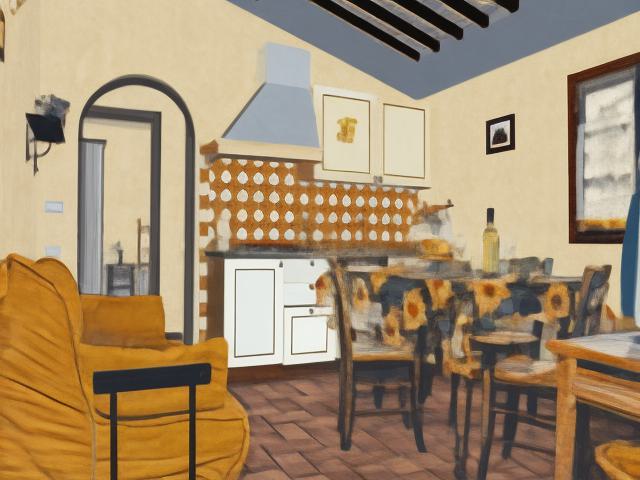}\\
        \small{\hspace{24pt}WHDR} & \small{\hspace{24pt}21.00} & \small{\hspace{24pt}\textbf{17.87}} & \small{\hspace{24pt}31.09} \\

        Input & \citet{ComplexInvIndoor} & \citet{ComplexInvIndoorMC} & Ours 
  \end{tabular}
  \caption{\textbf{WHDR metric imperfection.} 
The WHDR metric tends to favor smoother outcomes,
thus not providing an optimal evaluation of high-frequency details.
Despite our method producing significantly clearer and sharper predictions, the baseline achieves a better score even though their predictions exhibit burned-in lighting and over-smoothed texture. }
\label{fig:experiments:real:whdr}
\vspace*{-15pt}
\end{figure}

\subsection{Real-world results}
\label{sec:experiment:real}
\noindent\textbf{Qualitative evaluation. } 
We evaluate our method on the challenging IIW \cite{IIW} and ScanNet++ \cite{scannet++} datasets. 
\citet{ComplexInvIndoor} proposes an additional bilateral solver for refinement and found that it helps for real-world images.
This solver smoothes the material, making them more consistent overall but for the cost of losing high-frequency details. 
Following \citet{ComplexInvIndoor}, we use the bilateral solver for evaluating their method on real scenes. 

We show qualitative results in \cref{fig:experiments:real:iiw}. 
\cite{ComplexInvIndoor} often produces over-smoothed predictions but with overall correct colors.
\cite{ComplexInvIndoorMC} gives more detailed albedos but still struggles with baked-in shadows, lighting, and specularity, especially close to light sources. 
Our method predicts detailed and well-textured materials over the objects and also for real-world samples. 
Similarly to the synthetic results, small, specular, and emissive objects have higher variance. 
For roughness and metallic results, see supplemental. 

\vspace{3pt}
\noindent\textbf{Quantitative evaluation. } 
We report the WHDR metric \cite{IIW} in \cref{tab:experiments:real}, which compares the predicted albedo to human judgments of relative intensities.
Our method achieves similar performance quantitatively as the baselines. 
However, the WHDR metric assesses predictions sparsely and tends to favor smoother outcomes, thus not providing an optimal evaluation of high-frequency details, as shown in \cref{fig:experiments:real:whdr} and also found in \cite{MAW, DIPR, IIUP}.

As an additional quantitative evaluation, we have conducted a user-study 
 with $27$ participants on assessing the albedo quality. 
We show the albedo prediction of three baselines \cite{ComplexInvIndoor, ComplexInvIndoorMC, IIW} and our method together with the input image for a total of $40$ scenes of the IIW dataset \cite{IIW}. 
The user needs to choose the best prediction. 
To get the users more acquainted with assessing the quality of albedo predictions, we first show them three synthetic images with their ground-truth albedo. 
We report the percentage of being chosen as \textit{Perceptual Quality} in \cref{tab:experiments:real}.

\begin{figure}
  \setlength\tabcolsep{1.25pt}
  \centering
  \begin{tabular}{ccc|c}
        \includegraphics[width=0.23\columnwidth]{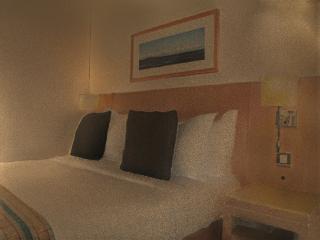}&
        \includegraphics[width=0.23\columnwidth]{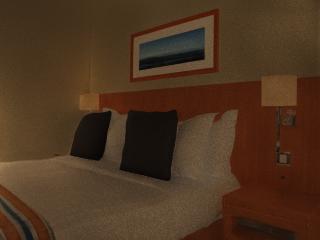}&
        \includegraphics[width=0.23\columnwidth]{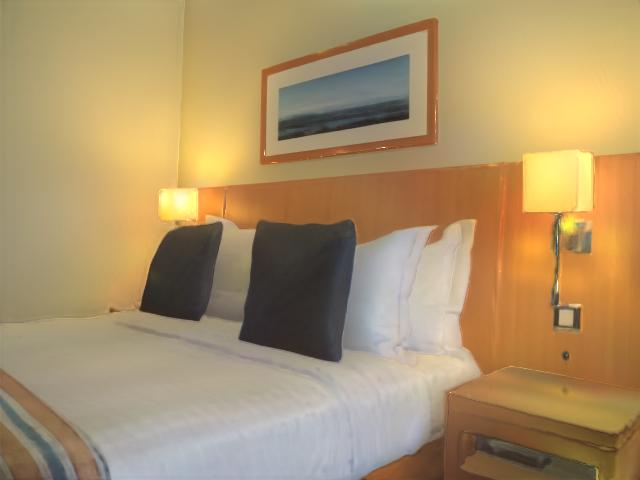}&
        \includegraphics[width=0.23\columnwidth]{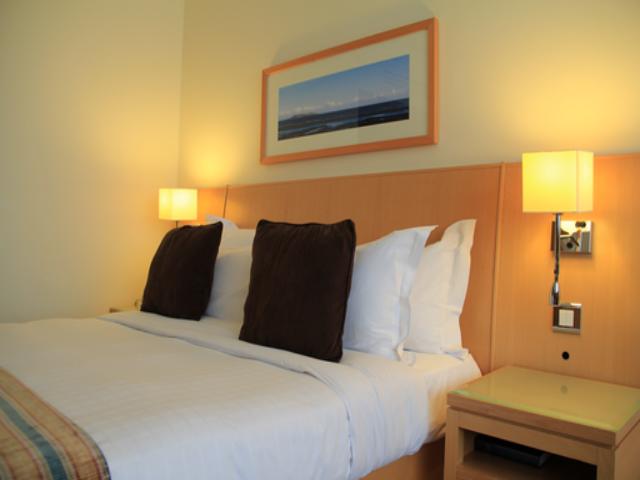}\\
        
        \includegraphics[width=0.23\columnwidth]{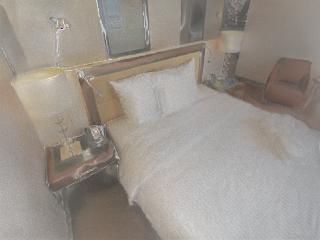}&
        \includegraphics[width=0.23\columnwidth]{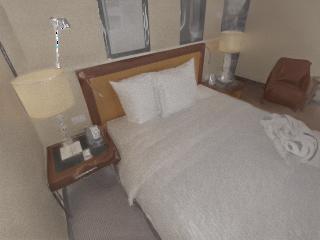}&
        \includegraphics[width=0.23\columnwidth]{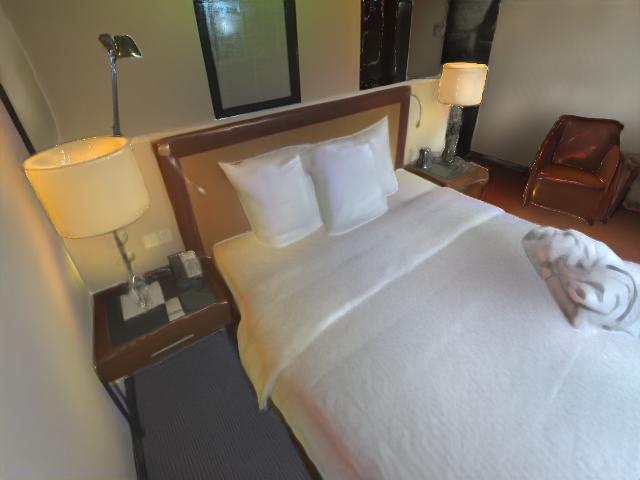}&
        \includegraphics[width=0.23\columnwidth]{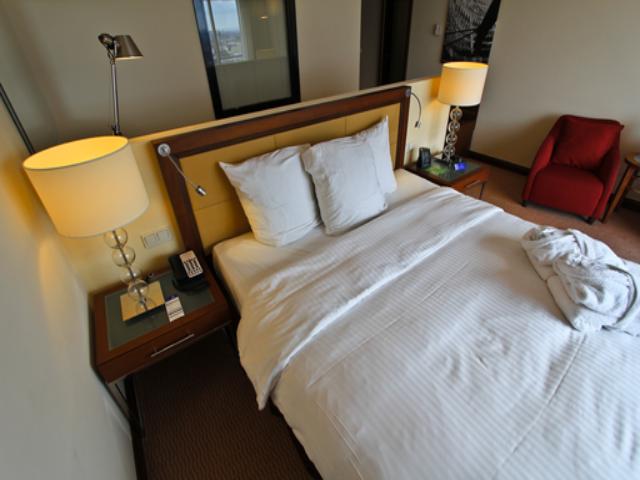}\\
        \citet{ComplexInvIndoorMC} & \cite{ComplexInvIndoorMC} w/ Ours  & Ours Full & GT
  \end{tabular}
  \caption{\textbf{Image reconstruction.}
  We compare the rerendering of \cite{ComplexInvIndoorMC} on real images \cite{IIW} first using their, then our material predictions against our full inverse rendering pipeline. 
  Being trained on synthetic data, \cite{ComplexInvIndoorMC} suffers from domain gap on real-world samples, while our material predictions help to maintain sharp details, enabling to fit controllable detailed lighting. }
  \label{fig:experiments:reconstruction}
  \vspace{-6pt}
\end{figure}

\vspace{3pt}
\noindent\textbf{Image Reconstruction. } 
We evaluate the image reconstruction quality of our method. 
First, we use the lighting and geometry prediction module of \cite{ComplexInvIndoorMC} with our material predictions, which makes the renderings more consistent; however, fails to capture small light sources (\cref{fig:experiments:reconstruction}).
Our lighting optimization reconstructs all light sources faithfully outperforming the baseline with a high margin (\cref{tab:experiments:reconstruction})

\begin{table}
    \begin{center}
        \resizebox{\columnwidth}{!}{
        \begin{tabular}{l|cccc}
            \toprule
            & PSNR $\uparrow$ & SSIM $\uparrow$ & LPIPS $\downarrow$\\
            \midrule 

            \citet{ComplexInvIndoorMC} & 13.54 {\footnotesize $\pm$ 2.22} & 0.51 {\footnotesize $\pm$ 0.12} & 0.43 {\footnotesize $\pm$ 0.08}\\
            \cite{ComplexInvIndoorMC} w/ Ours & 14.07 {\footnotesize $\pm$ 1.91} & 0.55 {\footnotesize $\pm$ 0.11} & 0.38 {\footnotesize $\pm$ 0.08}  \\
            Ours Full & \textbf{21.96}{\footnotesize $\pm$ 3.22} & \textbf{0.70}{\footnotesize $\pm$ 0.10} & \textbf{0.22} {\footnotesize $\pm$ 0.08}  \\
            
            \bottomrule
        \end{tabular}
        }
    \end{center}
    \vspace{-9pt}
    \caption{\textbf{Image reconstruction.}
    We compare the image rerendering of \cite{ComplexInvIndoorMC} using their and our material predictions against our full pipeline with lighting estimation (\cref{fig:method:lighting}). 
    Our material predictions generalize better to real-world images thanks to the diffusion prior and improve the rendering of \cite{ComplexInvIndoorMC}, even though their lighting estimation network is trained with predicted materials of \cite{ComplexInvIndoorMC} and forced to learn to compensate for their discrepancies. 
    Our consistent materials enable us to fit detailed and also controllable lighting, improving the rerendering quality by a high margin. 
    }
    \label{tab:experiments:reconstruction}
    \vspace{-15pt}
\end{table}

\subsection{Ablations}
\label{sec:experiment:ablations}
\begin{figure}
  \setlength\tabcolsep{1.25pt}
  \centering
  \begin{tabular}{c|ccc}
        \includegraphics[width=0.235\columnwidth]{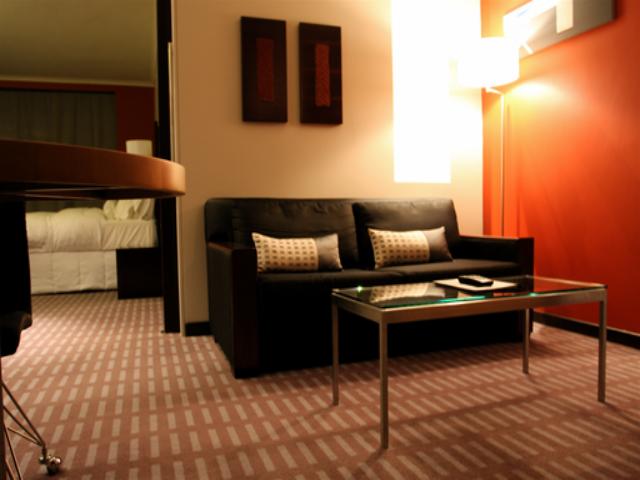}&
        \includegraphics[width=0.235\columnwidth]{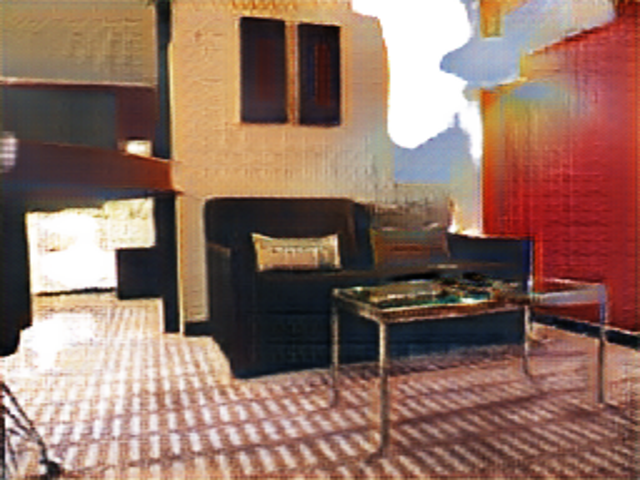}&
        \includegraphics[width=0.235\columnwidth]{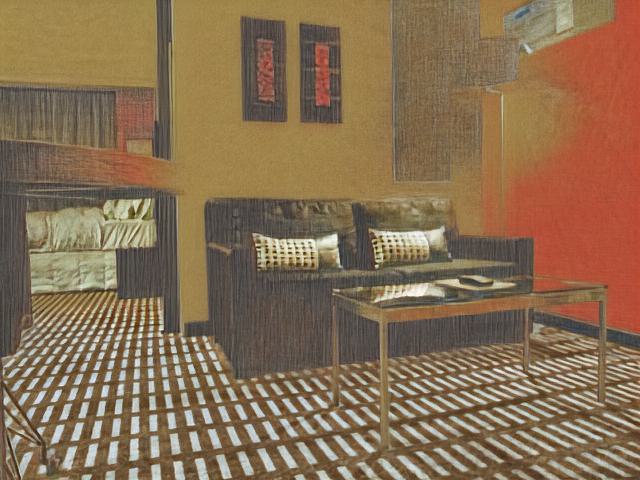}&
        \includegraphics[width=0.235\columnwidth]{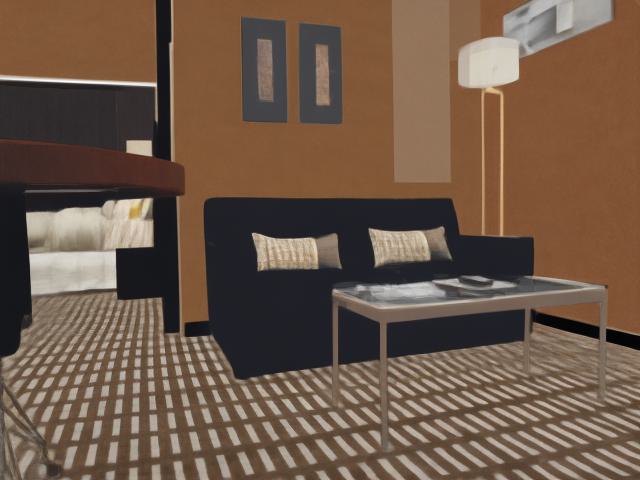}\\
        Input & Pix2Pix \cite{pix2pix} & Ours w/o PT & Ours
  \end{tabular}
  \caption{\textbf{Model ablation.} 
  Leveraging Stable Diffusion's \cite{LDM} real image prior significantly improves our material diffusion. }
  \label{fig:experiments:ablations:pt}
  \vspace{-6pt}
\end{figure}

\noindent\textbf{Effect of pre-training. } 
Our main design choice is the fine-tuning of a pre-trained diffusion model.
We evaluate the albedo predictions on the InteriorVerse dataset \cite{ComplexInvIndoorMC} and on the IIW dataset \cite{IIW}. 
Both qualitatively \cref{fig:experiments:ablations:pt} and quantitatively (\cref{tab:experiments:ablations:pt}), using the pre-trained model, gives significant improvement. 
Notably, pre-training not only aids in reducing the domain gap but also proves beneficial in synthetic scenarios, showing that the prior knowledge obtained from real imagery can be effectively adapted and leveraged for material estimation.
Without pre-training, the model struggles with under and over-saturated regions, but pre-training gives a strong prior over various objects, enabling high-quality material estimation in challenging regions.

\subsection{Limitations and Future Work}
We have shown significant improvement for single-view material estimation by formulating the task probabilistically and utilizing the strong learned prior of a pre-trained diffusion model.
Currently, we train on paired synthetic samples and use the real-world image prior of a pre-trained diffusion model to reduce the domain gap.
Extending this method to weak supervision, even with real-world samples, is a great avenue for future research. 
Furthermore, our current work focuses on material estimation and optimizes for the lighting independently. 
A fused inverse rendering framework that effectively leverages the probabilistic nature of our method for lighting estimation could be beneficial.
Finally, generative models present numerous innovative possibilities that offer greater control over renderings, such as text-guided material editing.

\begin{table}
\begin{center}
\resizebox{\columnwidth}{!}{
\begin{tabular}{l|cccc|c}
  \toprule
   & \multicolumn{4}{c|}{InteriorVerse} & IIW \\
   & PSNR $\uparrow$ & SSIM $\uparrow$ & LPIPS $\downarrow$ & FID $\downarrow$ & WHDR $\downarrow$ \\
   \midrule 
  Pix2Pix \cite{pix2pix} & 13.69 {\footnotesize $\pm$ 2.94} & 0.67 {\footnotesize $\pm$ 0.11} & 0.50 {\footnotesize $\pm$ 0.11} & 84.28 & 36.42 {\footnotesize $\pm$ 12.28} \\
   \midrule 
  Ours w/o PT - Mean & 13.38 {\footnotesize $\pm$ 2.44} & 0.69 {\footnotesize $\pm$ 0.09} & 0.42 {\footnotesize $\pm$ 0.10} & 113.31 & 35.60 {\footnotesize $\pm$ 12.94}\\
  Ours w/o PT - Best  & 13.85 {\footnotesize $\pm$ 2.46} & 0.61 {\footnotesize $\pm$ 0.12} & 0.61 {\footnotesize $\pm$ 0.14} &  & 36.34 {\footnotesize $\pm$ 11.78}\\
  Ours - Mean  & 17.42 {\footnotesize $\pm$ 3.08} & \textbf{0.80} {\footnotesize $\pm$ 0.08} & \textbf{0.22} {\footnotesize $\pm$ 0.08} & \textbf{25.42} & 22.02 {\footnotesize $\pm$ 11.99}\\
  Ours - Best  & \textbf{18.43} {\footnotesize $\pm$ 2.97} & 0.77 {\footnotesize $\pm$ 0.09} & 0.26 {\footnotesize $\pm$ 0.11}  &  & \textbf{21.84} {\footnotesize $\pm$ 11.63}\\
  \bottomrule
\end{tabular}
}
\end{center}
\vspace{-9pt}
\caption{\textbf{Model ablation.}
Using a pre-trained diffusion model brings substantial improvement. 
Using the GAN-based pix2pix \cite{pix2pix} performs slightly weaker as diffusion trained from scratch. 
Stable Diffusion's \cite{LDM} prior of real imagery not only reduces the domain gap but can also be adapted for material estimation, resulting in improved synthetic results as well.
}
\label{tab:experiments:ablations:pt}
\vspace{-15pt}
\end{table}

\section{Conclusion}
We have introduced Intrinsic Image Diffusion, a generative model for material estimation.
Rather than predicting a single solution, we propose probabilistically formulating the highly ambiguous appearance decomposition. 
We have shown that leveraging the strong image prior of recent diffusion models improves the material prediction quality and reduces the domain gap. 
Our material predictions are significantly cleaner and more consistent than the ones of prior works, enabling direct optimization-based emissive lighting fitting.
We believe our work can pave the way for novel approaches to tackling the challenging task of the ill-posed appearance decomposition by applying conditional generative models and utilizing image prior.

\vspace{3pt}
\noindent\textbf{Acknowledgements.}
This work was supported by the ERC Starting Grant Scan2CAD (804724), the German Research Foundation (DFG) Grant ``Making Machine Learning on Static and Dynamic 3D Data Practical'', and the German Research Foundation (DFG) Research Unit ``Learning and Simulation in Visual Computing''. Vincent Sitzmann was supported by the National Science Foundation under Grant No. 2211259, and by the Singapore DSTA under DST00OECI20300823 (New Representations for Vision).

{
    \small
    \bibliographystyle{ieeenat_fullname}
    \bibliography{bibliography}

\begin{thebibliography}{50}
\providecommand{\natexlab}[1]{#1}
\providecommand{\url}[1]{\texttt{#1}}
\expandafter\ifx\csname urlstyle\endcsname\relax
  \providecommand{\doi}[1]{doi: #1}\else
  \providecommand{\doi}{doi: \begingroup \urlstyle{rm}\Url}\fi

\bibitem[Azinovic et~al.(2019)Azinovic, Li, Kaplanyan, and Nie{\ss}ner]{IPT}
Dejan Azinovic, Tzu{-}Mao Li, Anton Kaplanyan, and Matthias Nie{\ss}ner.
\newblock Inverse path tracing for joint material and lighting estimation.
\newblock In \emph{{IEEE} Conference on Computer Vision and Pattern Recognition, {CVPR} 2019, Long Beach, CA, USA, June 16-20, 2019}, pages 2447--2456. Computer Vision Foundation / {IEEE}, 2019.

\bibitem[Barron and Malik(2013)]{DBLP:conf/cvpr/BarronM13}
Jonathan~T. Barron and Jitendra Malik.
\newblock Intrinsic scene properties from a single {RGB-D} image.
\newblock In \emph{2013 {IEEE} Conference on Computer Vision and Pattern Recognition, Portland, OR, USA, June 23-28, 2013}, pages 17--24. {IEEE} Computer Society, 2013.

\bibitem[Barron and Malik(2015)]{ShapeFromShading}
Jonathan~T. Barron and Jitendra Malik.
\newblock Shape, illumination, and reflectance from shading.
\newblock \emph{{IEEE} Trans. Pattern Anal. Mach. Intell.}, 37\penalty0 (8):\penalty0 1670--1687, 2015.

\bibitem[Bell et~al.(2014)Bell, Bala, and Snavely]{IIW}
Sean Bell, Kavita Bala, and Noah Snavely.
\newblock Intrinsic images in the wild.
\newblock \emph{{ACM} Trans. Graph.}, 33\penalty0 (4):\penalty0 159:1--159:12, 2014.

\bibitem[Bhattad and Forsyth(2023)]{bhattad2023StylitGAN}
Anand Bhattad and D.A. Forsyth.
\newblock Stylitgan: Prompting stylegan to generate new illumination conditions.
\newblock In \emph{arXiv}, 2023.

\bibitem[Bhattad and Forsyth(2022)]{DIPR}
Anand Bhattad and David~A. Forsyth.
\newblock Cut-and-paste object insertion by enabling deep image prior for reshading.
\newblock In \emph{International Conference on 3D Vision, 3DV 2022, Prague, Czech Republic, September 12-16, 2022}, pages 332--341. {IEEE}, 2022.

\bibitem[Bhattad et~al.(2023)Bhattad, McKee, Hoiem, and Forsyth]{DBLP:conf/nips/BhattadMHF23}
Anand Bhattad, Daniel McKee, Derek Hoiem, and David~A. Forsyth.
\newblock Stylegan knows normal, depth, albedo, and more.
\newblock In \emph{Advances in Neural Information Processing Systems 36: Annual Conference on Neural Information Processing Systems 2023, NeurIPS 2023, New Orleans, LA, USA, December 10 - 16, 2023}, 2023.

\bibitem[Chen and Koltun(2013)]{DBLP:conf/iccv/ChenK13}
Qifeng Chen and Vladlen Koltun.
\newblock A simple model for intrinsic image decomposition with depth cues.
\newblock In \emph{{IEEE} International Conference on Computer Vision, {ICCV} 2013, Sydney, Australia, December 1-8, 2013}, pages 241--248. {IEEE} Computer Society, 2013.

\bibitem[Choi et~al.(2023)Choi, Lee, Park, Jung, Kim, and Cho]{MAIR}
Junyong Choi, SeokYeong Lee, Haesol Park, Seung{-}Won Jung, Ig{-}Jae Kim, and Junghyun Cho.
\newblock {MAIR:} multi-view attention inverse rendering with 3d spatially-varying lighting estimation.
\newblock \emph{CoRR}, abs/2303.12368, 2023.

\bibitem[Du et~al.(2023)Du, Kolkin, Shakhnarovich, and Bhattad]{IntrinsicLora}
Xiaodan Du, Nicholas~I. Kolkin, Greg Shakhnarovich, and Anand Bhattad.
\newblock Generative models: What do they know? do they know things? let's find out!
\newblock \emph{CoRR}, abs/2311.17137, 2023.

\bibitem[Eftekhar et~al.(2021)Eftekhar, Sax, Malik, and Zamir]{OmniData}
Ainaz Eftekhar, Alexander Sax, Jitendra Malik, and Amir Zamir.
\newblock Omnidata: {A} scalable pipeline for making multi-task mid-level vision datasets from 3d scans.
\newblock In \emph{2021 {IEEE/CVF} International Conference on Computer Vision, {ICCV} 2021, Montreal, QC, Canada, October 10-17, 2021}, pages 10766--10776. {IEEE}, 2021.

\bibitem[Forsyth and Rock(2022)]{IIUP}
David~A. Forsyth and Jason~J. Rock.
\newblock Intrinsic image decomposition using paradigms.
\newblock \emph{{IEEE} Trans. Pattern Anal. Mach. Intell.}, 44\penalty0 (11):\penalty0 7624--7637, 2022.

\bibitem[Garces et~al.(2012)Garces, Munoz, Lopez-Moreno, and Gutierrez]{ChromacityClustering}
Elena Garces, Adolfo Munoz, Jorge Lopez-Moreno, and Diego Gutierrez.
\newblock Intrinsic images by clustering.
\newblock In \emph{Computer graphics forum}, pages 1415--1424. Wiley Online Library, 2012.

\bibitem[Grosse et~al.(2009)Grosse, Johnson, Adelson, and Freeman]{MITII}
Roger~B. Grosse, Micah~K. Johnson, Edward~H. Adelson, and William~T. Freeman.
\newblock Ground truth dataset and baseline evaluations for intrinsic image algorithms.
\newblock In \emph{{IEEE} 12th International Conference on Computer Vision, {ICCV} 2009, Kyoto, Japan, September 27 - October 4, 2009}, pages 2335--2342. {IEEE} Computer Society, 2009.

\bibitem[Ho et~al.(2020)Ho, Jain, and Abbeel]{DDPM}
Jonathan Ho, Ajay Jain, and Pieter Abbeel.
\newblock Denoising diffusion probabilistic models.
\newblock In \emph{Advances in Neural Information Processing Systems 33: Annual Conference on Neural Information Processing Systems 2020, NeurIPS 2020, December 6-12, 2020, virtual}, 2020.

\bibitem[Ilharco et~al.(2021)Ilharco, Wortsman, Wightman, Gordon, Carlini, Taori, Dave, Shankar, Namkoong, Miller, Hajishirzi, Farhadi, and Schmidt]{OpenCLIP}
Gabriel Ilharco, Mitchell Wortsman, Ross Wightman, Cade Gordon, Nicholas Carlini, Rohan Taori, Achal Dave, Vaishaal Shankar, Hongseok Namkoong, John Miller, Hannaneh Hajishirzi, Ali Farhadi, and Ludwig Schmidt.
\newblock Openclip, 2021.
\newblock If you use this software, please cite it as below.

\bibitem[Isola et~al.(2017)Isola, Zhu, Zhou, and Efros]{pix2pix}
Phillip Isola, Jun-Yan Zhu, Tinghui Zhou, and Alexei~A Efros.
\newblock Image-to-image translation with conditional adversarial networks.
\newblock \emph{CVPR}, 2017.

\bibitem[Jin et~al.(2023)Jin, Li, Yang, and Tan]{jin2023estimating}
Yeying Jin, Ruoteng Li, Wenhan Yang, and Robby~T Tan.
\newblock Estimating reflectance layer from a single image: Integrating reflectance guidance and shadow/specular aware learning.
\newblock In \emph{Proceedings of the AAAI Conference on Artificial Intelligence}, pages 1069--1077, 2023.

\bibitem[Kingma and Ba(2015)]{Adam}
Diederik~P. Kingma and Jimmy Ba.
\newblock Adam: {A} method for stochastic optimization.
\newblock In \emph{3rd International Conference on Learning Representations, {ICLR} 2015, San Diego, CA, USA, May 7-9, 2015, Conference Track Proceedings}, 2015.

\bibitem[Kingma et~al.(2021)Kingma, Salimans, Poole, and Ho]{DBLP:journals/corr/abs-2107-00630}
Diederik~P. Kingma, Tim Salimans, Ben Poole, and Jonathan Ho.
\newblock Variational diffusion models.
\newblock \emph{CoRR}, abs/2107.00630, 2021.

\bibitem[Kovacs et~al.(2017)Kovacs, Bell, Snavely, and Bala]{SAW}
Balazs Kovacs, Sean Bell, Noah Snavely, and Kavita Bala.
\newblock Shading annotations in the wild.
\newblock In \emph{2017 {IEEE} Conference on Computer Vision and Pattern Recognition, {CVPR} 2017, Honolulu, HI, USA, July 21-26, 2017}, pages 850--859. {IEEE} Computer Society, 2017.

\bibitem[Land and McCann(1971)]{Retinex}
Edwin Land and John McCann.
\newblock Lightness and retinex theory.
\newblock \emph{Journal of the Optical Society of America}, 61:\penalty0 1--11, 1971.

\bibitem[Lee et~al.(2023)Lee, Tseng, Lee, and Yang]{DMP}
Hsin{-}Ying Lee, Hung{-}Yu Tseng, Hsin{-}Ying Lee, and Ming{-}Hsuan Yang.
\newblock Exploiting diffusion prior for generalizable pixel-level semantic prediction.
\newblock \emph{CoRR}, abs/2311.18832, 2023.

\bibitem[Li and Snavely(2018{\natexlab{a}})]{CGIntrinsics}
Zhengqi Li and Noah Snavely.
\newblock Cgintrinsics: Better intrinsic image decomposition through physically-based rendering.
\newblock In \emph{Computer Vision - {ECCV} 2018 - 15th European Conference, Munich, Germany, September 8-14, 2018, Proceedings, Part {III}}, pages 381--399. Springer, 2018{\natexlab{a}}.

\bibitem[Li and Snavely(2018{\natexlab{b}})]{DBLP:conf/cvpr/LiS18a}
Zhengqi Li and Noah Snavely.
\newblock Learning intrinsic image decomposition from watching the world.
\newblock In \emph{2018 {IEEE} Conference on Computer Vision and Pattern Recognition, {CVPR} 2018, Salt Lake City, UT, USA, June 18-22, 2018}, pages 9039--9048. Computer Vision Foundation / {IEEE} Computer Society, 2018{\natexlab{b}}.

\bibitem[Li et~al.(2018)Li, Xu, Ramamoorthi, Sunkavalli, and Chandraker]{DBLP:journals/tog/LiXRSC18}
Zhengqin Li, Zexiang Xu, Ravi Ramamoorthi, Kalyan Sunkavalli, and Manmohan Chandraker.
\newblock Learning to reconstruct shape and spatially-varying reflectance from a single image.
\newblock \emph{{ACM} Trans. Graph.}, 37\penalty0 (6):\penalty0 269, 2018.

\bibitem[Li et~al.(2020)Li, Shafiei, Ramamoorthi, Sunkavalli, and Chandraker]{ComplexInvIndoor}
Zhengqin Li, Mohammad Shafiei, Ravi Ramamoorthi, Kalyan Sunkavalli, and Manmohan Chandraker.
\newblock Inverse rendering for complex indoor scenes: Shape, spatially-varying lighting and {SVBRDF} from a single image.
\newblock In \emph{2020 {IEEE/CVF} Conference on Computer Vision and Pattern Recognition, {CVPR} 2020, Seattle, WA, USA, June 13-19, 2020}, pages 2472--2481. Computer Vision Foundation / {IEEE}, 2020.

\bibitem[Li et~al.(2022)Li, Shi, Bi, Zhu, Sunkavalli, Hasan, Xu, Ramamoorthi, and Chandraker]{PBLightEditing}
Zhengqin Li, Jia Shi, Sai Bi, Rui Zhu, Kalyan Sunkavalli, Milos Hasan, Zexiang Xu, Ravi Ramamoorthi, and Manmohan Chandraker.
\newblock Physically-based editing of indoor scene lighting from a single image.
\newblock In \emph{Computer Vision - {ECCV} 2022 - 17th European Conference, Tel Aviv, Israel, October 23-27, 2022, Proceedings, Part {VI}}, pages 555--572. Springer, 2022.

\bibitem[Liu et~al.(2023)Liu, Wu, Hoorick, Tokmakov, Zakharov, and Vondrick]{Zero123D}
Ruoshi Liu, Rundi Wu, Basile~Van Hoorick, Pavel Tokmakov, Sergey Zakharov, and Carl Vondrick.
\newblock Zero-1-to-3: Zero-shot one image to 3d object.
\newblock \emph{CoRR}, abs/2303.11328, 2023.

\bibitem[Loshchilov and Hutter(2019)]{AdamW}
Ilya Loshchilov and Frank Hutter.
\newblock Decoupled weight decay regularization.
\newblock In \emph{7th International Conference on Learning Representations, {ICLR} 2019, New Orleans, LA, USA, May 6-9, 2019}. OpenReview.net, 2019.

\bibitem[Nimier{-}David et~al.(2021)Nimier{-}David, Dong, Jakob, and Kaplanyan]{Nimier}
Merlin Nimier{-}David, Zhao Dong, Wenzel Jakob, and Anton Kaplanyan.
\newblock Material and lighting reconstruction for complex indoor scenes with texture-space differentiable rendering.
\newblock In \emph{32nd Eurographics Symposium on Rendering, {EGSR} 2021 - Digital Library Only Track, Saarbr{\"{u}}cken, Germany, June 29 - July 2, 2021}, pages 73--84. Eurographics Association, 2021.

\bibitem[Philip et~al.(2021)Philip, Morgenthaler, Gharbi, and Drettakis]{INR}
Julien Philip, S{\'{e}}bastien Morgenthaler, Micha{\"{e}}l Gharbi, and George Drettakis.
\newblock Free-viewpoint indoor neural relighting from multi-view stereo.
\newblock \emph{{ACM} Trans. Graph.}, 40\penalty0 (5):\penalty0 194:1--194:18, 2021.

\bibitem[Po et~al.(2023)Po, Yifan, Golyanik, Aberman, Barron, Bermano, Chan, Dekel, Holynski, Kanazawa, Liu, Liu, Mildenhall, Nie{\ss}ner, Ommer, Theobalt, Wonka, and Wetzstein]{DiffusionSTAR}
Ryan Po, Wang Yifan, Vladislav Golyanik, Kfir Aberman, Jonathan~T. Barron, Amit~H. Bermano, Eric~Ryan Chan, Tali Dekel, Aleksander Holynski, Angjoo Kanazawa, C.~Karen Liu, Lingjie Liu, Ben Mildenhall, Matthias Nie{\ss}ner, Bj{\"{o}}rn Ommer, Christian Theobalt, Peter Wonka, and Gordon Wetzstein.
\newblock State of the art on diffusion models for visual computing.
\newblock \emph{CoRR}, abs/2310.07204, 2023.

\bibitem[Radford et~al.(2021)Radford, Kim, Hallacy, Ramesh, Goh, Agarwal, Sastry, Askell, Mishkin, Clark, Krueger, and Sutskever]{CLIP}
Alec Radford, Jong~Wook Kim, Chris Hallacy, A. Ramesh, Gabriel Goh, Sandhini Agarwal, Girish Sastry, Amanda Askell, Pamela Mishkin, Jack Clark, Gretchen Krueger, and Ilya Sutskever.
\newblock Learning transferable visual models from natural language supervision.
\newblock In \emph{ICML}, 2021.

\bibitem[Rodr{\'{\i}}guez{-}Pardo et~al.(2023)Rodr{\'{\i}}guez{-}Pardo, Dominguez{-}Elvira, Pascual{-}Hern{\'{a}}ndez, and Garces]{UMat}
Carlos Rodr{\'{\i}}guez{-}Pardo, Henar Dominguez{-}Elvira, David Pascual{-}Hern{\'{a}}ndez, and Elena Garces.
\newblock Umat: Uncertainty-aware single image high resolution material capture.
\newblock In \emph{{IEEE/CVF} Conference on Computer Vision and Pattern Recognition, {CVPR} 2023, Vancouver, BC, Canada, June 17-24, 2023}, pages 5764--5774. {IEEE}, 2023.

\bibitem[Rombach et~al.(2021)Rombach, Blattmann, Lorenz, Esser, and Ommer]{LDM}
Robin Rombach, Andreas Blattmann, Dominik Lorenz, Patrick Esser, and Bj{\"{o}}rn Ommer.
\newblock High-resolution image synthesis with latent diffusion models.
\newblock \emph{CoRR}, abs/2112.10752, 2021.

\bibitem[Ronneberger et~al.(2015)Ronneberger, Fischer, and Brox]{UNet}
Olaf Ronneberger, Philipp Fischer, and Thomas Brox.
\newblock U-net: Convolutional networks for biomedical image segmentation.
\newblock In \emph{Medical Image Computing and Computer-Assisted Intervention - {MICCAI} 2015 - 18th International Conference Munich, Germany, October 5 - 9, 2015, Proceedings, Part {III}}, pages 234--241. Springer, 2015.

\bibitem[Schuhmann et~al.(2022)Schuhmann, Beaumont, Vencu, Gordon, Wightman, Cherti, Coombes, Katta, Mullis, Wortsman, Schramowski, Kundurthy, Crowson, Schmidt, Kaczmarczyk, and Jitsev]{LAION5B}
Christoph Schuhmann, Romain Beaumont, Richard Vencu, Cade Gordon, Ross Wightman, Mehdi Cherti, Theo Coombes, Aarush Katta, Clayton Mullis, Mitchell Wortsman, Patrick Schramowski, Srivatsa Kundurthy, Katherine Crowson, Ludwig Schmidt, Robert Kaczmarczyk, and Jenia Jitsev.
\newblock {LAION-5B:} an open large-scale dataset for training next generation image-text models.
\newblock In \emph{NeurIPS}, 2022.

\bibitem[Sohl{-}Dickstein et~al.(2015)Sohl{-}Dickstein, Weiss, Maheswaranathan, and Ganguli]{DBLP:conf/icml/Sohl-DicksteinW15}
Jascha Sohl{-}Dickstein, Eric~A. Weiss, Niru Maheswaranathan, and Surya Ganguli.
\newblock Deep unsupervised learning using nonequilibrium thermodynamics.
\newblock In \emph{Proceedings of the 32nd International Conference on Machine Learning, {ICML} 2015, Lille, France, 6-11 July 2015}, pages 2256--2265. JMLR.org, 2015.

\bibitem[Song et~al.(2020{\natexlab{a}})Song, Meng, and Ermon]{DDIM}
Jiaming Song, Chenlin Meng, and Stefano Ermon.
\newblock Denoising diffusion implicit models.
\newblock \emph{arXiv:2010.02502}, 2020{\natexlab{a}}.

\bibitem[Song et~al.(2020{\natexlab{b}})Song, Sohl{-}Dickstein, Kingma, Kumar, Ermon, and Poole]{DBLP:journals/corr/abs-2011-13456}
Yang Song, Jascha Sohl{-}Dickstein, Diederik~P. Kingma, Abhishek Kumar, Stefano Ermon, and Ben Poole.
\newblock Score-based generative modeling through stochastic differential equations.
\newblock \emph{CoRR}, abs/2011.13456, 2020{\natexlab{b}}.

\bibitem[Walter et~al.(2007)Walter, Marschner, Li, and Torrance]{GGX}
Bruce Walter, Stephen~R. Marschner, Hongsong Li, and Kenneth~E. Torrance.
\newblock Microfacet models for refraction through rough surfaces.
\newblock In \emph{Proceedings of the Eurographics Symposium on Rendering Techniques, Grenoble, France, 2007}, pages 195--206. Eurographics Association, 2007.

\bibitem[Wang et~al.(2021)Wang, Philion, Fidler, and Kautz]{3DSVLIndoor}
Zian Wang, Jonah Philion, Sanja Fidler, and Jan Kautz.
\newblock Learning indoor inverse rendering with 3d spatially-varying lighting.
\newblock In \emph{2021 {IEEE/CVF} International Conference on Computer Vision, {ICCV} 2021, Montreal, QC, Canada, October 10-17, 2021}, pages 12518--12527. {IEEE}, 2021.

\bibitem[Wu et~al.(2023)Wu, Chowdhury, Shanmugaraja, Jacobs, and Sengupta]{MAW}
Jiaye Wu, Sanjoy Chowdhury, Hariharmano Shanmugaraja, David Jacobs, and Soumyadip Sengupta.
\newblock Measured albedo in the wild: Filling the gap in intrinsics evaluation.
\newblock \emph{COPR}, 2023.

\bibitem[Yang et~al.(2022)Yang, Zhang, Song, Hong, Xu, Zhao, Shao, Zhang, Yang, and Cui]{DiffusionSurvey}
Ling Yang, Zhilong Zhang, Yang Song, Shenda Hong, Runsheng Xu, Yue Zhao, Yingxia Shao, Wentao Zhang, Ming{-}Hsuan Yang, and Bin Cui.
\newblock Diffusion models: {A} comprehensive survey of methods and applications.
\newblock \emph{CoRR}, abs/2209.00796, 2022.

\bibitem[Yeshwanth et~al.(2023)Yeshwanth, Liu, Nie{\ss}ner, and Dai]{scannet++}
Chandan Yeshwanth, Yueh-Cheng Liu, Matthias Nie{\ss}ner, and Angela Dai.
\newblock Scannet++: A high-fidelity dataset of 3d indoor scenes.
\newblock In \emph{Proceedings of the International Conference on Computer Vision ({ICCV})}, 2023.

\bibitem[Zhang and Agrawala(2023)]{ControlNet}
Lvmin Zhang and Maneesh Agrawala.
\newblock Adding conditional control to text-to-image diffusion models.
\newblock \emph{CoRR}, abs/2302.05543, 2023.

\bibitem[Zhu et~al.(2022{\natexlab{a}})Zhu, Luan, Huo, Lin, Zhong, Xi, Wang, Bao, Zheng, and Tang]{ComplexInvIndoorMC}
Jingsen Zhu, Fujun Luan, Yuchi Huo, Zihao Lin, Zhihua Zhong, Dianbing Xi, Rui Wang, Hujun Bao, Jiaxiang Zheng, and Rui Tang.
\newblock Learning-based inverse rendering of complex indoor scenes with differentiable monte carlo raytracing.
\newblock In \emph{{SIGGRAPH} Asia 2022 Conference Papers, {SA} 2022, Daegu, Republic of Korea, December 6-9, 2022}, pages 6:1--6:8. {ACM}, 2022{\natexlab{a}}.

\bibitem[Zhu et~al.(2023)Zhu, Huo, Ye, Luan, Li, Xi, Wang, Tang, Hua, Bao, and Wang]{I2SDF}
Jingsen Zhu, Yuchi Huo, Qi Ye, Fujun Luan, Jifan Li, Dianbing Xi, Lisha Wang, Rui Tang, Wei Hua, Hujun Bao, and Rui Wang.
\newblock I\({}^{\mbox{2}}\)-sdf: Intrinsic indoor scene reconstruction and editing via raytracing in neural sdfs.
\newblock \emph{CoRR}, abs/2303.07634, 2023.

\bibitem[Zhu et~al.(2022{\natexlab{b}})Zhu, Li, Matai, Porikli, and Chandraker]{IRISFormer}
Rui Zhu, Zhengqin Li, Janarbek Matai, Fatih Porikli, and Manmohan Chandraker.
\newblock Irisformer: Dense vision transformers for single-image inverse rendering in indoor scenes.
\newblock In \emph{{IEEE/CVF} Conference on Computer Vision and Pattern Recognition, {CVPR} 2022, New Orleans, LA, USA, June 18-24, 2022}, pages 2812--2821. {IEEE}, 2022{\natexlab{b}}.

\end{thebibliography}
}

\maketitlesupplementary
\appendix
\setcounter{page}{1}

In this supplementary material, first, we show application results (\cref{sec:supp:applications}). 
Then, we give additional details on our method (\cref{sec:supp:method}) and on the experimental setting (\cref{sec:supp:baselines}).
We show more qualitative results on the full material prediction, including the albedo, roughness, and metallic properties on synthetic and real data in \cref{sec:supp:results}.
Finally, we present an additional ablation (\cref{sec:supp:ablation}).

\section{Applications}
\label{sec:supp:applications}
Our predicted materials and optimized lighting enable intrinsic image editing, i.e., changing solely specific aspects of the image, such as only the materials or lighting. 

\subsection{Material Editing}
\label{sec:supp:applications:material_editing}
Our sharp material predictions have smooth but sharp features for every object with minimal or no baked-in lighting, enabling simple image editing in the material space. 
An example editing is shown in \cref{fig:supp:applications}, where we change the wall color from beige to cyan. 
Note how the reflections on the wall turn greenish since the lamp emission mostly contains red and green components, but the wall reflects green and blue the most. 

\subsection{Lighting Editing}
\label{sec:supp:applications:lighting_editing}
\begin{figure}[t]
  \centering
  \setlength\tabcolsep{1.25pt}
  \begin{tabular}{ccc}
        \includegraphics[width=0.32\columnwidth]{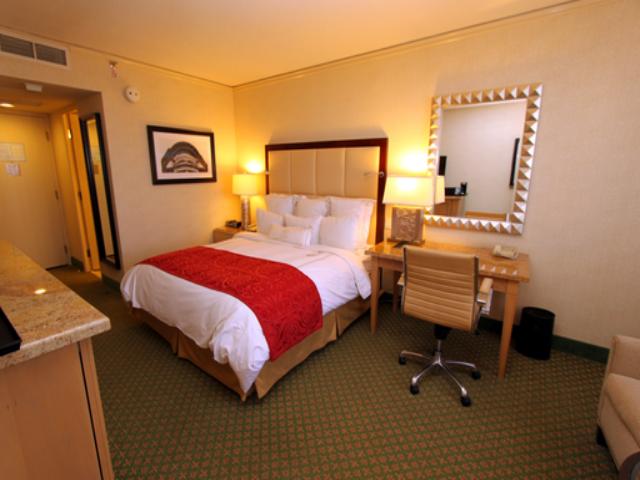}&
        \includegraphics[width=0.32\columnwidth]{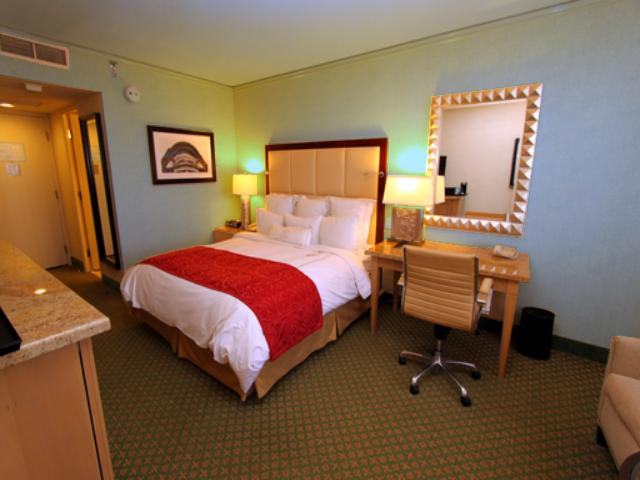}&
        \includegraphics[width=0.32\columnwidth]{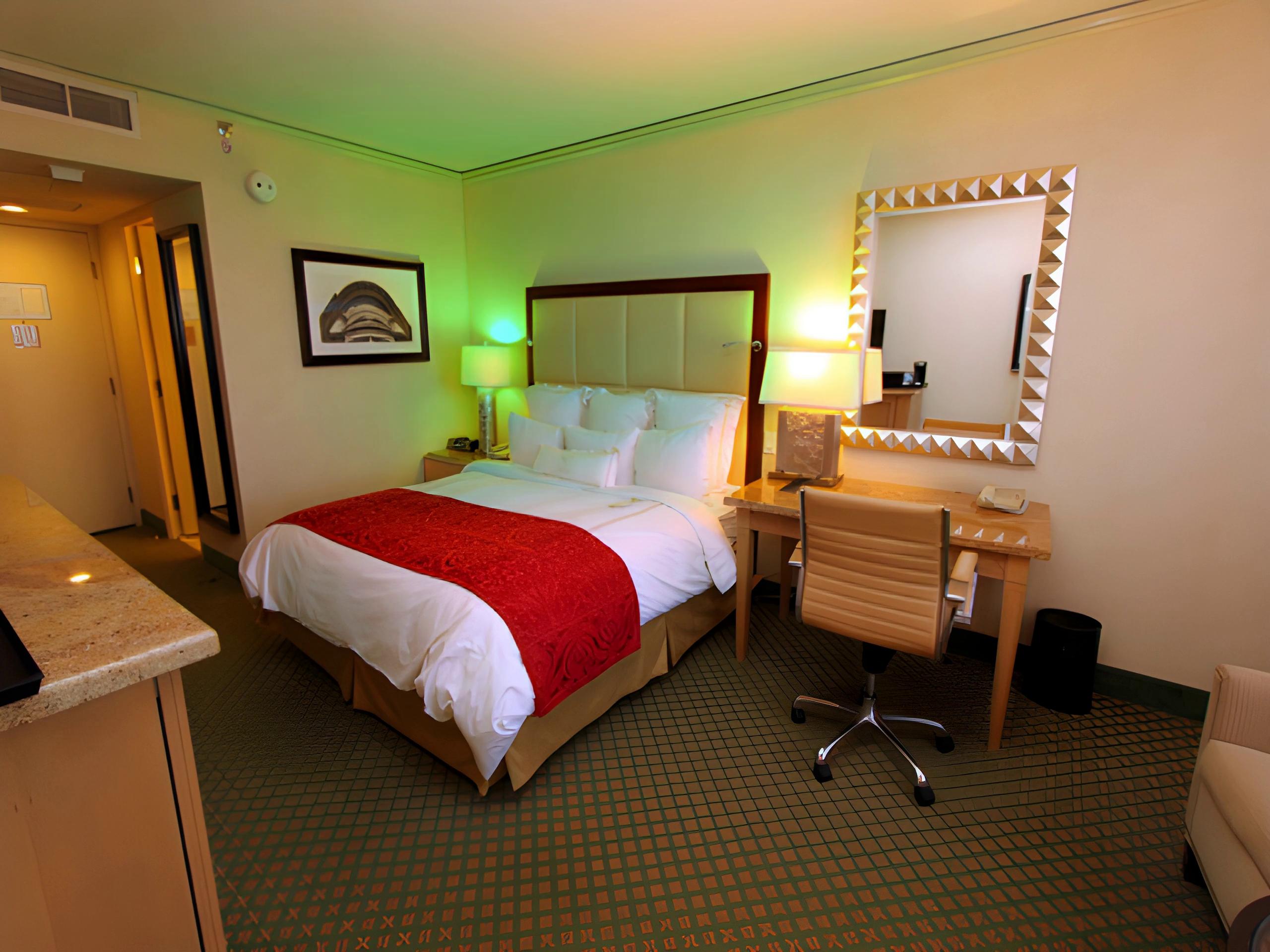}\\
        \includegraphics[width=0.32\columnwidth]{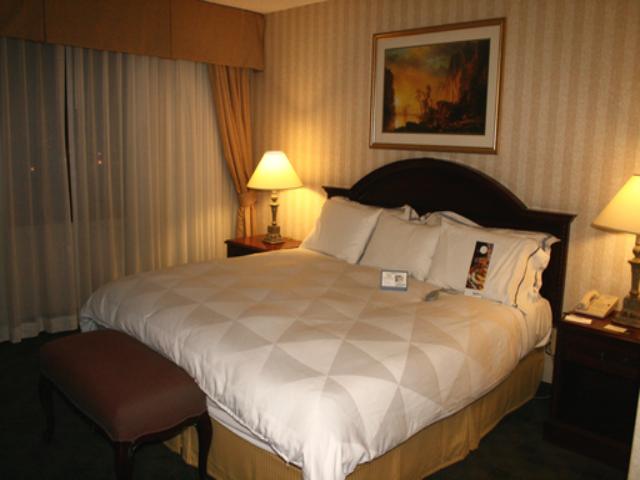}&
        \includegraphics[width=0.32\columnwidth]{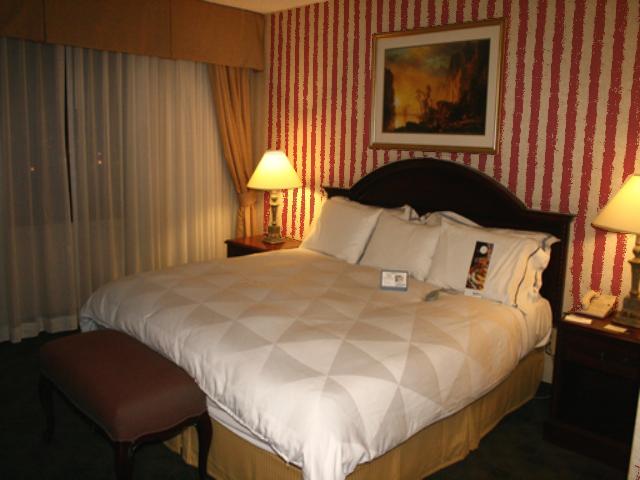}&
        \includegraphics[width=0.32\columnwidth]{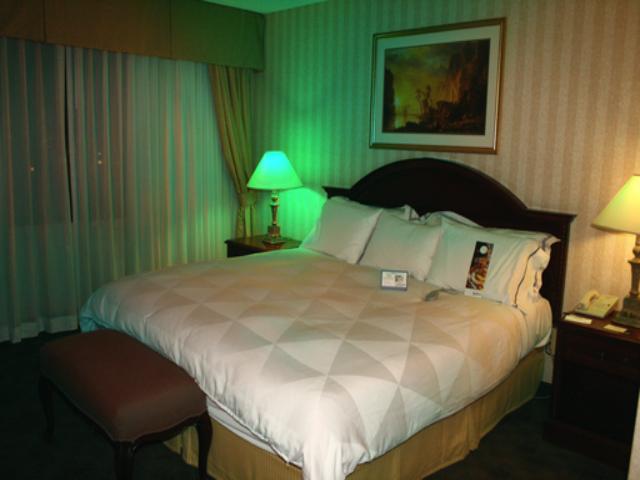}\\
        Input & Material Editing & Lighting Editing 
  \end{tabular}
  \caption{\textbf{Applications.} 
  We show two key intrinsic image editing applications. 
  Our method produces sharp and consistent materials without baked-in lighting, enabling convincing material editing.
  Our lighting representation allows fitting to small light sources, even close to objects, and editing them independently. }
  \label{fig:supp:applications}
\end{figure}

Our lighting provides a flexible yet controllable way to represent lighting in the scene. 
After fitting, the emission weights of the light sources can be edited independently. 
Thanks to the emissive representation, we can achieve physically realistic relighting (\cref{fig:supp:applications}).

\section{Method Details}
\label{sec:supp:method}
\subsection{Lighting Optimization}
We use a hybrid lighting representation. 
For global and out-of-view lighting effects, we use a pre-integrated environment lighting parametrized by Spherical Gaussians (SG) \cite{ComplexInvIndoor}. 
However, such representation alone is not sufficient in our case since indoor scenes often have multiple light sources close to objects, even with different emission profiles and varying colors requiring a spatially-varying lighting representation. 
To achieve a controllable yet expressive representation, we additionally use $N_{light}$ point light sources. 
We use SG emission profile for the point lights to further improve the expressivity. 

Specifically, for the global environment map and also for each point lights, we use $N_{sg}$ SGs with separate 3-channel weights. 
The point lights positions are initialized over a grid in image space and are backprojected to 3D with $1e{-}2$ offset from the surface in the normal direction in normalized depth space. 
The emission profiles are initialized with minimal uniform emission.

We use our predicted materials, and OmniData \cite{OmniData} normal estimation to rerender the scene and optimize the lighting parameters with L2 reconstruction loss. 
We consider every light source for each pixel but without occlusions. 
We found that using the absolute value of the geometry term makes the optimization more stable because otherwise if a light source happens to move behind an object out of the scene, it receives no gradients anymore. 

Without any regularization, this representation might end up representing a single light source with multiple point lights distributed over a sphere around the true source. 
To avoid such a scenario, we apply two regularization terms and an adaptive pruning scheme to motivate using the minimal number of point lights.
We regularize the emission weight $w_{j}$ of all SGs and penalize the inverse distance to the nearest surface $d_{near}$ to move the lights further away from the reflections (\cref{eq:supp:method:lighting}).

\begin{equation}
\label{eq:supp:method:lighting}
\begin{split}
    L_{pos} & = \sum_{i}^{N_{light}} 1 / d_{i, near} \\
    L_{val} & = \sum_{i}^{N_{light}}\sum_{j}^{N_{sg}} w_{i,j} \\
    L & = L_{rec} + \lambda_{pos} L_{pos} + \lambda_{val} L_{val}
\end{split}
\end{equation}

\vspace{12pt}
We use Adam optimizer \cite{Adam} with initial learning rate $5e{-}2$, $\lambda_{pos}{=}1e{-}6$, $\lambda_{val}{=}1e{-}4$, $N_{light}{=}6{\times}8$, $N_{sg}{=}2{\times}6$. 
If the loss starts to stagnate, we reduce the learning rate by a factor of $0.5$ and also prune the weakest light sources. 
We disable every light sources, which total intensity is smaller then $5\%$ of the strongest light source. 
We stop the optimization if the performance stagnates longer. 
The whole optimization usually takes $5{-}10$ minutes on a single A6000 GPU depending on the scene complexity.

\section{Experiment Details}
\label{sec:supp:baselines}
\noindent\textbf{Baselines. } Both baselines \cite{ComplexInvIndoor, ComplexInvIndoorMC} have been trained for $320 \times 240$ resolution.
As reported in the original papers, evaluating them on higher resolution leads to performance degradation; thus, we also evaluate them on this resolution. 

\section{Additional Results}
\label{sec:supp:results}
\subsection{Synthetic Results}
\label{sec:supp:synthetic}
We provide additional material estimation results on the InteriorVerse dataset \cite{ComplexInvIndoorMC} in \cref{fig:supp:synthetic}. 

\begin{figure}[t]
  \centering
  \begin{tabular}{cc}
        \includegraphics[width=0.465\columnwidth]{res/experiments/synthetic/diversity/albedo_0_std.jpg}&
        \includegraphics[width=0.432\columnwidth]{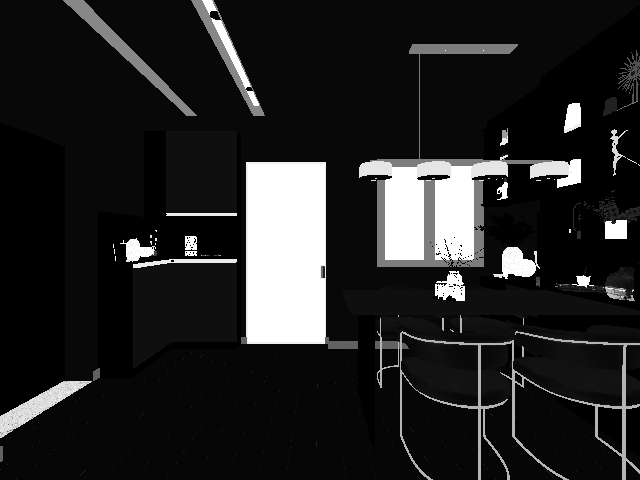}\\
        Albedo Variance & Metallic 
  \end{tabular}
  \caption{\textbf{Variance evaluation.} 
  To investigate one source of ambiguity, we show a comparison between the variance of our albedo predictions and the true metallic map. 
  Glossy objects tend to have higher variance due to the specular ambiguity.
  Further ambiguity arises e.g. from emissive objects, small, under- or over-exposed objects. }
  \label{fig:supp:synthetic:variance}
\end{figure}

\vspace{6pt}
\noindent\textbf{Variance evaluation}
\label{sec:supp:synthetic:variance}
Single-view albedo estimation is an inherently ambiguous tasks, where specularity is one major source of ambiguity. 
We show the correlation between the metallic and albedo variance maps in \cref{fig:supp:synthetic:variance}. 
Glossy objects tend to have higher uncertainty, as also found quantitatively in the main text. 
Note that perfect correlation can not be expected, since specularity is not the only source of ambiguity.

\subsection{Real Results}
\label{sec:supp:real}
We provide additional material estimation results on the IIW dataset \cite{IIW} in \cref{fig:supp:real}. 

\vspace{6pt}
\noindent\textbf{User study}
\label{sec:supp:user_study}
We conduct a user study to additionally evaluate the real-world predictions perceptually too. 

\begin{figure*}[t]
  \centering
  \setlength\tabcolsep{1.25pt}
  \resizebox{\textwidth}{!}{
      \begin{tabular}{ccccccccc}
           
            \rotatebox{90}{\footnotesize{Ours}}&
            \includegraphics[width=0.11\textwidth]{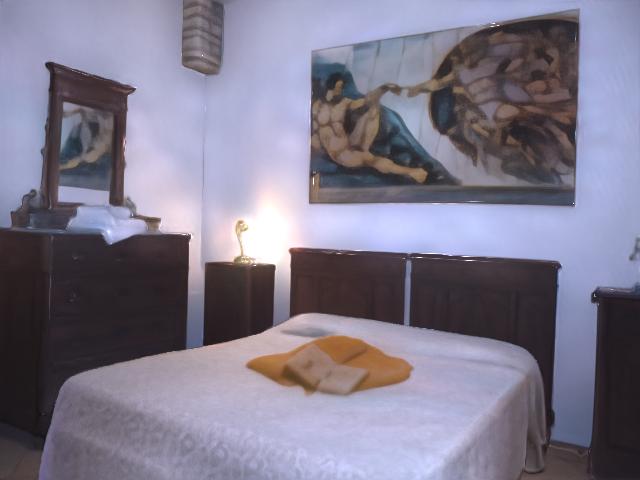}&
            \includegraphics[width=0.11\textwidth]{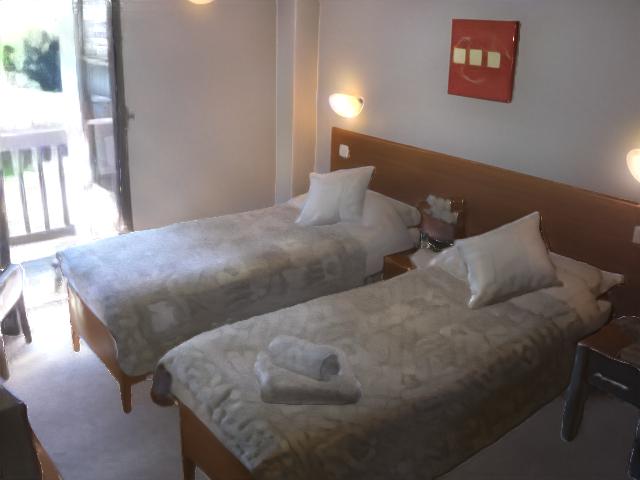}&
            \includegraphics[width=0.11\textwidth]{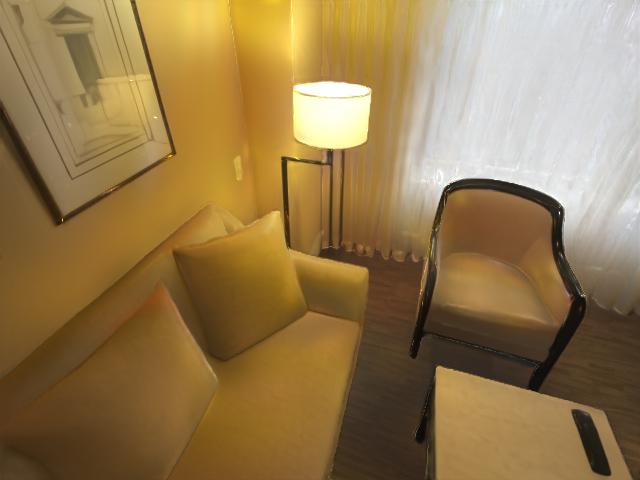}&
            \includegraphics[width=0.11\textwidth]{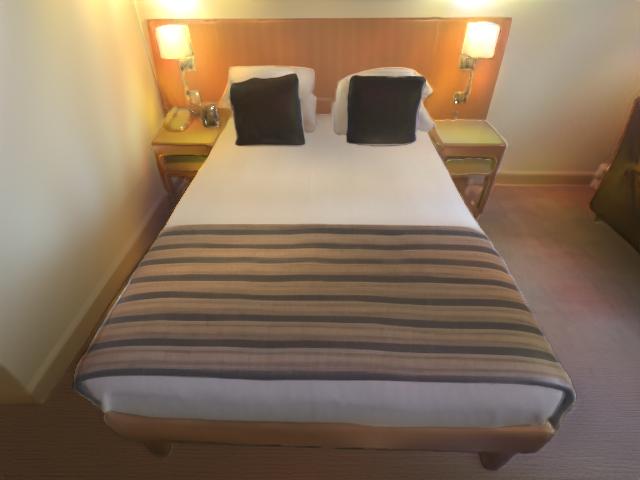}&
            \includegraphics[width=0.11\textwidth]{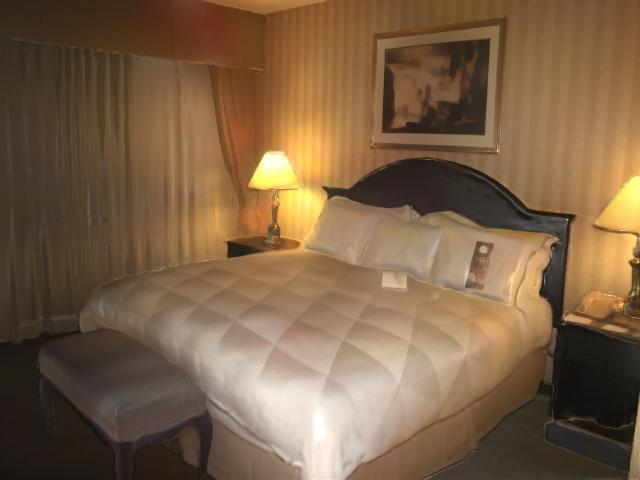}&
            \includegraphics[width=0.11\textwidth]{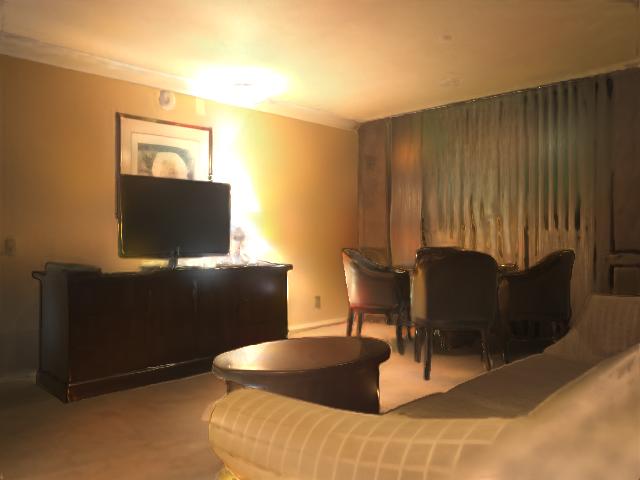}&
            \includegraphics[width=0.11\textwidth]{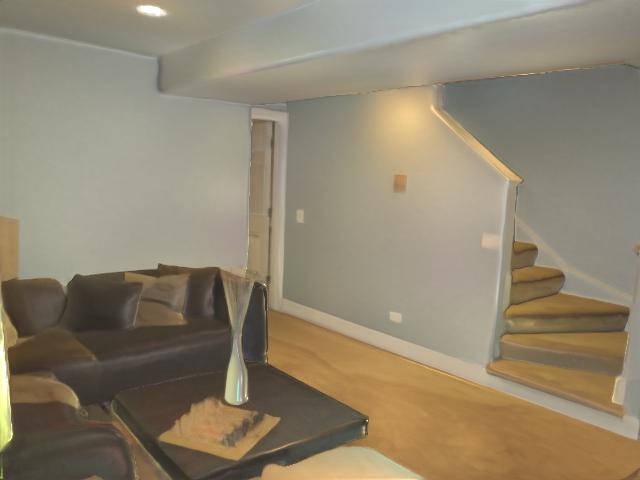}&
            \includegraphics[width=0.11\textwidth]{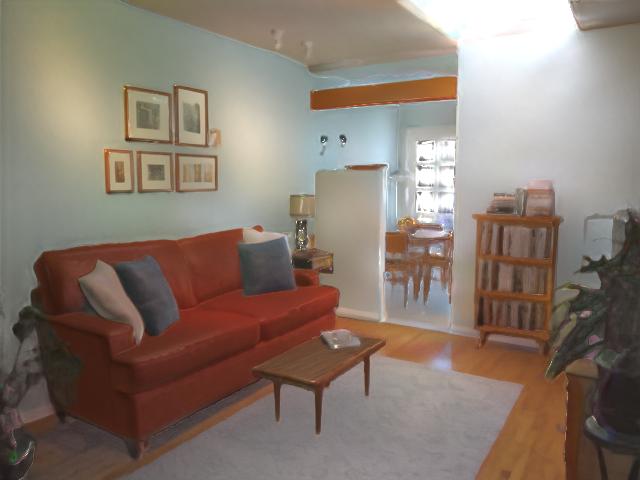}
            \\
            
            \rotatebox{90}{\footnotesize{GT}}&
            \includegraphics[width=0.11\textwidth]{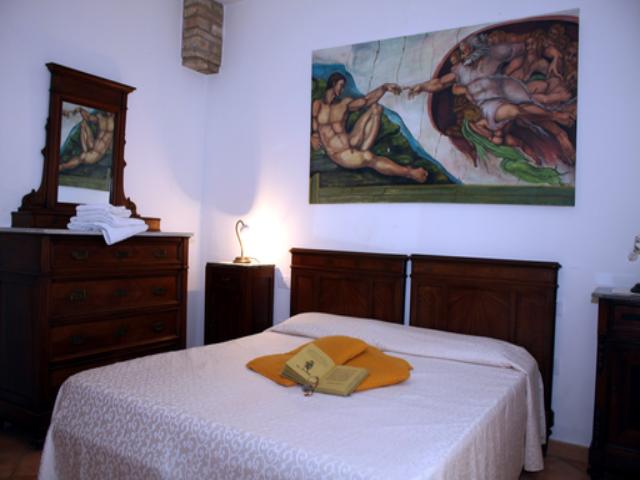}&
            \includegraphics[width=0.11\textwidth]{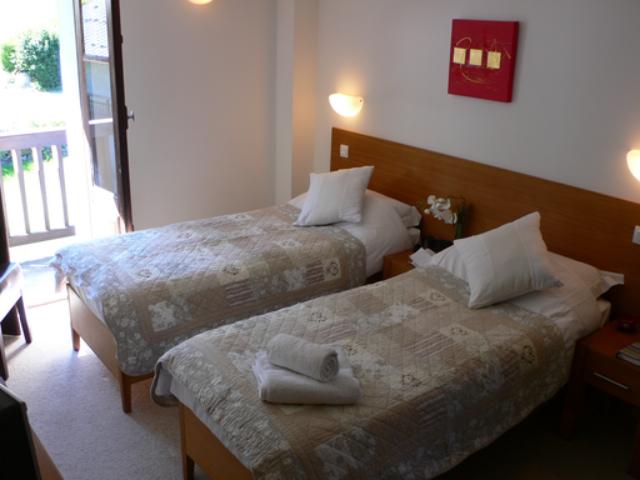}&
            \includegraphics[width=0.11\textwidth]{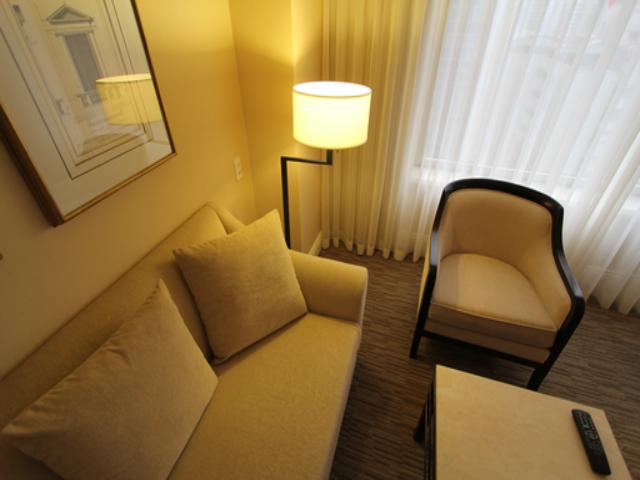}&
            \includegraphics[width=0.11\textwidth]{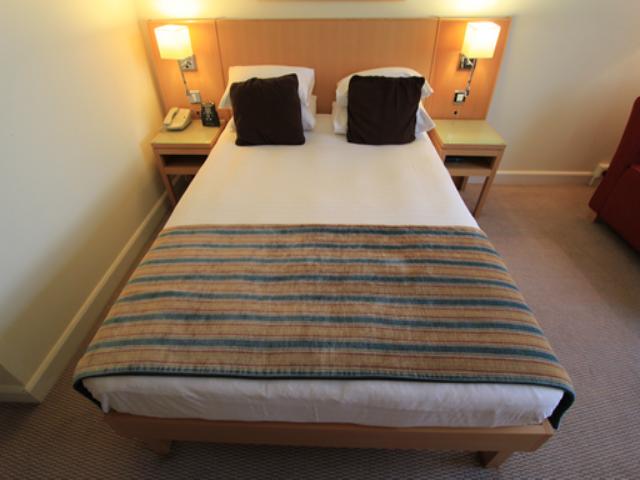}&
            \includegraphics[width=0.11\textwidth]{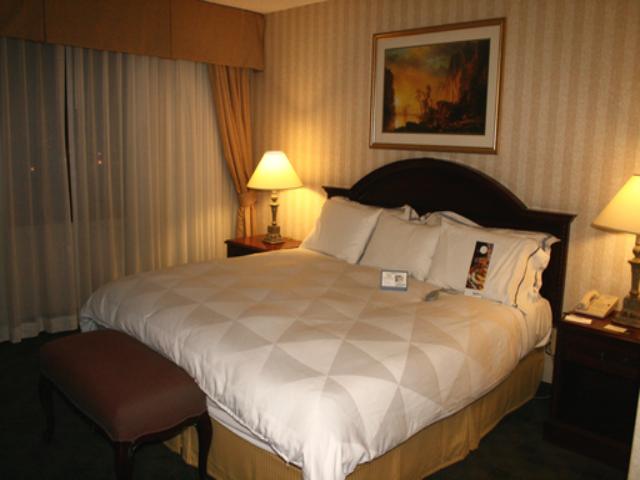}&
            \includegraphics[width=0.11\textwidth]{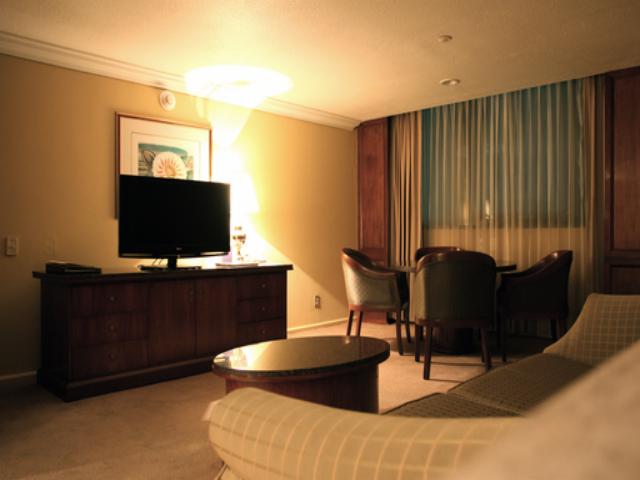}&
            \includegraphics[width=0.11\textwidth]{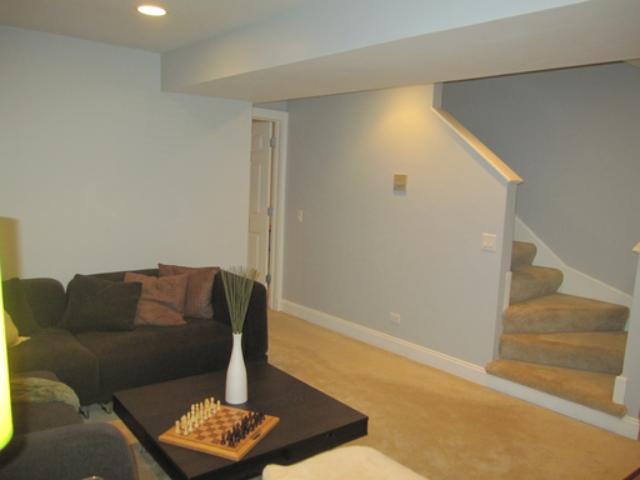}&
            \includegraphics[width=0.11\textwidth]{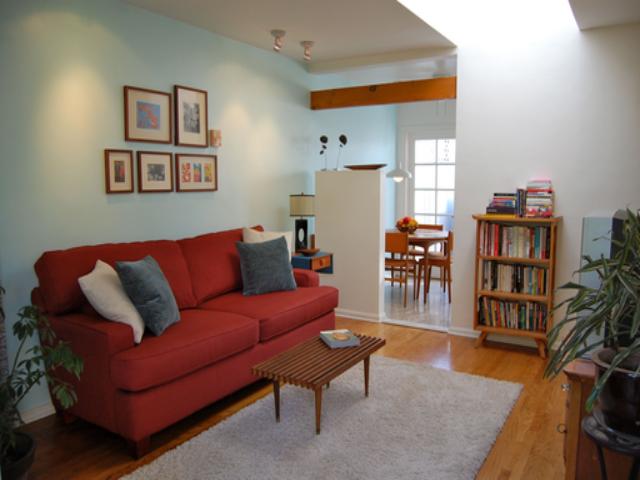}
            \\
          \end{tabular}  }    
  \caption{\textbf{Image reconstruction.} 
  Additional image reconstruction results using our full pipeline. }
  \label{fig:supp:real:rerendering}
\end{figure*}

\vspace{6pt}
\noindent\textbf{Image reconstruction}
\label{sec:supp:user_study}
We provide additional image rerenderings using our full pipeline in \cref{fig:supp:real:rerendering}. 
We thank the authors of \cite{ComplexInvIndoorMC} for providing the code for running their method and for discussing the results.

\begin{figure*}[t]
  \setlength\tabcolsep{1.25pt}
  \centering
  \resizebox{\textwidth}{!}{
  \begin{tabular}{cccccc}
        \rotatebox{90}{\footnotesize{Input}}&
        \includegraphics[width=0.2\linewidth]{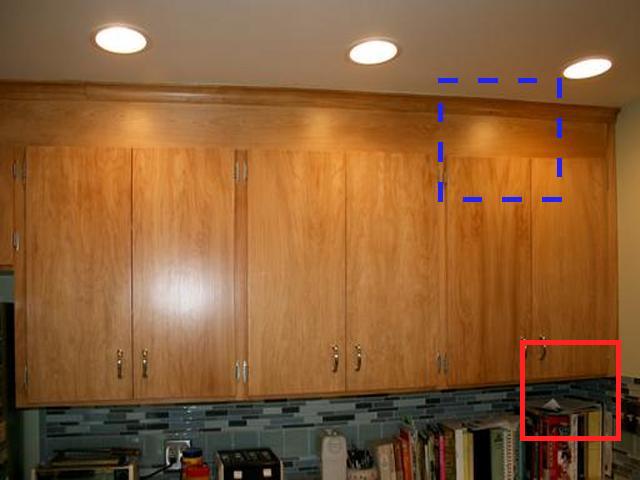}&
        \includegraphics[width=0.2\linewidth]{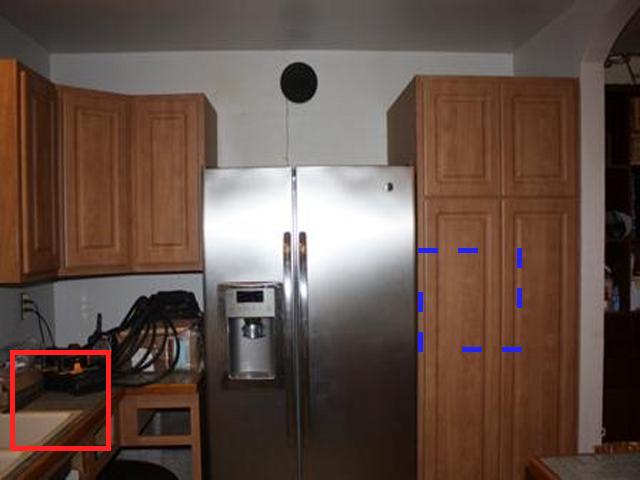}&
        \includegraphics[width=0.2\linewidth]{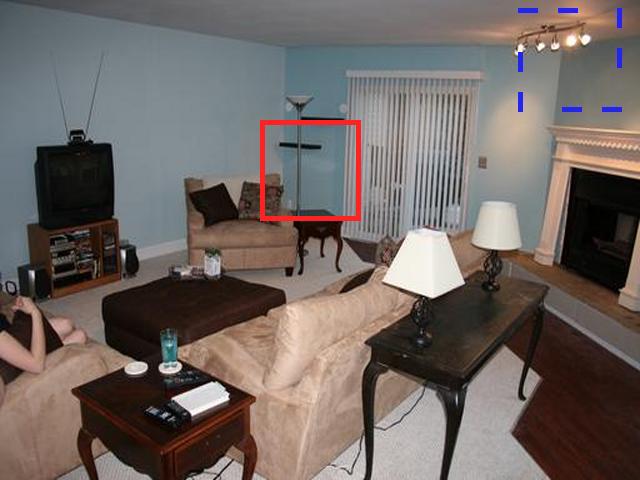}&
        \includegraphics[width=0.2\linewidth]{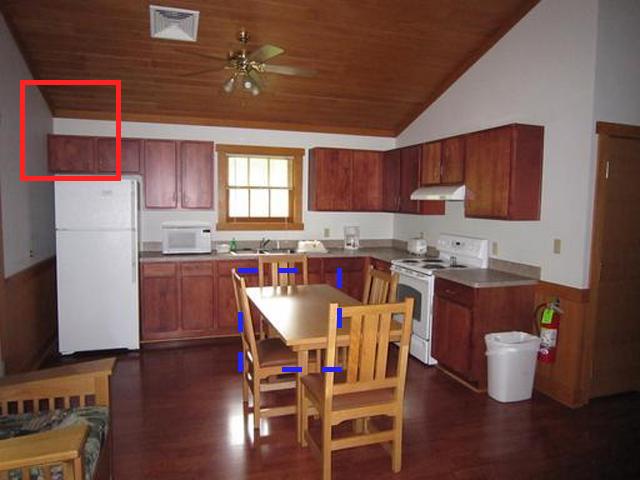}&
        \includegraphics[width=0.2\linewidth]{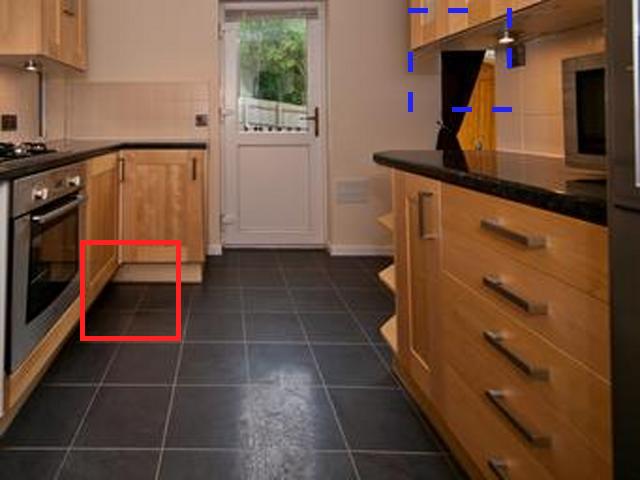}
        \\

        \rotatebox{90}{\footnotesize{S-AWARE \cite{jin2023estimating}}}&
        \includegraphics[width=0.2\linewidth]{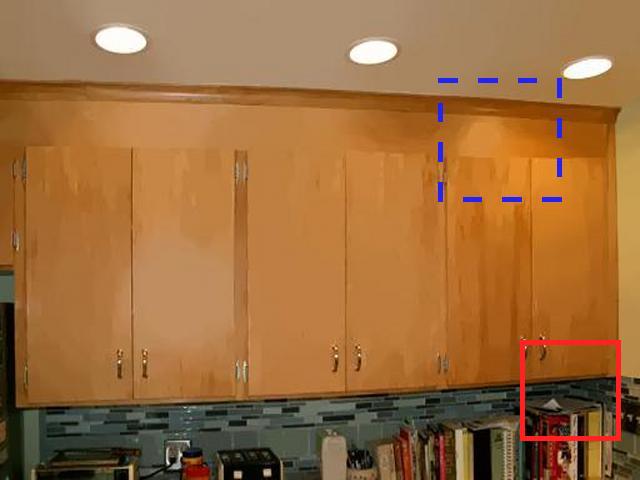}&
        \includegraphics[width=0.2\linewidth]{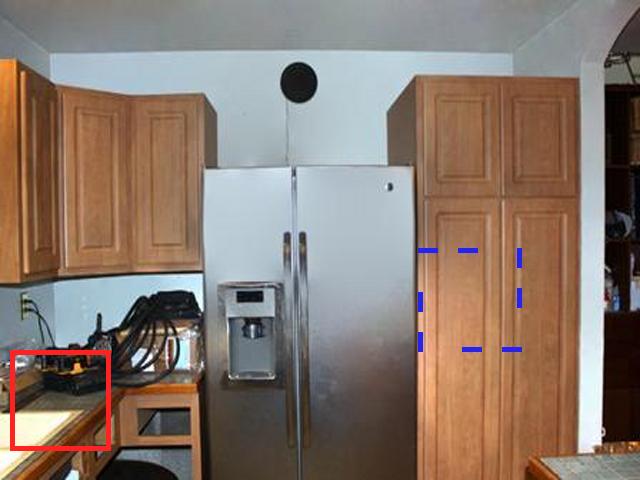}&
        \includegraphics[width=0.2\linewidth]{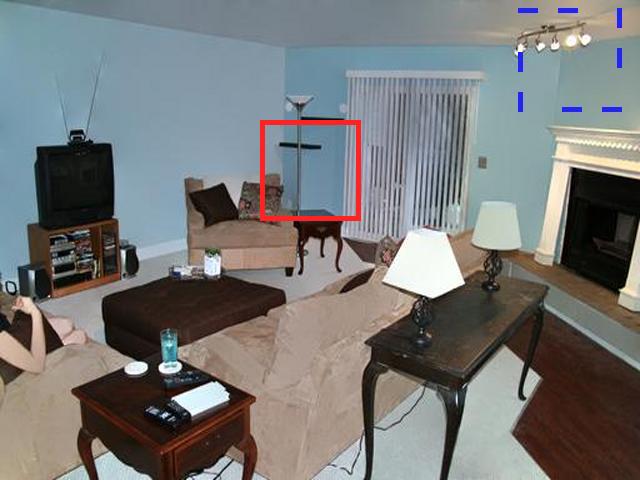}&
        \includegraphics[width=0.2\linewidth]{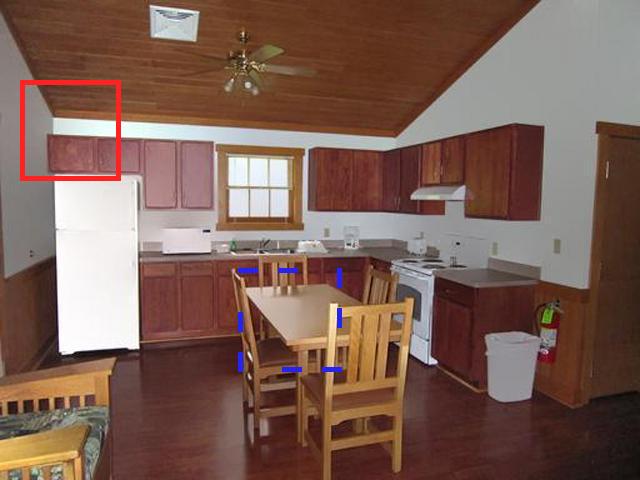}&
        \includegraphics[width=0.2\linewidth]{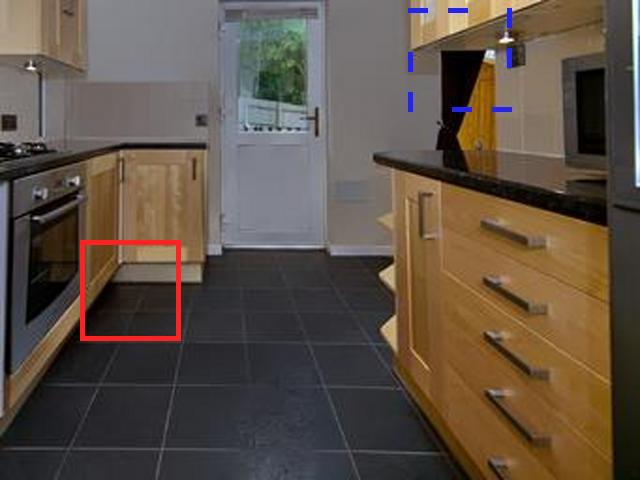}
        \\
        
        \rotatebox{90}{\footnotesize{Ours}}&
        \includegraphics[width=0.2\linewidth]{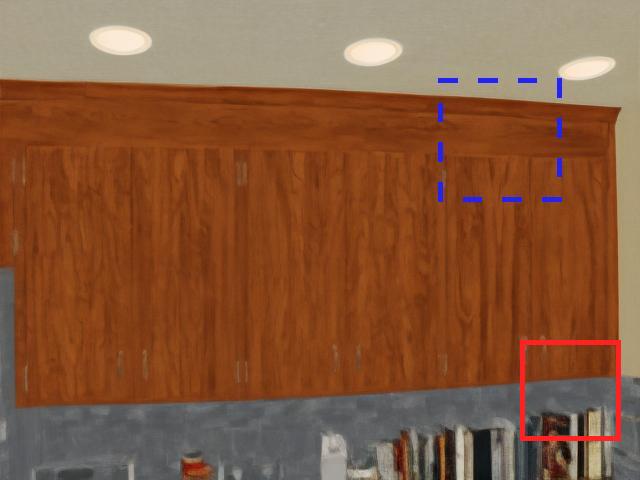}&
        \includegraphics[width=0.2\linewidth]{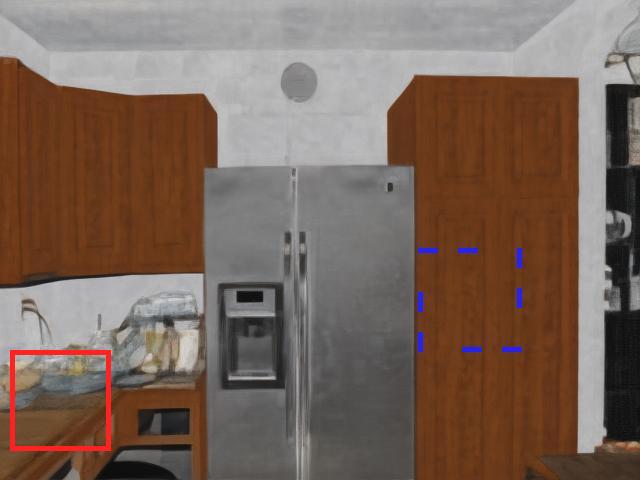}&
        \includegraphics[width=0.2\linewidth]{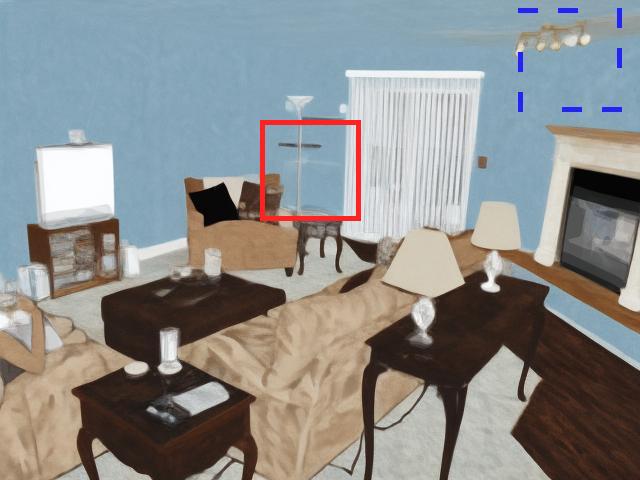}&
        \includegraphics[width=0.2\linewidth]{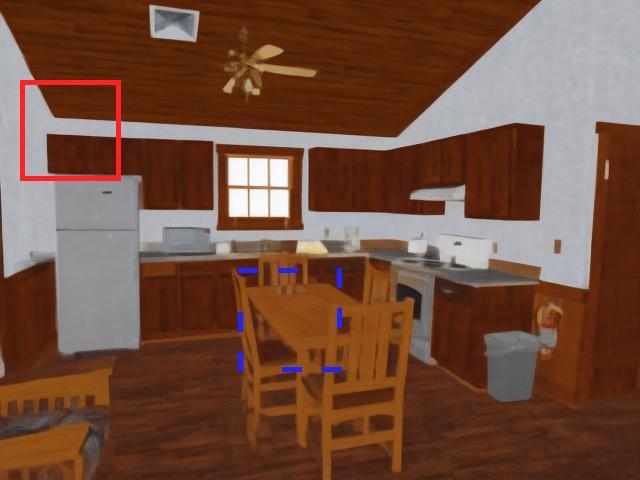}&
        \includegraphics[width=0.2\linewidth]{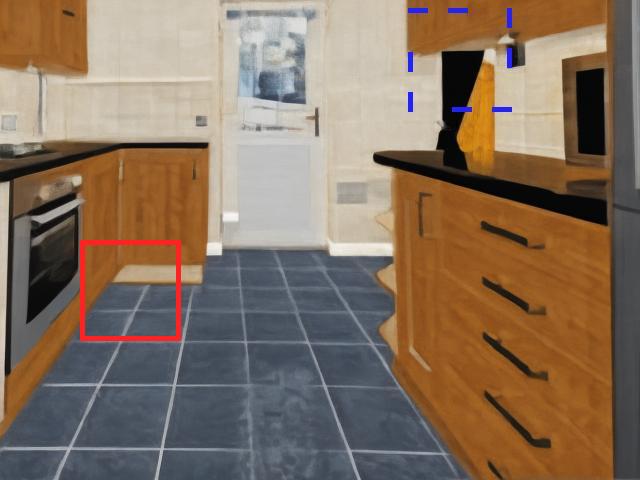}
        \\
  \end{tabular} }
  \vspace{-9pt}
  \caption{\textbf{Comparison to S-AWARE \cite{jin2023estimating}.}
  Our method gives complex material maps and also improved albedo without baked-in \textcolor{blue}{lighting} and \textcolor{red}{shadows}. }
  \label{fig:supp:s_aware}
  \vspace{-12pt}
\end{figure*}

\vspace{6pt}
\noindent\textbf{S-AWARE Network \cite{jin2023estimating} }
We compare against S-AWARE \cite{jin2023estimating} in \cref{fig:supp:s_aware}. 
Our method predicts roughness and metallic properties as well and improves upon the albedo estimation by avoiding baked-in lighting or shadows. 
We thank the authors for providing us with their results.

\begin{table}
\begin{center}
\resizebox{\columnwidth}{!}{
\begin{tabular}{l|cccc|c}
  \toprule
   & \multicolumn{4}{c|}{InteriorVerse} & IIW \\
   & PSNR $\uparrow$ & SSIM $\uparrow$ & LPIPS $\downarrow$ & FID $\downarrow$ & WHDR $\downarrow$ \\
   \midrule 
  Ours - ImageOnly   & 17.42 {\footnotesize $\pm$ 3.08} & 0.80 {\footnotesize $\pm$ 0.08} & 0.22 {\footnotesize $\pm$ 0.08} & 25.42 & 22.02 {\footnotesize $\pm$ 11.99}\\
  Ours - GTDepth  & 16.57 {\footnotesize $\pm$ 3.79} & 0.80 {\footnotesize $\pm$ 0.08} & 0.22 {\footnotesize $\pm$ 0.08} & 24.36 & 17.05 {\footnotesize $\pm$ 10.29}\\
  Ours - PredDepth   & \textbf{18.31} {\footnotesize $\pm$ 3.44} & \textbf{0.82} {\footnotesize $\pm$ 0.08} & \textbf{0.19} {\footnotesize $\pm$ 0.07} & \textbf{22.60} & \textbf{16.66} {\footnotesize $\pm$ 10.21}\\
  \bottomrule
\end{tabular}
}
\end{center}
\caption{\textbf{Effect of depth-conditioning.} 
Geometry information can give helpful cues for appearance decomposition and improve our models performance. 
We compare our image-conditioned method with additional conditioning on depth and normal maps. 
During training, one variant uses ground-truth geometry, the other uses predicted geometry using OmniData \cite{OmniData}. 
In test time, we use the predicted geometry as conditioning. 
Nevertheless, to stay consistent with the baselines, we kept the image-conditioning variant. }
\label{tab:experiments:ablations:depth}
\end{table}

\section{Additional Ablations}
\label{sec:supp:ablation}
\noindent\textbf{Effect of depth-conditioning. } 
Our approach uses only a single image as input.
However, geometry information can give beneficial cues for the appearance decomposition, since a physically-based renderer would require the normals for the shading and the depth for the global illumination estimation. 
To test this hypothesis, we train two other variants of our model. 
Both variants use additional depth and normal inputs.
To provide a fair comparison between the variants, we evaluate all the methods without ground-truth depth and normal data. 
We use OmniData \cite{OmniData} to predict the depth and normal maps of the input view and use the predicted geometry as conditioning. 
One of our variants was trained with ground-truth geometry information, the other with predicted geometry.

We show qualitative results in \cref{tab:experiments:ablations:depth}. 
Here, we use the mean of $10$ samples. 
Indeed, geometry information provides helpful cues for appearance decomposition and can improve the performance of our model.
However, to stay consistent with all the other baselines, we kept the image-conditioned version as our main model.

\begin{figure*}
  \setlength\tabcolsep{1.25pt}
  \centering
  \begin{tabular}{c|cc|c|cc}
        \includegraphics[width=0.19\textwidth]{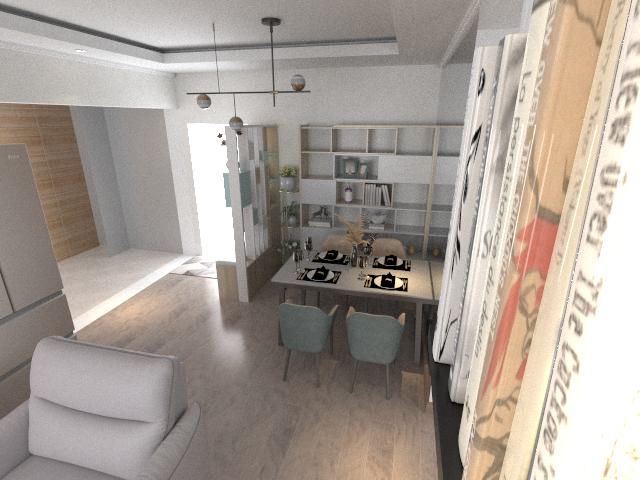}&
        \includegraphics[width=0.19\textwidth]{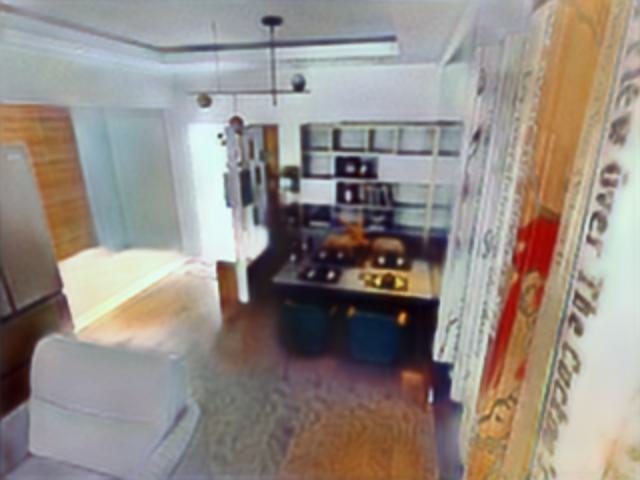}&
        \includegraphics[width=0.19\textwidth]{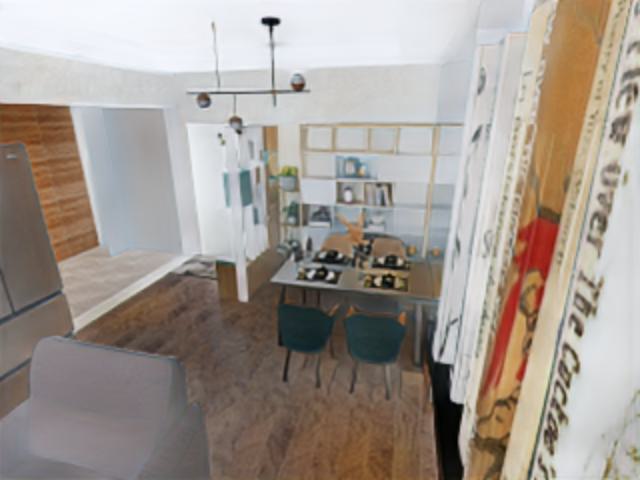}&
        \includegraphics[width=0.19\textwidth]{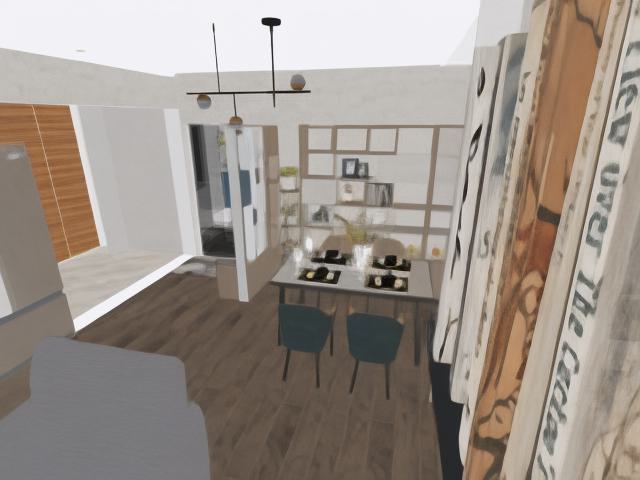}&
        \includegraphics[width=0.19\textwidth]{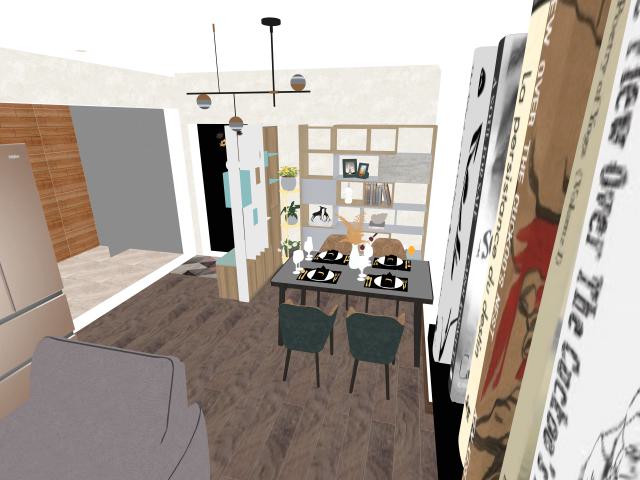}&
        \rotatebox{90}{Albedo}\\     
        &
        \includegraphics[width=0.19\textwidth]{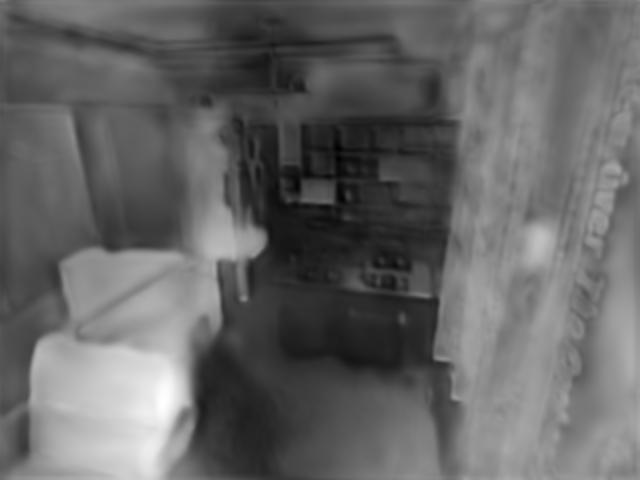}&
        \includegraphics[width=0.19\textwidth]{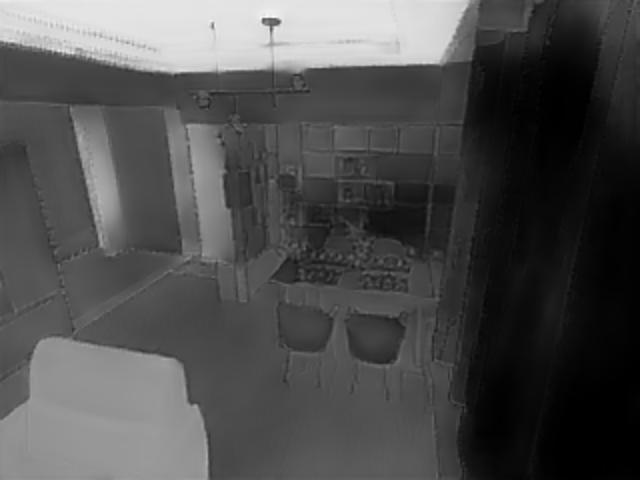}&
        \includegraphics[width=0.19\textwidth]{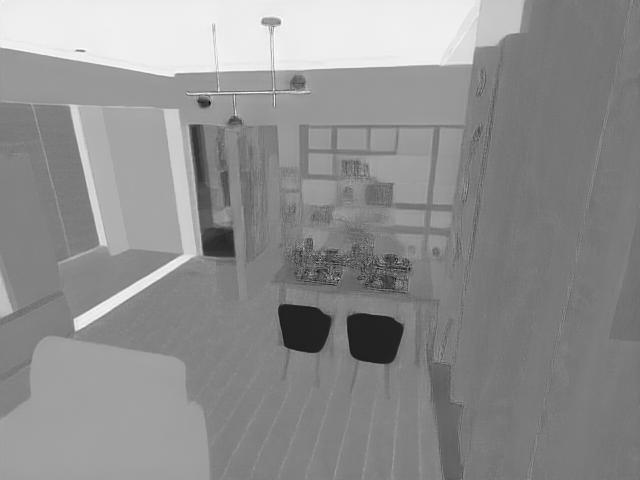}&
        \includegraphics[width=0.19\textwidth]{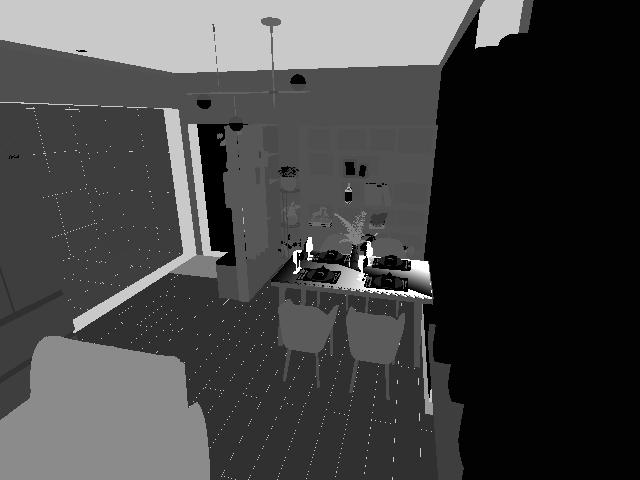}&
        \rotatebox{90}{Roughness}\\       
        &
        \includegraphics[width=0.19\textwidth]{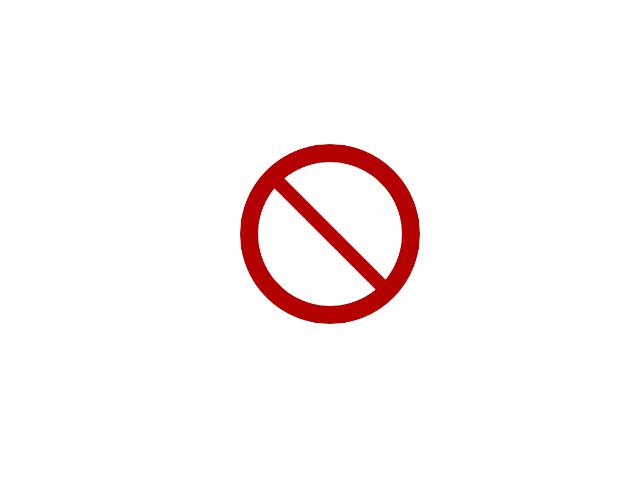}&
        \includegraphics[width=0.19\textwidth]{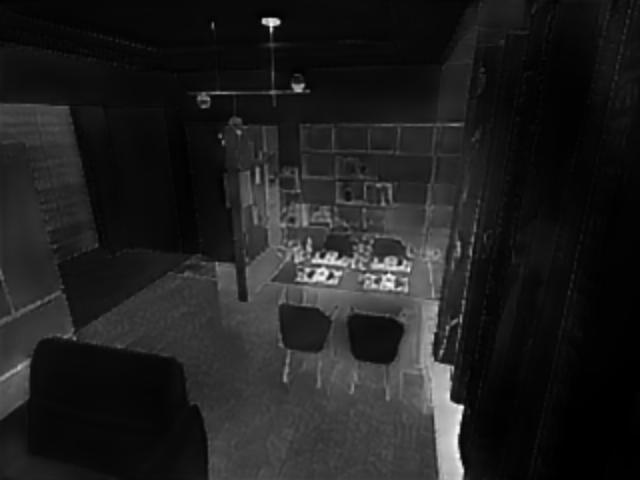}&
        \includegraphics[width=0.19\textwidth]{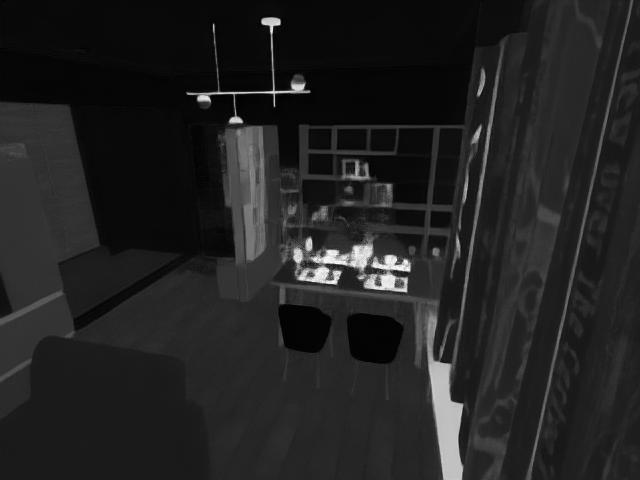}&
        \includegraphics[width=0.19\textwidth]{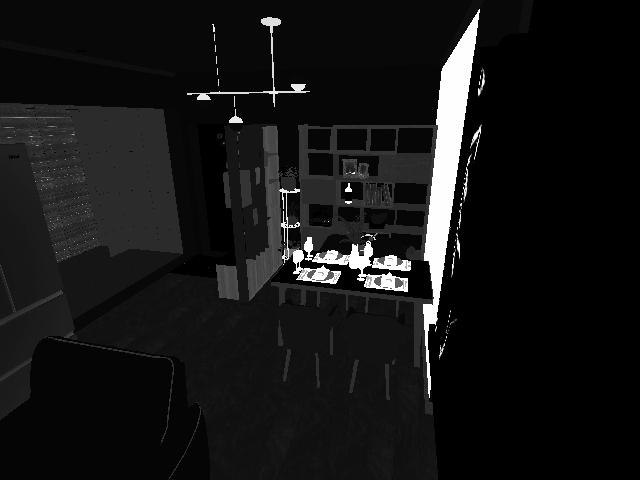}&
        \rotatebox{90}{Metallic}\\

        \includegraphics[width=0.19\textwidth]{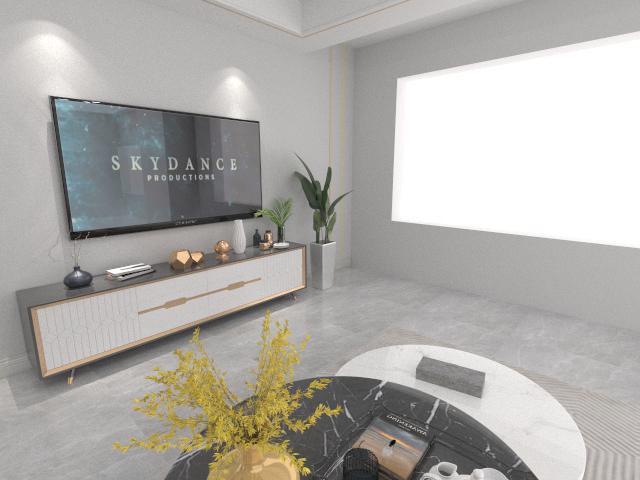}&
        \includegraphics[width=0.19\textwidth]{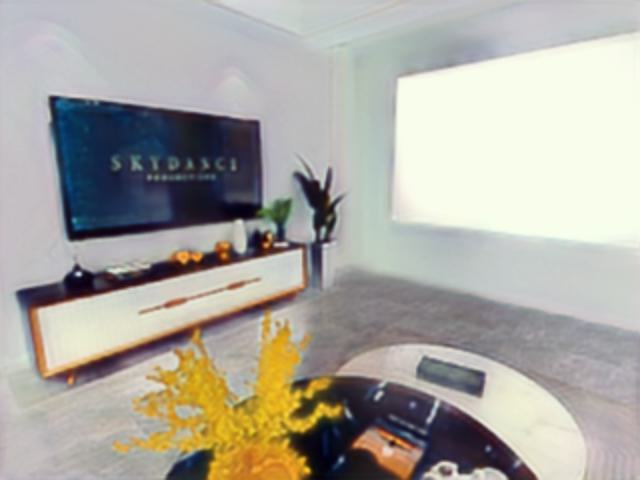}&
        \includegraphics[width=0.19\textwidth]{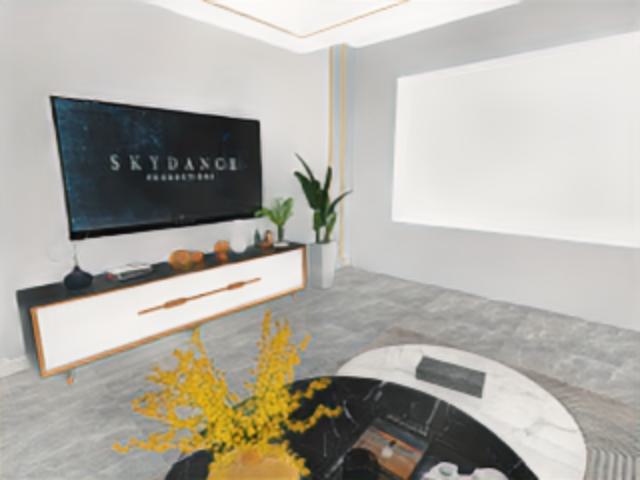}&
        \includegraphics[width=0.19\textwidth]{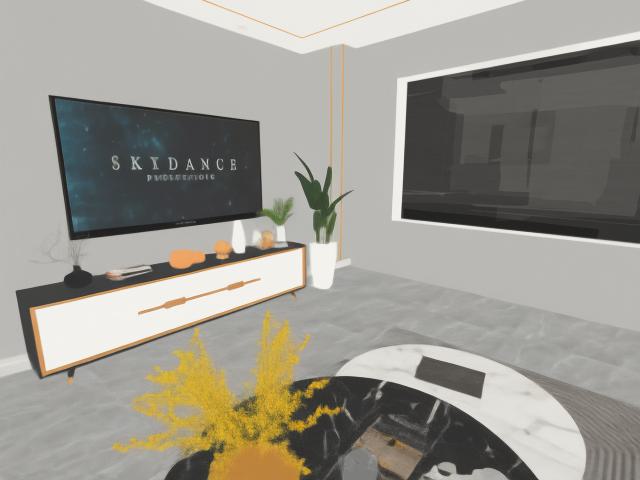}&
        \includegraphics[width=0.19\textwidth]{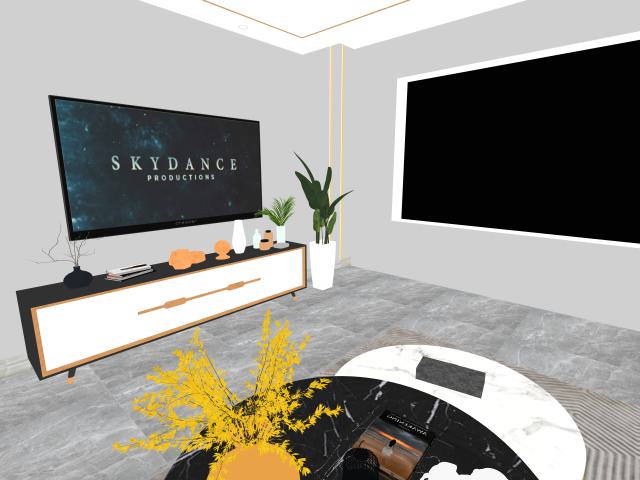}&
        \rotatebox{90}{Albedo}\\     
        &
        \includegraphics[width=0.19\textwidth]{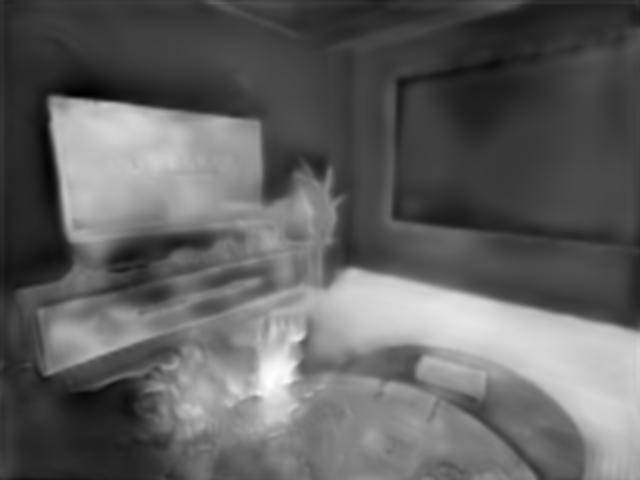}&
        \includegraphics[width=0.19\textwidth]{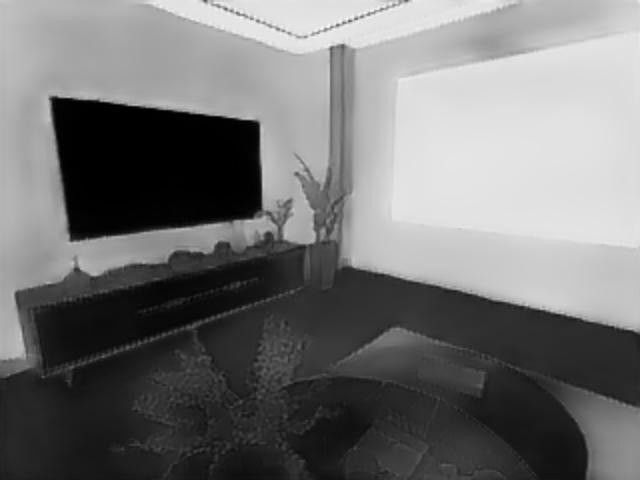}&
        \includegraphics[width=0.19\textwidth]{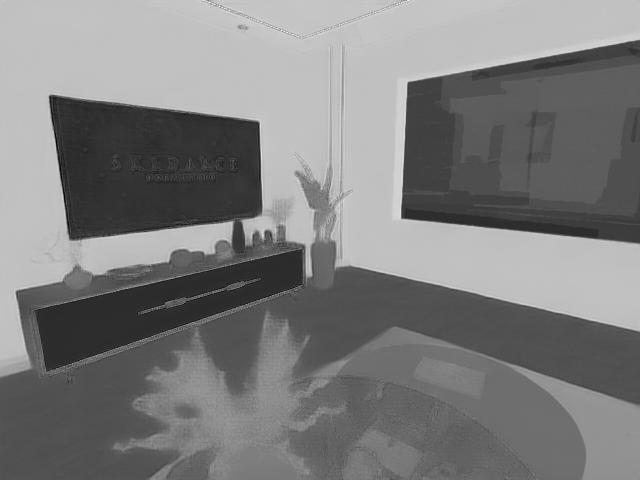}&
        \includegraphics[width=0.19\textwidth]{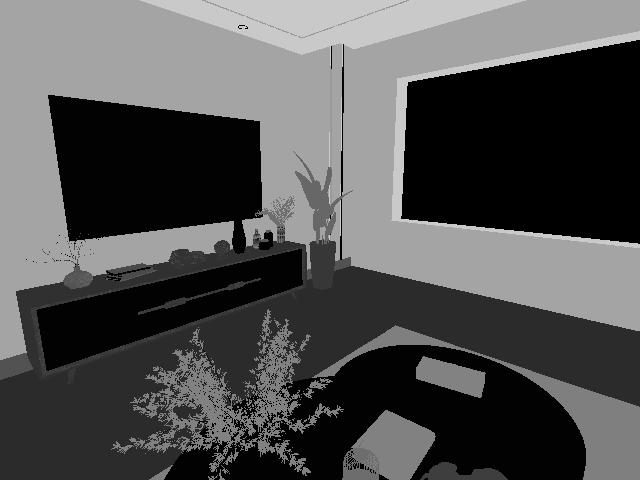}&
        \rotatebox{90}{Roughness}\\       
        &
        \includegraphics[width=0.19\textwidth]{res/supp/synthetic/metallic_no.jpg}&
        \includegraphics[width=0.19\textwidth]{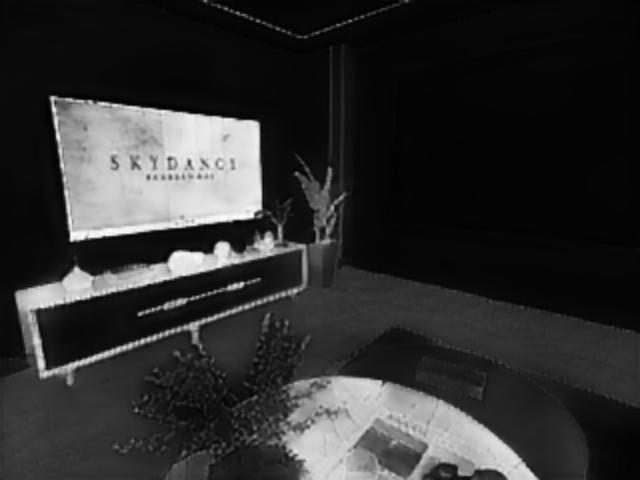}&
        \includegraphics[width=0.19\textwidth]{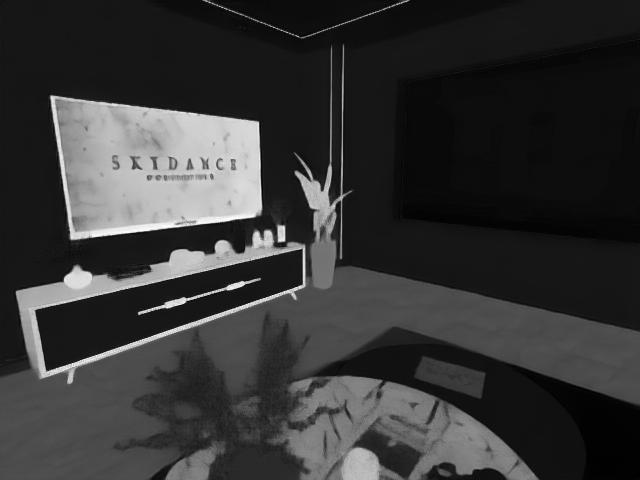}&
        \includegraphics[width=0.19\textwidth]{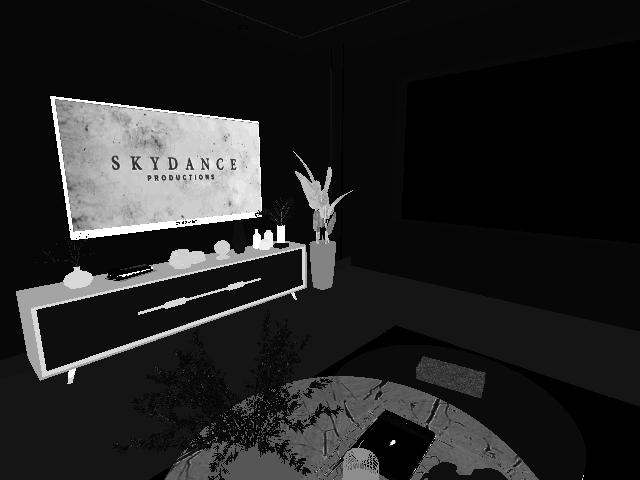}&
        \rotatebox{90}{Metallic}\\ 

          \end{tabular}
  \caption{\textbf{Synthetic material estimation.} 
  Continues on the next page.}
\end{figure*}
\begin{figure*}\ContinuedFloat
  \setlength\tabcolsep{1.25pt}
  \centering
  \begin{tabular}{c|cc|c|cc}
        
        \includegraphics[width=0.19\textwidth]{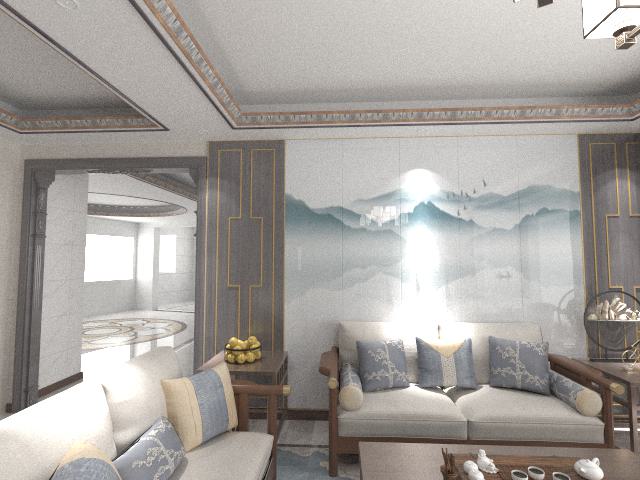}&
        \includegraphics[width=0.19\textwidth]{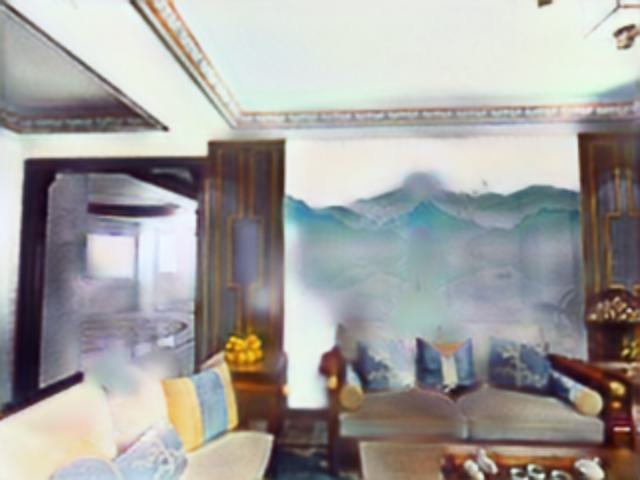}&
        \includegraphics[width=0.19\textwidth]{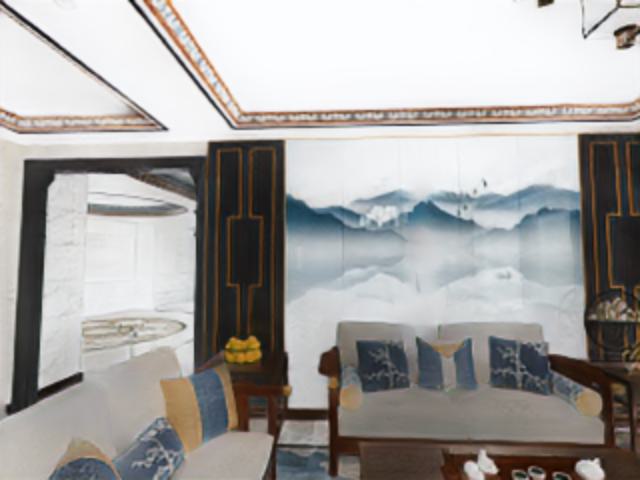}&
        \includegraphics[width=0.19\textwidth]{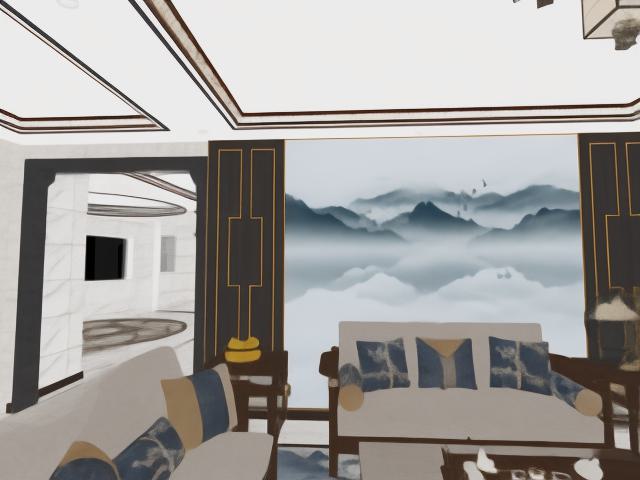}&
        \includegraphics[width=0.19\textwidth]{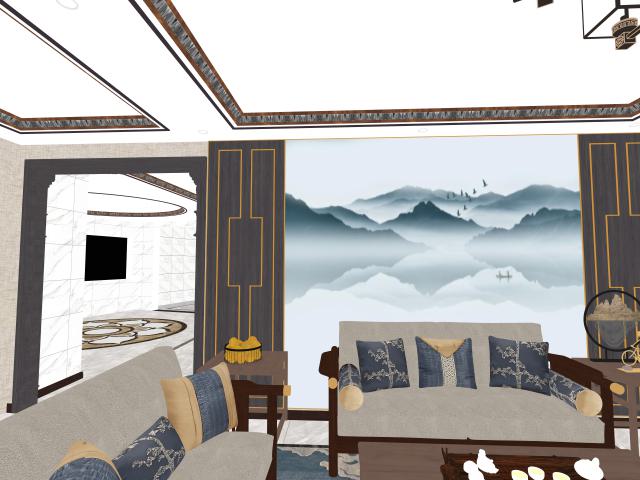}&
        \rotatebox{90}{Albedo}\\     
        &
        \includegraphics[width=0.19\textwidth]{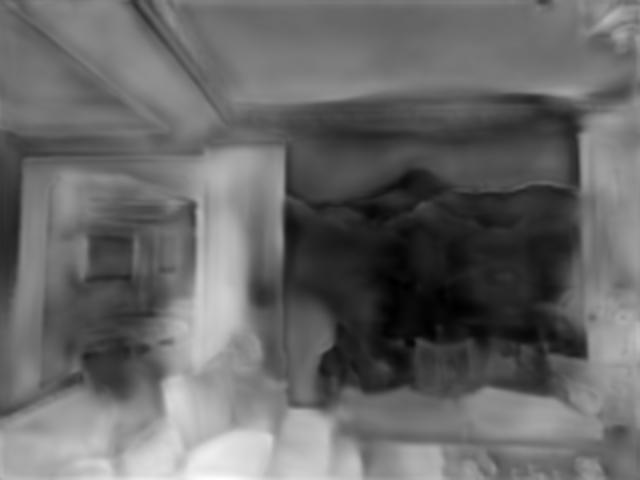}&
        \includegraphics[width=0.19\textwidth]{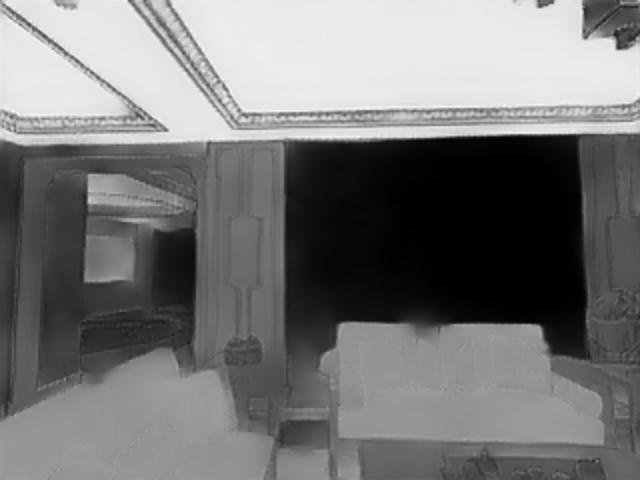}&
        \includegraphics[width=0.19\textwidth]{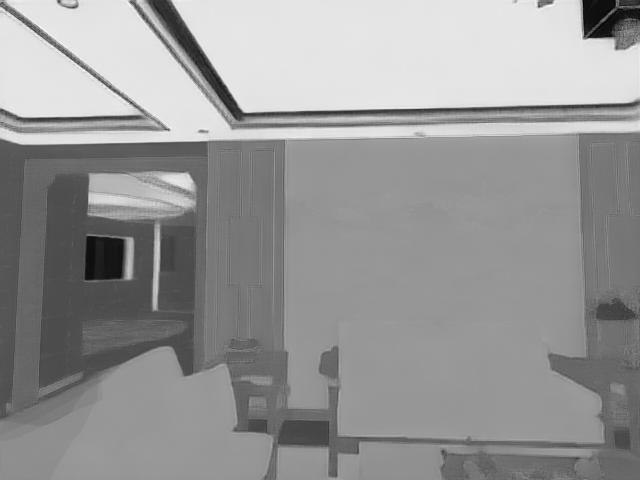}&
        \includegraphics[width=0.19\textwidth]{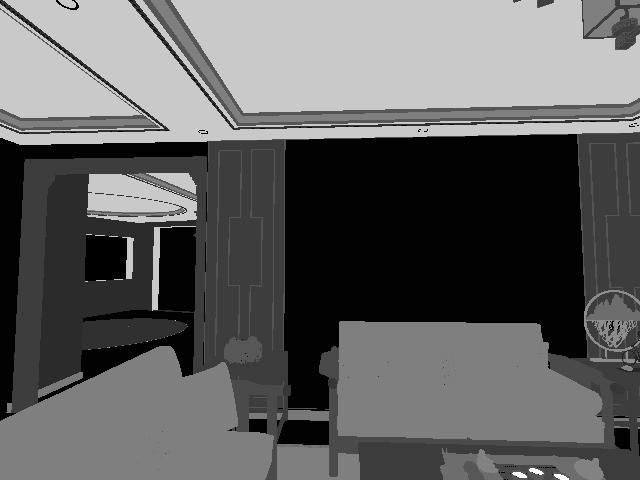}&
        \rotatebox{90}{Roughness}\\       
        &
        \includegraphics[width=0.19\textwidth]{res/supp/synthetic/metallic_no.jpg}&
        \includegraphics[width=0.19\textwidth]{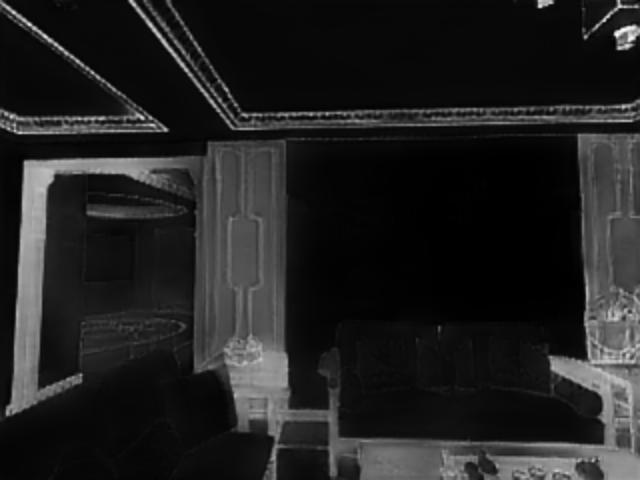}&
        \includegraphics[width=0.19\textwidth]{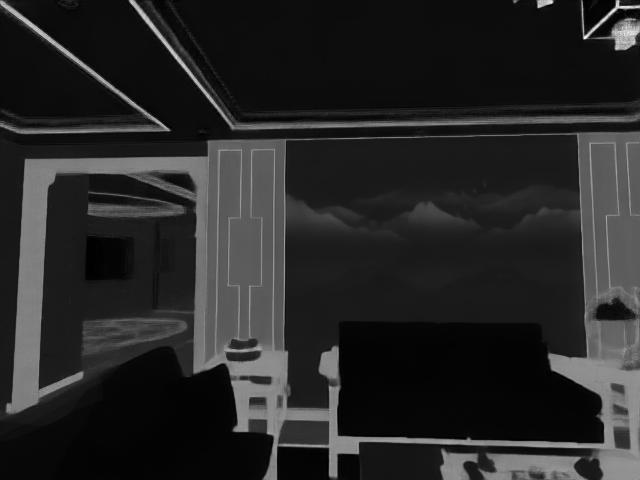}&
        \includegraphics[width=0.19\textwidth]{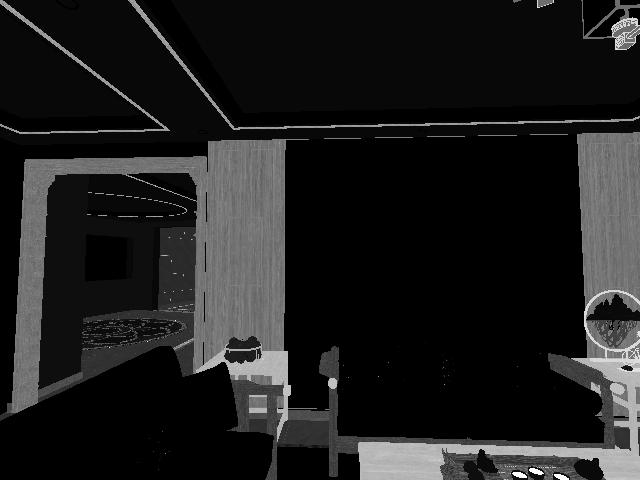}&
        \rotatebox{90}{Metallic}\\ 
        
        \includegraphics[width=0.19\textwidth]{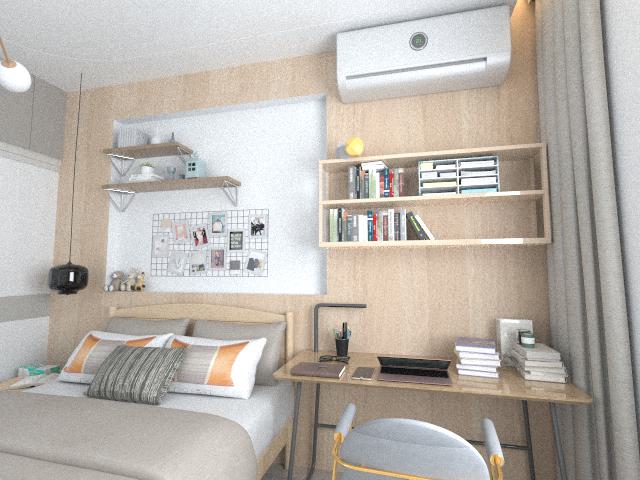}&
        \includegraphics[width=0.19\textwidth]{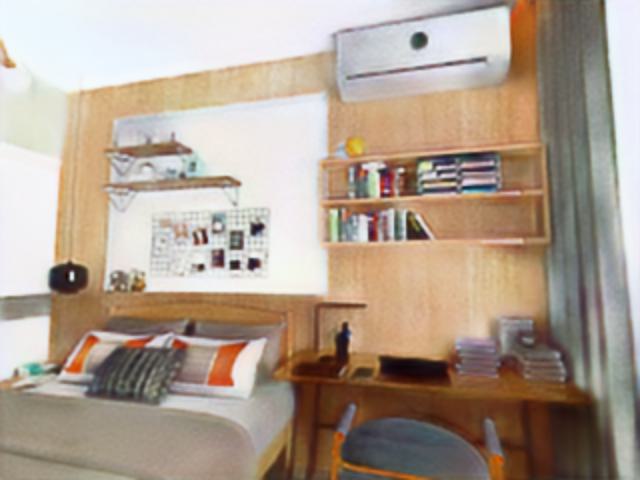}&
        \includegraphics[width=0.19\textwidth]{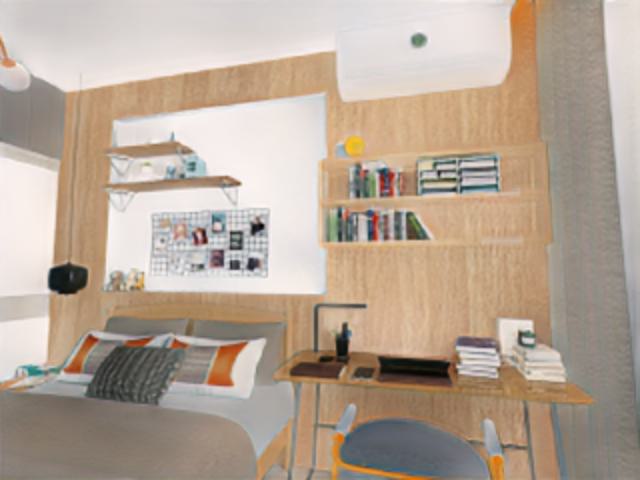}&
        \includegraphics[width=0.19\textwidth]{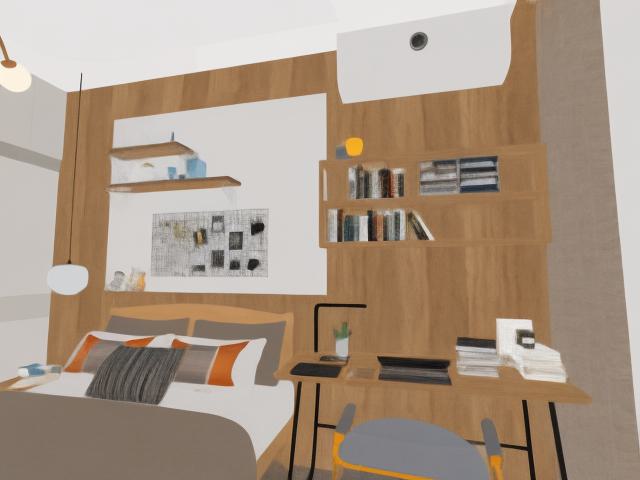}&
        \includegraphics[width=0.19\textwidth]{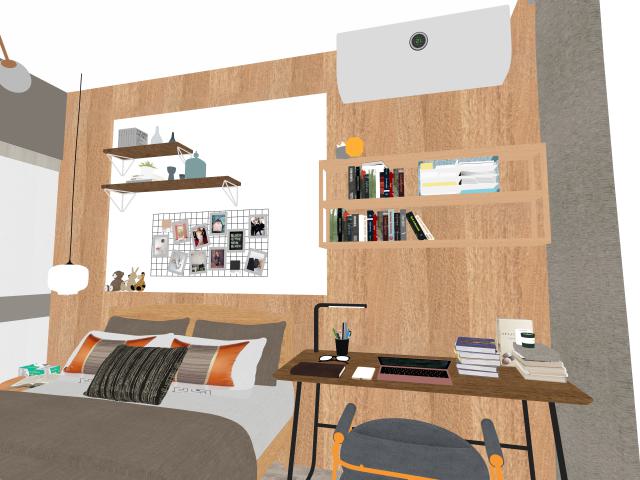}&
        \rotatebox{90}{Albedo}\\     
        &
        \includegraphics[width=0.19\textwidth]{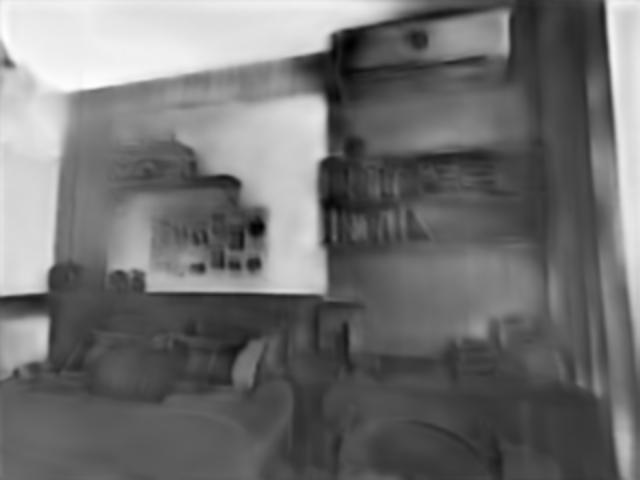}&
        \includegraphics[width=0.19\textwidth]{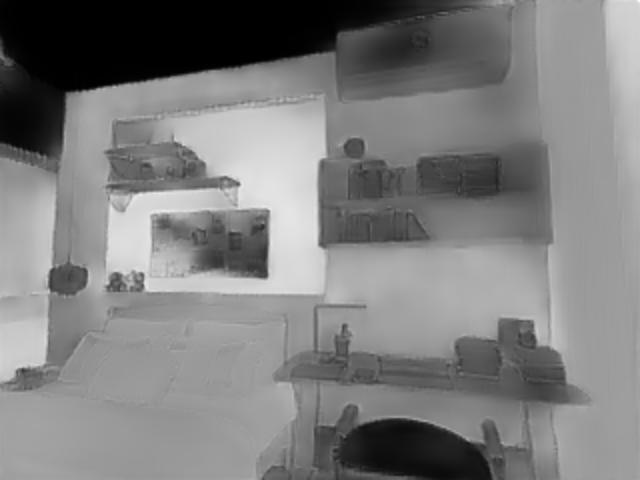}&
        \includegraphics[width=0.19\textwidth]{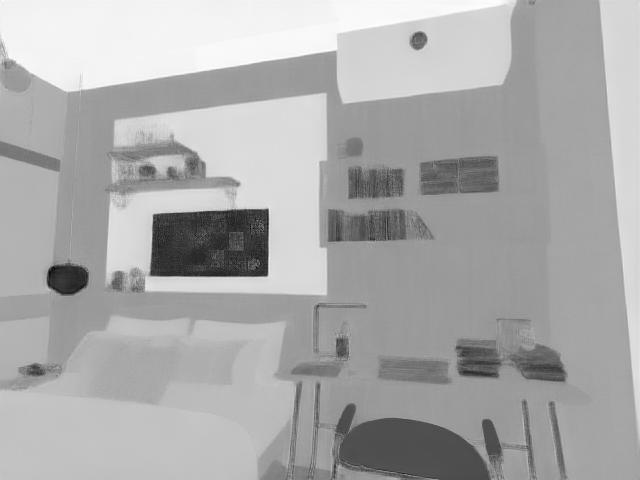}&
        \includegraphics[width=0.19\textwidth]{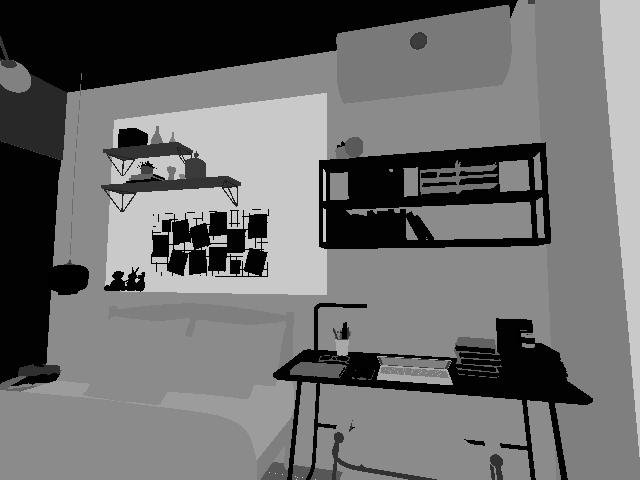}&
        \rotatebox{90}{Roughness}\\       
        &
        \includegraphics[width=0.19\textwidth]{res/supp/synthetic/metallic_no.jpg}&
        \includegraphics[width=0.19\textwidth]{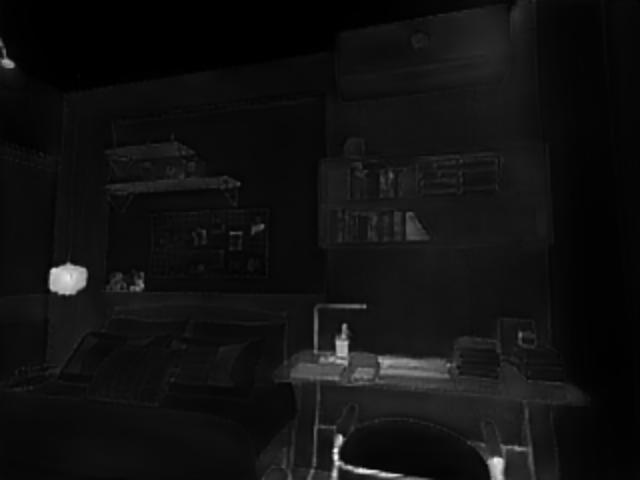}&
        \includegraphics[width=0.19\textwidth]{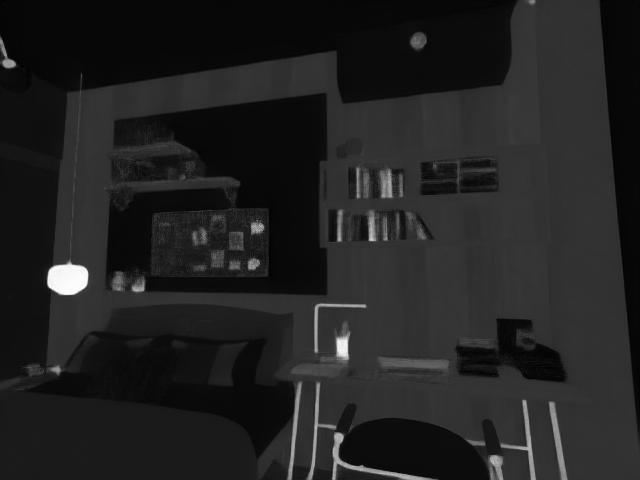}&
        \includegraphics[width=0.19\textwidth]{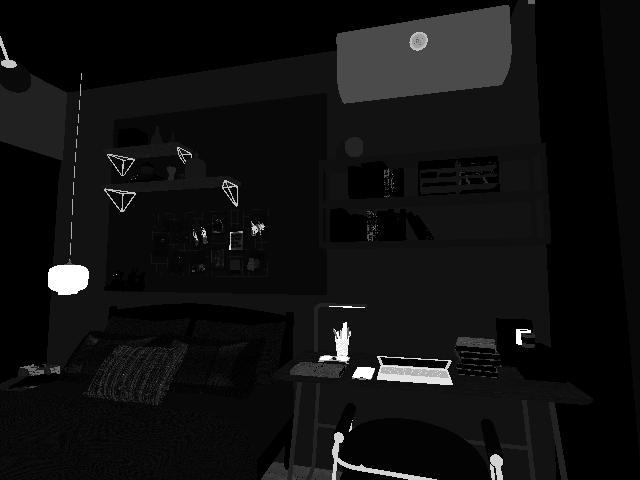}&
        \rotatebox{90}{Metallic}\\ 
        Input & \citet{ComplexInvIndoor} & \citet{ComplexInvIndoorMC} & Ours - Mean & GT
  \end{tabular}
  \caption{\textbf{Synthetic material estimation.} 
  We compare our material estimation against the baselines \cite{ComplexInvIndoor, ComplexInvIndoorMC}. 
  Both baselines produce good albedo colors overall, but they tend to bake in the lighting and specularities into the albedo map. 
  In contrast, our method can produce clear materials with sharp edges and fine details. }
  \label{fig:supp:synthetic}
\end{figure*}

\begin{figure*}
  \setlength\tabcolsep{1.25pt}
  \centering
  \begin{tabular}{c|cc|cc}
        \includegraphics[width=0.235\textwidth]{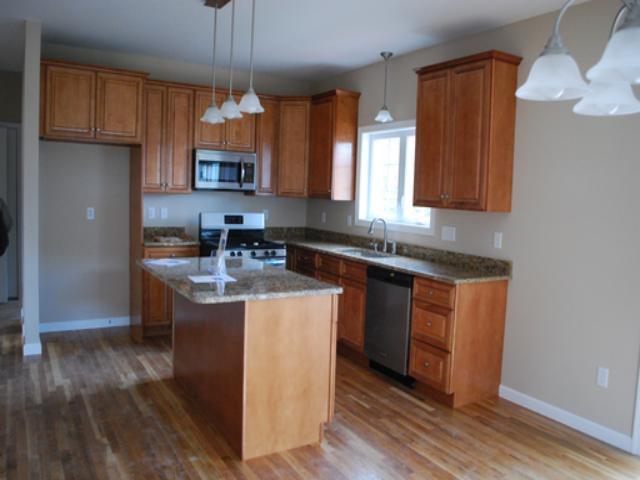}&
        \includegraphics[width=0.235\textwidth]{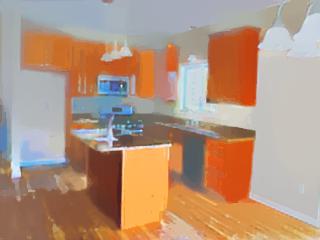}&
        \includegraphics[width=0.235\textwidth]{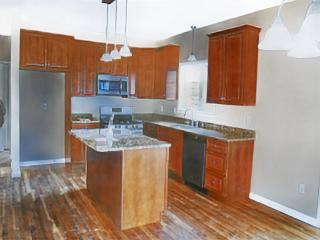}&
        \includegraphics[width=0.235\textwidth]{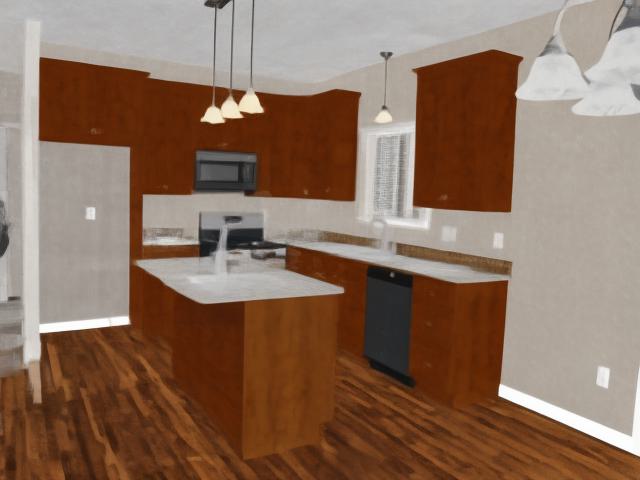}&
        \rotatebox{90}{Albedo}\\     
        &
        \includegraphics[width=0.235\textwidth]{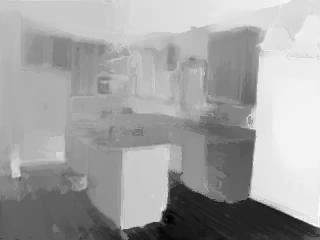}&
        \includegraphics[width=0.235\textwidth]{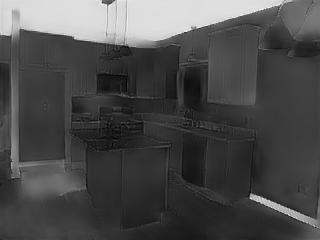}&
        \includegraphics[width=0.235\textwidth]{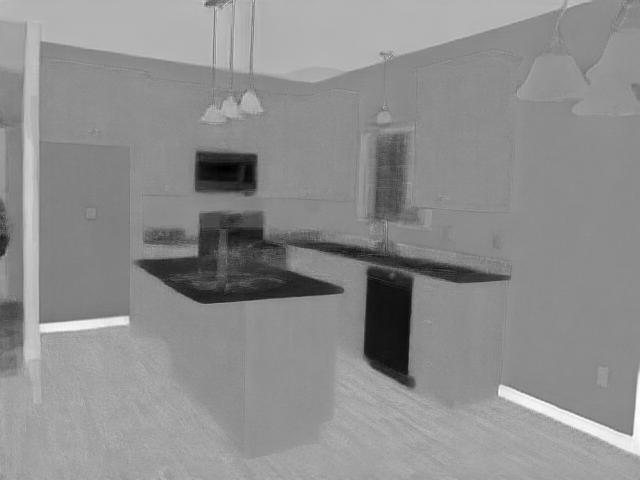}&
        \rotatebox{90}{Roughness}\\       
        &
        \includegraphics[width=0.235\textwidth]{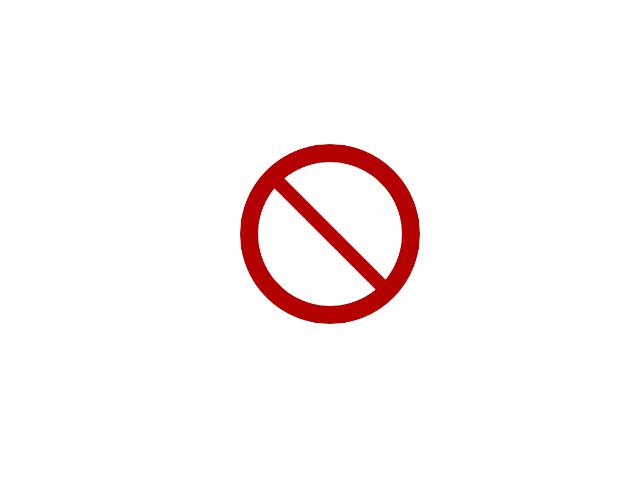}&
        \includegraphics[width=0.235\textwidth]{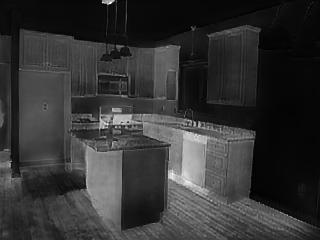}&
        \includegraphics[width=0.235\textwidth]{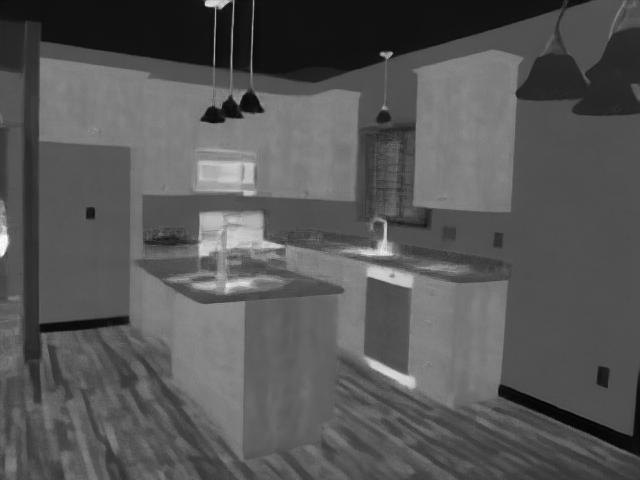}&
        \rotatebox{90}{Metallic}\\

        \includegraphics[width=0.235\textwidth]{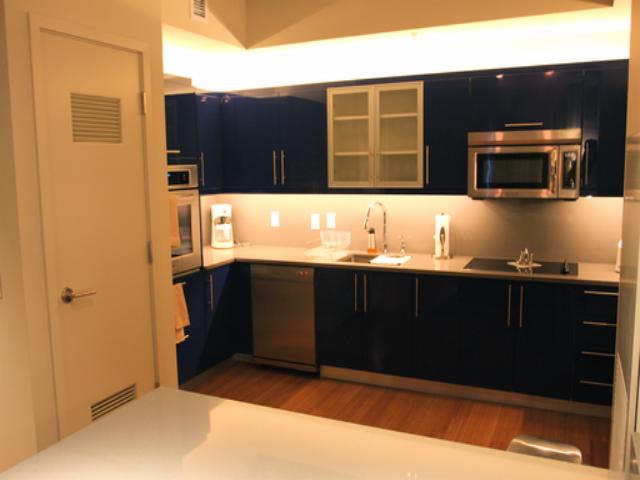}&
        \includegraphics[width=0.235\textwidth]{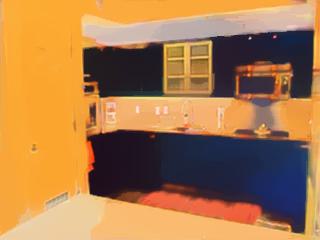}&
        \includegraphics[width=0.235\textwidth]{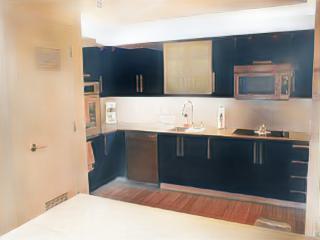}&
        \includegraphics[width=0.235\textwidth]{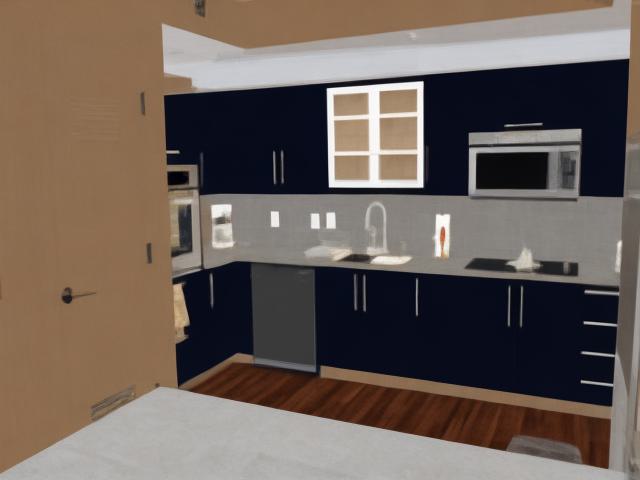}&
        \rotatebox{90}{Albedo}\\     
        &
        \includegraphics[width=0.235\textwidth]{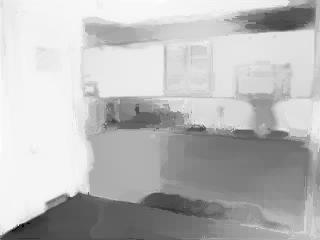}&
        \includegraphics[width=0.235\textwidth]{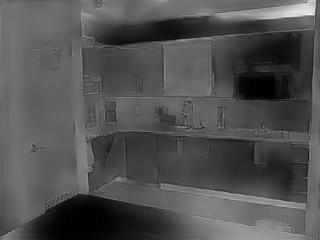}&
        \includegraphics[width=0.235\textwidth]{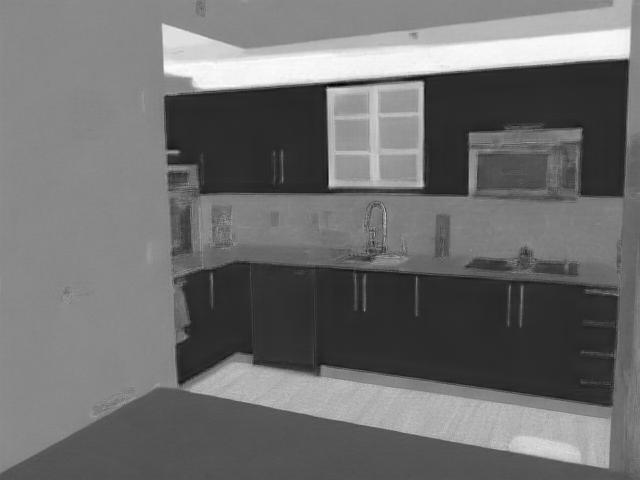}&
        \rotatebox{90}{Roughness}\\       
        &
        \includegraphics[width=0.235\textwidth]{res/supp/real/metallic_no.jpg}&
        \includegraphics[width=0.235\textwidth]{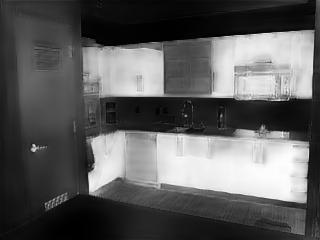}&
        \includegraphics[width=0.235\textwidth]{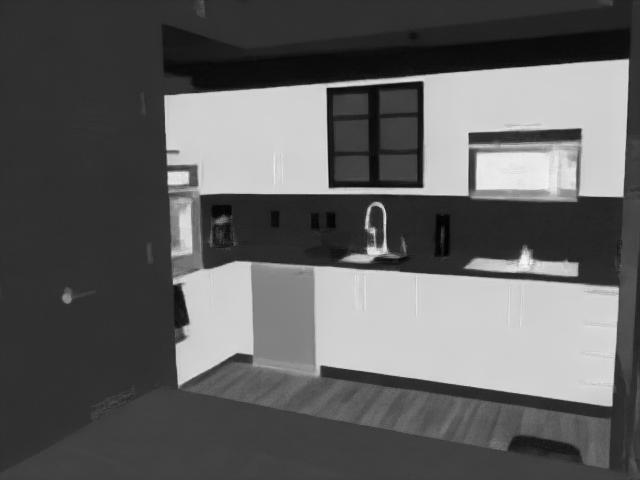}&
        \rotatebox{90}{Metallic}\\ 
  \end{tabular}
  \caption{\textbf{Real-world material estimation.} 
  Continues on the next page.}
\end{figure*}
\begin{figure*}\ContinuedFloat
  \setlength\tabcolsep{1.25pt}
  \centering
  \begin{tabular}{c|cc|c|cc}
        \includegraphics[width=0.235\textwidth]{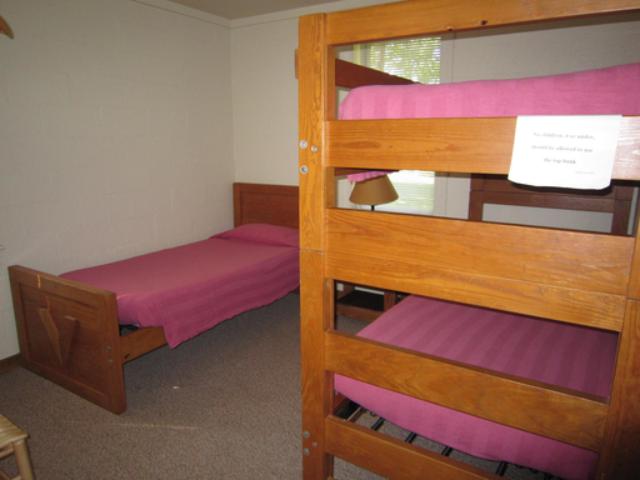}&
        \includegraphics[width=0.235\textwidth]{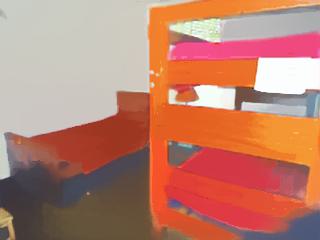}&
        \includegraphics[width=0.235\textwidth]{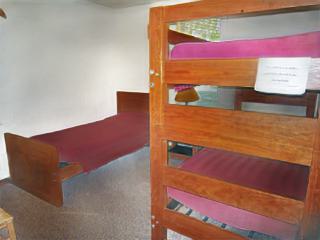}&
        \includegraphics[width=0.235\textwidth]{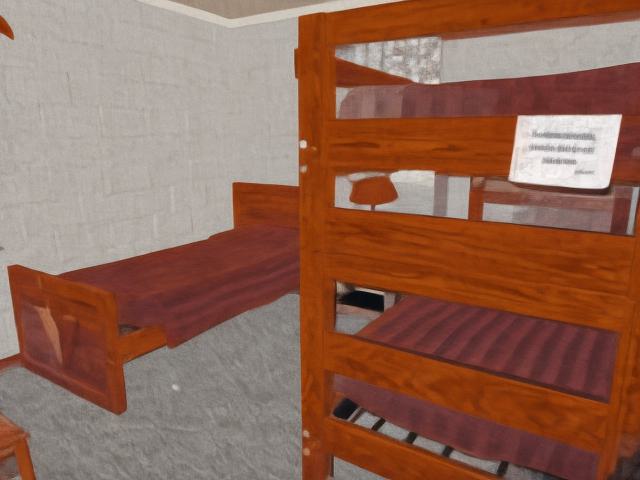}&
        \rotatebox{90}{Albedo}\\     
        &
        \includegraphics[width=0.235\textwidth]{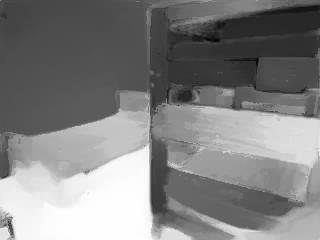}&
        \includegraphics[width=0.235\textwidth]{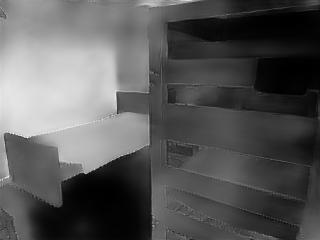}&
        \includegraphics[width=0.235\textwidth]{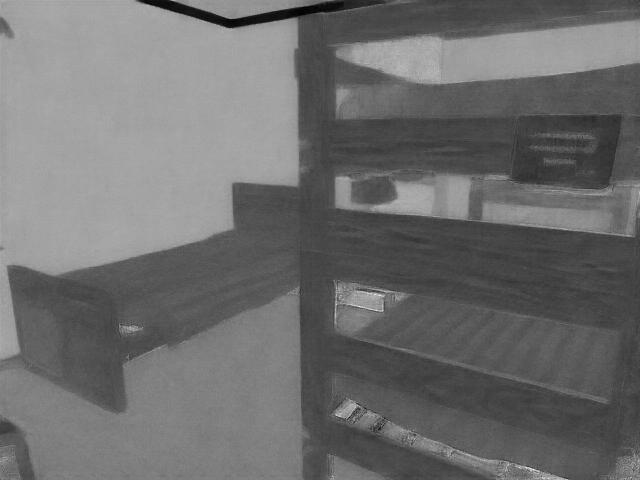}&
        \rotatebox{90}{Roughness}\\       
        &
        \includegraphics[width=0.235\textwidth]{res/supp/real/metallic_no.jpg}&
        \includegraphics[width=0.235\textwidth]{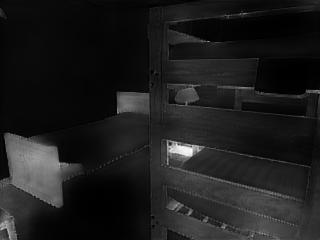}&
        \includegraphics[width=0.235\textwidth]{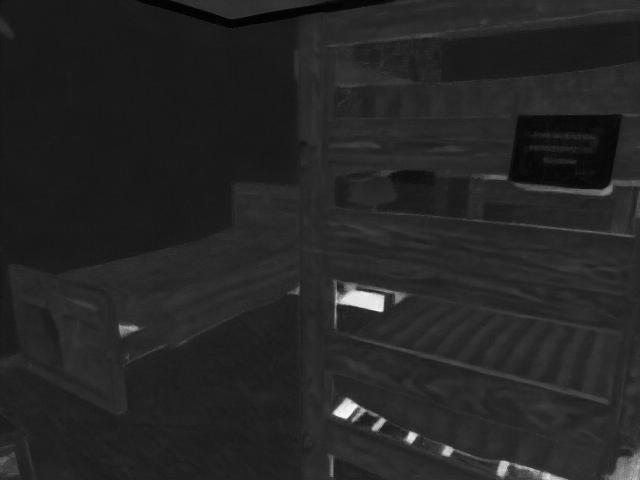}&
        \rotatebox{90}{Metallic}\\

        \includegraphics[width=0.235\textwidth]{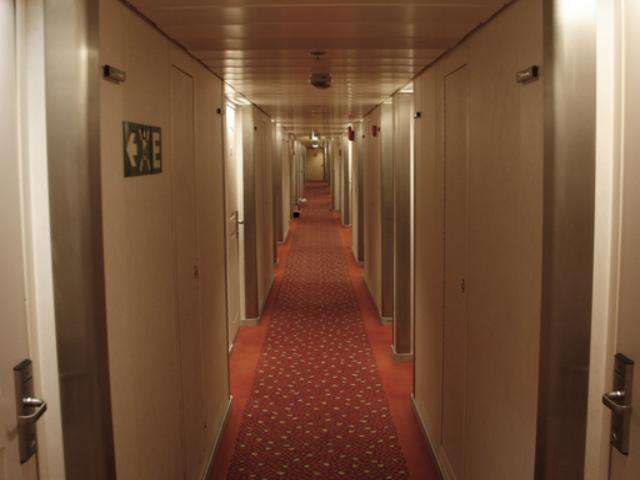}&
        \includegraphics[width=0.235\textwidth]{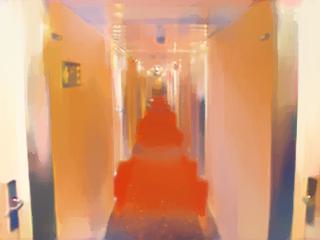}&
        \includegraphics[width=0.235\textwidth]{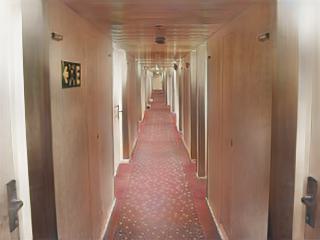}&
        \includegraphics[width=0.235\textwidth]{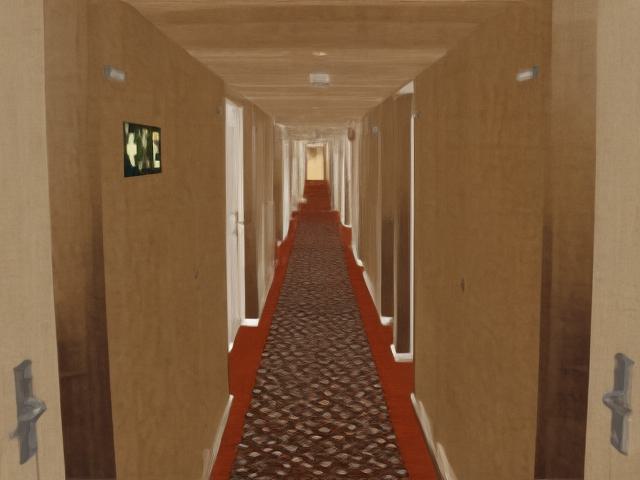}&
        \rotatebox{90}{Albedo}\\     
        &
        \includegraphics[width=0.235\textwidth]{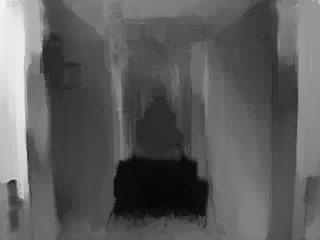}&
        \includegraphics[width=0.235\textwidth]{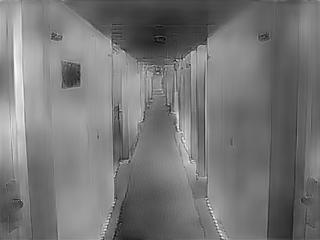}&
        \includegraphics[width=0.235\textwidth]{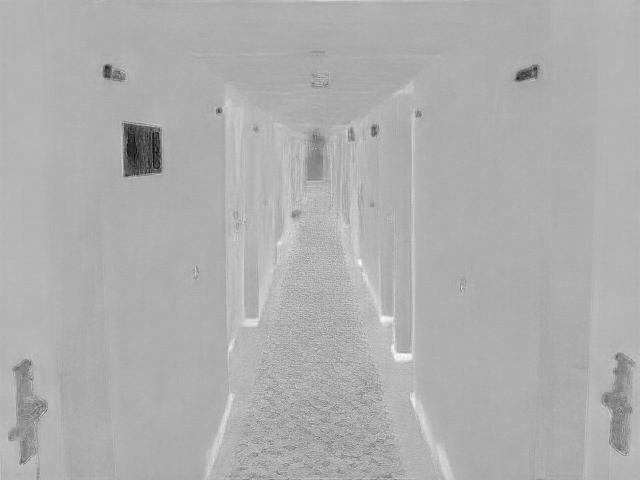}&
        \rotatebox{90}{Roughness}\\       
        &
        \includegraphics[width=0.235\textwidth]{res/supp/real/metallic_no.jpg}&
        \includegraphics[width=0.235\textwidth]{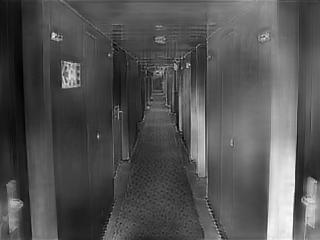}&
        \includegraphics[width=0.235\textwidth]{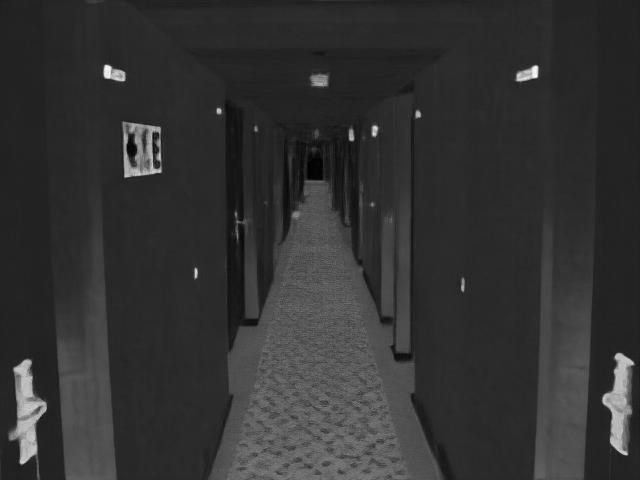}&
        \rotatebox{90}{Metallic}\\ 
        Input & \citet{ComplexInvIndoor} & \citet{ComplexInvIndoorMC} & Ours - Mean & 
  \end{tabular}
  \caption{\textbf{Real-world material estimation.} 
  We compare our material estimation against the baselines \cite{ComplexInvIndoor, ComplexInvIndoorMC}. 
  Real-world lighting and shadows pose a bigger challenge for the baselines and they often bake them into the albedo map. 
  Our method can produce sharp and detailed materials even in challenging real-world settings. }
  \label{fig:supp:real}
\end{figure*}

\end{document}